\documentclass[twoside,11pt]{article}
\usepackage{amsmath}
\usepackage{ascmac}
\usepackage{amssymb}
\usepackage{amsthm}
\usepackage{times}
\usepackage{color}
\usepackage{multirow}
\usepackage{algpseudocode}
\usepackage{bm}
\usepackage{xr}
\usepackage{url}
\usepackage{natbib}
\usepackage{enumerate}
\usepackage{tabularx}
\usepackage{booktabs}
\usepackage[labelfont=bf,format=plain,justification=raggedright,singlelinecheck=false]{caption}
\usepackage{lastpage}
\usepackage{subfigure}

\newcommand{\argmin}{\mathop{\rm argmin}\limits}
\renewcommand{\and}{\text{and}}
\newcommand{\R}[1]{\mathbb{R}^{#1}}

\newcommand{\dx}{d_{\mathrm{x}}}

\newcommand{\du}{d_{\mathrm{u}}}

\newcommand{\Dx}{D_{\mathrm{x}}}
\newcommand{\Du}{D_{\mathrm{u}}}

\newcommand{\calX}{\mathcal{X}}

\newcommand{\calYx}{\mathcal{Y}_{\mathrm{x}}}
\newcommand{\calYu}{\mathcal{Y}_{\mathrm{u}}}

\newcommand{\calU}{\mathcal{U}}

\newcommand{\calD}{\mathcal{D}}

\newcommand{\hx}[1]{h_{\mathrm{x},#1}}
\newcommand{\hu}[1]{h_{\mathrm{u},#1}}
\newcommand{\hq}[1]{\lambda_{#1}}
\newcommand{\ru}[1]{r_{\mathrm{u},#1}}
\renewcommand{\rq}[1]{r_{\mathrm{q},#1}}
\newcommand{\bhx}{\bm{h}_{\mathrm{x}}}

\newcommand{\bhu}{\bm{h}_{\mathrm{u}}}
\newcommand{\bhq}{\bm{\lambda}}
\newcommand{\bru}{\bm{r}_{\mathrm{u}}}
\newcommand{\brq}{\bm{r}_{\mathrm{q}}}

\newcommand{\Ex}{E_{\mathrm{x}}}

\newcommand{\Exu}{E_{\mathrm{xu}}}
\newcommand{\Extu}{E_{\mathrm{x}\times\mathrm{u}}}
\newcommand{\Eu}{E_{\mathrm{u}}}
\newcommand{\bExu}{\bar{E}_{\mathrm{xu}}}
\newcommand{\bExtu}{\bar{E}_{\mathrm{x}\times\mathrm{u}}}

\newcommand{\intd}{\mathrm{d}}
\newcommand{\parder}[2]{\frac{\partial #1}{\partial #2}}

\newcommand{\bx}{\bm{x}}
\newcommand{\bu}{\bm{u}}
\newcommand{\bup}{\bm{u}_{\mathrm{p}}}

\newcommand{\Jlr}{J_\mathrm{LR}}
\newcommand{\Jdv}{J_\mathrm{DV}}

\newcommand{\Jga}{J_\mathrm{\gamma}}
\newcommand{\bJga}{\bar{J}_\mathrm{\gamma}}

\newcommand{\wJdv}{\widehat{J}_\mathrm{DV}}
\newcommand{\tJdv}{\widetilde{J}_\mathrm{DV}}
\newcommand{\wJga}{\widehat{J}_\mathrm{\gamma}}
\newcommand{\tJga}{\widetilde{J}_\mathrm{\gamma}}

\newcommand{\tJnce}{\widetilde{J}_\mathrm{NCE}}

\newcommand{\byx}{\bm{y}_{\mathrm{x}}}
\newcommand{\byu}{\bm{y}_{\mathrm{u}}}

\newcommand{\rank}{\mathrm{rank}}

\newcommand{\vtheta}{\bm{\theta}}

\newcommand{\IFdv}{\mathrm{IF}_{\mathrm{DV}}}
\newcommand{\IFga}{\mathrm{IF}_{\gamma}}

\newcommand{\Vdv}{\bm{V}_{\mathrm{DV}}}

\newcommand{\bVdv}{\bar{\bm{V}}_{\mathrm{DV}}}

\newcommand{\bX}{\bm{X}}
\newcommand{\bU}{\bm{U}}
\newcommand{\bJ}{\bm{J}}

\usepackage[abbrvbib,preprint]{jmlr2e}

\jmlrheading{23}{2022}{1-\pageref{LastPage}}{12/20; Revised 6/22}{8/22}{20-1469}{Hiroaki Sasaki and Takashi Takenouchi}

\ShortHeadings{Representation learning with robust desity ratio estimation}{Sasaki and Takenouchi} \firstpageno{1}

\begin{document}
\title{Representation learning for maximization of MI, nonlinear ICA and
nonlinear subspaces with robust density ratio estimation}
\author{\name Hiroaki Sasaki \email hsasaki@fun.ac.jp \\ 
\addr Department of Complexity
and Intelligent Systems\\ Future University Hakodate\\ Hokkaido, Japan \\
\name Takashi
Takenouchi \email t-takenouchi@grips.ac.jp \\ \addr
National Graduate Institute for Policy Studies 
\\ RIKEN AIP\\ Tokyo, Japan}
%
\editor{Stefan Harmeling}
\maketitle
\begin{abstract}
 Unsupervised representation learning is one of the most important
 problems in machine learning. A recent promising approach is
 contrastive learning: A feature representation of data is learned by
 solving a \emph{pseudo} classification problem where class labels are
 automatically generated from unlabelled data. However, it is not
 straightforward to understand what representation contrastive learning
 yields through the classification problem. In addition, most of
 practical methods for contrastive learning are based on the maximum
 likelihood estimation, which is often vulnerable to the contamination
 by outliers. In order to promote the understanding to contrastive
 learning, this paper first theoretically shows a connection to
 maximization of mutual information (MI). Our result indicates that
 density ratio estimation is necessary and sufficient for maximization
 of MI under some conditions. Since popular objective functions for
 classification can be regarded as estimating density ratios,
 contrastive learning related to density ratio estimation can be
 interpreted as maximizing MI. Next, in terms of density ratio
 estimation, we establish new recovery conditions for the latent source
 components in nonlinear independent component analysis (ICA). In
 contrast with existing work, the established conditions include a novel
 insight for the dimensionality of data, which is clearly supported by
 numerical experiments. Furthermore, inspired by nonlinear ICA, we
 propose a novel framework to estimate a nonlinear subspace for
 lower-dimensional latent source components, and some theoretical
 conditions for the subspace estimation are established with density
 ratio estimation. Motivated by the theoretical results, we propose a
 practical method through outlier-robust density ratio estimation, which
 can be seen as performing maximization of MI, nonlinear ICA or
 nonlinear subspace estimation.  Moreover, a sample-efficient nonlinear
 ICA method is also proposed based on a variational lower-bound of
 MI. Then, we theoretically investigate outlier-robustness of the
 proposed methods. Finally, we numerically demonstrate usefulness of the
 proposed methods in nonlinear ICA and through application to a
 downstream task for linear classification.
\end{abstract}
 \section{Introduction}
 \label{sec:Intro}
 Unsupervised representation learning has been a long-term issue in
 machine learning~\citep{raina2007self,bengio2013representation}. In
 contrast with supervised learning, a feature representation of data is
 learned only from unlabelled data. Since the acquisition cost of
 unlabelled data is often low, the learned representation would be
 useful particularly when the number of data is very limited in a
 downstream task (e.g., supervised tasks). Successful examples include
 video
 classification~\citep{Wang_2015_ICCV,arandjelovic2017look,sun2019learning},
 motion capture~\citep{NIPS2017_ab452534}, and natural language
 processing~\citep{peters2018deep,devlin2019bert}. More examples can be
 found in recent review papers for representation
 learning~\citep{jing2020self,liu2020self}.

 Recently, a number of methods for unsupervised representation learning
 have been proposed, and most of them can be roughly divided into three
 categories based on variational autoencoder
 (VAE)~\citep{kingma2014stochastic,Higgins2017betaVAELB,pmlr-v108-khemakhem20a},
 generative adversarial network
 (GAN)~\citep{goodfellow2014generative,chen2016infogan} and contrastive
 learning~\citep{dosovitskiy2014discriminative,doersch2015unsupervised,noroozi2016unsupervised}.
 VAE learns a useful representation of data by maximizing a tractable
 lower-bound of the likelihood function. However, VAE makes a
 restrictive prior assumption on densities such as the Gaussian
 assumption. The GAN approach learns both a representation and generator
 of data, but it requires to solve a max-min optimization problem, which
 could be numerically unstable. On the other hand, contrastive learning
 does not make strong assumptions on densities, and simply solves a
 straightforward optimization problem.
 
 In order to learn a feature representation of data, contrastive
 learning solves a \emph{pseudo} classification problem where class
 labels are automatically generated from unlabelled data. For example,
 \citet{arandjelovic2017look} make a dataset for binary classification
 from unlabelled video data: The positive labels are assigned to pairs
 of video frames and audio clips taken from the same video, while the
 negative labels are given to pairs from different videos. Then,
 representations of both video frames and audio clips are learned by
 solving the binary classification problem. \citet{misra2016shuffle}
 also solve a binary classification problem for unsupervised
 representation learning where the positive data is temporally
 consecutive video frames in a video, but the negative one is temporally
 shuffled frames in the same video. This learning scheme based on
 classification problems is also called \emph{self-supervised learning}
 (SSL). See a recent review article for SSL~\citep{liu2020self} and
 references therein. However, it is not straightforward to understand
 what representation contrastive learning yields by solving
 classification problems.
 
 Contrastive learning has been justified as being closely related to a
 classical framework so called \emph{maximization of mutual information}
 (MI)~\citep{linsker1989application,bell1995information}.  This comes
 from an intuitive idea that MI must be high if a classifier can
 accurately distinguish positive samples drawn from the joint density
 (e.g., for image frames and audio clips) and negative ones drawn from
 the product of the marginal
 densities~\citep{belghazi2018mutual,hjelm2019learning,tschannen2019mutual}.
 However, this intuitive idea is rather superficial because it is not
 necessarily obvious why solving the classification problem leads to
 maximization of MI. A recent promising approach is based on
 maximization of tractable variational lower-bounds of MI and also
 related to contrastive
 learning~\citep{nguyen2008estimating,belghazi2018mutual,oord2018representation},
 but the original purpose of maximizing the variational lower-bounds is
 \emph{estimation} of MI, not necessarily maximization of MI.
 More rigorous justification would be preferable to understand how and
 when contrastive learning can be regarded as performing maximization of
 MI.
 
 Contrastive learning has been employed in recent algorithms of
 \emph{nonlinear independent component analysis}
 (ICA)~\citep{hyvarinen2016unsupervised,pmlr-v54-hyvarinen17a,hyvarinen2018nonlinear},
 which is a solid framework for unsupervised representation learning.
 Nonlinear ICA rigorously defines a generative model where the input
 data is assumed to be observed as a general nonlinear mixing of the
 latent source components. Then, the problem is to recover the source
 components from the observations of input data. Regarding the linear
 ICA where the mixing function is linear, the recovery conditions are
 well-established and the important condition is mutual independence of
 the source components~\citep{comon1994independent}. On the other hand,
 the problem of nonlinear ICA has been proved to be fundamentally
 ill-posed under the same independence condition because there exist an
 infinite number of decompositions of a random vector into mutually
 independent
 variables~\citep{hyvarinen1999nonlinear,pmlr-v97-locatello19a}.  Very
 recently, novel recovery conditions for nonlinear ICA have been
 established~\citep{sprekeler2014extension,hyvarinen2016unsupervised,pmlr-v54-hyvarinen17a,hyvarinen2018nonlinear}.
 The main idea is to introduce additional data called the
 \emph{complementary data} in this paper, and alternative condition to
 mutual independence is conditional independence of the source
 components given complementary data. An example of the complementary
 data is time segment labels obtained by dividing time series data into
 a number of time segments~\citep{hyvarinen2016unsupervised}, while
 \citet{pmlr-v54-hyvarinen17a} employed the history of time-series data
 as complementary data. With the complementary data, contrastive
 learning based on the logistic regression has been performed for
 nonlinear ICA.

 Most of practical methods in contrastive learning have been based on
 maximum likelihood estimation (MLE): In logistic regression, the
 conditional probability (i.e., posterior probability) is estimated by
 MLE, while maximizing a variational lower-bound of MI can be
 interpreted as performing MLE as we show later. MLE has a number of
 excellent properties such the asymptotic
 efficiency~\citep{wasserman2006all}, but it is known not to be robust
 against outliers: If data is contaminated by outliers, MLE often
 suffers from a severe bias. Thus, estimation of existing methods for
 maximization of MI and nonlinear ICA might be magnified by
 outliers. This is a serious problem because outliers are observed in
 many practical situations.  For instance, the contamination of outliers
 has been an important issue in functional MRI
 data~\citep{poldrack2012future} to which ICA methods have been
 applied~\citep{monti2019causal}.
 
 This paper first shows that unsupervised representation learning on
 three frameworks can be performed through density ratio estimation.
 Previous work has already discussed that contrastive learning can be
 seen as density ratio estimation when popular objective functions are
 used for classification such as the cross
 entropy~\citep{nguyen2008estimating,belghazi2018mutual,oord2018representation}.
 Our primary contributions are to perform theoretical analysis on the
 three frameworks in terms of density ratio estimation. The first
 framework to be tackled is maximization of MI. We show that density
 ratio estimation is necessary and sufficient for maximization of MI
 under some conditions. This result supports the intuitive belief above
 between contrastive learning and maximization of MI, and clarifies when
 and how contrastive learning can be considered as performing
 maximization of MI. The second framework is nonlinear ICA, where we
 provide two new proofs for source recovery in terms of density ratio
 estimation. The key point is that the recovery conditions in both
 proofs include a novel insight, which has not been seen in previous
 work of nonlinear ICA~\citep{hyvarinen2018nonlinear}: The
 dimensionality of complementary data is an important factor for source
 recovery. This insight is clearly supported by experimental results
 that the latent sources are more accurately recovered as the
 dimensionality of complementary data increases. Furthermore, one of the
 proofs can be seen as a generalization
 of~\citet{pmlr-v54-hyvarinen17a}. The third framework is
 \emph{nonlinear subspace estimation} proposed in this paper. This
 framework is inspired by nonlinear ICA, and aimed at estimating a
 nonlinear subspace of lower-dimensional latent source components. To
 this end, we propose a novel generative model where data is generated
 as a nonlinear mixing of lower-dimensional source components and
 nuisance variables. Unlike nonlinear ICA, the source components are not
 necessarily assumed to be conditionally independent, and thus the
 proposed generative model is more general than nonlinear ICA in terms
 of the conditional independence. Moreover, as in nonlinear ICA, with
 density ratio estimation, we establish theoretical conditions that a
 nonlinear subspace of the lower-dimensional source components, which is
 separated from the nuisance variables, can be estimated.

 Motivated by the density-ratio view, we propose an outlier-robust
 method for unsupervised representation learning. The proposed method
 performs density ratio estimation based on a robust alternative to the
 KL-divergence called the
 \emph{$\gamma$-divergence}~\citep{fujisawa2008robust}, which has a
 favorable robustness property expressed as the \emph{strong
 robustness}~\citep{cichocki2010families,amari2016information}: The
 latent bias caused from outliers can be small even in the case of heavy
 contamination of outliers. Furthermore, we develop a new method of
 nonlinear ICA by applying an existing variational lower-bound of
 MI~\citep{belghazi2018mutual}. Then, we theoretically investigate
 outlier-robustness of these methods. Finally, we numerically
 demonstrate that the proposed method based on the $\gamma$-divergence
 is very robust against outliers both in nonlinear ICA and a downstream
 task for linear classification, while the nonlinear ICA method based on
 the variational lower-bound is experimentally shown to perform better
 than existing ICA methods when the number of data is small.

 This paper is organized as follows\footnote{A very preliminary version
 of this paper was published in~\citet{pmlr-v124-sasaki20b}.}:
 Section~\ref{sec:background} formulates the problem of density ratio
 estimation for unsupervised representation learning, and reviews
 existing works for contrastive learning, variational MI estimation and
 nonlinear ICA. Section~\ref{sec:unified-view} theoretically shows that
 unsupervised representation learning on the three frameworks can be
 performed through density ratio estimation, and discusses theoretical
 contributions in each of the
 frameworks. Section~\ref{sec:estimation-methods} proposes two practical
 methods for unsupervised representation learning and theoretically
 analyzes them in terms of outlier-robustness. Section~\ref{sec:exp}
 numerically demonstrates usefulness of the proposed methods in
 nonlinear ICA and a downstream task for linear
 classification. Section~\ref{sec:conclusion} concludes this paper.
 \section{Problem formulation and background}
 \label{sec:background}
 This section first formulates the problem of unsupervised
 representation learning based on density ratio estimation, and then
 reviews existing works for contrastive learning, variational estimation
 of mutual information and nonlinear independent component analysis.

 \begin{table}
  \caption{List of notations used in Section~\ref{sec:background}.}
  \vspace{1mm}
  \begin{tabular}{|p{0.4\textwidth}|p{0.6\textwidth}|}
   \hline
   $\bm{x}\in\R{Dx}$ & Input data\\ 
   $\bm{u}\in\R{\Du}$ & Complementary data\\
   $\Dx$ & Dimensionality of input data \\
   $\Du$ & Dimensionality of complementary data \\
   $p(\bm{x},\bm{u})$ & Joint probability density function of $\bm{x}$ and $\bm{u}$\\
   $p(\bm{x})$ & Marginal probability density function of $\bm{x}$\\
   $p(\bm{u})$ & Marginal probability density function of $\bm{u}$\\
   $\perp$ & Statistical independence \\ 
   $\bm{u}^*\in\R{\Du}$& $\bm{u}^*\sim{p}(\bm{u})$ and
       $\bm{u}^*\perp\bm{x}$\\ 
   $\bhx(\bm{x}):\R{\Dx}\to\R{\dx}$
   & Representation function of $\bm{x}$\\
   $\bhu(\bm{u}):\R{\Du}\to\R{\du}$ 
   & Representation function of $\bm{u}$\\
   $\dx$ & Dimensionality of representation function $\bhx(\bx)$\\
   $\du$ & Dimensionality of representation function $\bhu(\bu)$\\
   $r(\bm{x},\bm{u})$ & Model to approximate
       $\log\frac{p(\bm{x},\bm{u})}{p(\bm{x})p(\bm{u})}$\\
   $\psi:\R{\dx}\times\R{\du}\to\R{}$ & Function in $r(\bm{x},\bm{u})$ \\
   $a:\R{\dx}\to\R{}$ & Function in $r(\bm{x},\bm{u})$ \\
   $b:\R{\du}\to\R{}$ & Function in $r(\bm{x},\bm{u})$ \\
   $\Jlr$ & Cross entropy in logistic regression \\
   \hline
  \end{tabular}
 \end{table}

  \subsection{Problem formulation of density ratio estimation for representation learning}
  Suppose that we are given $T$ pairs of two data samples drawn from the
  joint distribution with density $p(\bm{x},\bm{u})$:
  \begin{align}
   \calD:=\left\{(\bm{x}(t)^{\top},\bm{u}(t)^{\top})^{\top}~|~%
   \bm{x}(t)=(x_{1}(t),\dots,x_{\Dx}(t))^{\top},
   \bm{u}(t)=(u_{1}(t),\dots,u_{\Du}(t))^{\top}\right\}_{t=1}^T,
   \label{dataset}
  \end{align}
  where $\bm{x}(t)$ and $\bm{u}(t)$ denote the $t$-th observations of
  $\bm{x}$ and $\bm{u}$, respectively. Throughout the paper, we call
  $\bm{x}$ and $\bm{u}$ as \emph{input} and \emph{complementary} data,
  respectively. Our primary goal is to estimate
  $\bhx(\bm{x})=(\hx{1}(\bx),\dots,\hx{\dx}(\bx))^{\top}$ called a
  \emph{representation function} of $\bm{x}$ such that the logarithmic
  ratio of the joint density to the product of the marginal densities
  $p(\bm{x})$ and $p(\bm{u})$,
  \begin{align}
   \log\frac{p(\bm{x},\bm{u})}{p(\bm{x})p(\bm{u})}, \label{ratio}
  \end{align}
  is accurately approximated up to a constant under the following basic
  form of a model $r(\bx,\bu)$:
  \begin{align}
   r(\bx,\bu):=\psi(\bhx(\bm{x}),\bhu(\bm{u}))+a(\bhx(\bx))+b(\bhu(\bu)),
   \label{defi-h}
  \end{align}
  where $\bhu(\bu)=(\hu{1}(\bu),\dots,\hu{\du}(\bu))^{\top}$ is a
  representation function of $\bu$, and $\psi(\cdot, \cdot)$, and
  $a(\cdot)$ and $b(\cdot)$ are scalar functions. For nonlinear ICA,
  $\psi(\bhx(\bm{x}),\bhu(\bm{u}))$ is slightly modified: In
  Theorem~\ref{theo:general-ICA}, $\psi(\bhx(\bm{x}),\bhu(\bm{u}))$ is
  expressed by the sum of elementwise functions with respect to
  $\bhx(\bm{x})$ (not $\bhu(\bm{u}))$ as
  $\psi(\bhx(\bm{x}),\bhu(\bm{u}))=\sum_{i=1}^{\dx}\psi_i(\hx{i}(\bm{x}),\bhu(\bm{u}))$.
  All of $\bhu$, $\psi$, $a$ and $b$ are also estimated from
  $\calD$. Here, we assume that the dimensionalities of the
  representation functions are smaller than or equal to ones of input
  and complementary data, i.e., $\du\leq\Du$ and
  $\dx\leq\Dx$. Furthermore, in order to simplify the review of existing
  works and our theoretical results in Section~\ref{sec:unified-view},
  it is supposed that data samples is \emph{not} contaminated by
  outliers, while we propose a practical method and theoretically
  investigate its robustness under the contamination of outliers in
  Section~\ref{sec:estimation-methods}.

  Function $\psi(\bhx(\bx),\bhu(\bu))$ in~\eqref{defi-h} takes a role of
  capturing the statistical dependencies between $\bx$ and $\bu$ with
  dimensionality reduction, and has been modeled previously such as
  $\psi(\bhx,\bhu)=\bhx^\top\bhu$~\citep{bachman2019learning} and
  $\psi(\bhx,\bhu)=\bhx^\top\bm{W}\bhu$~\citep{oord2018representation,tian2019contrastive}
  where $\bm{W}$ is a $\dx$ by $\du$ matrix.  More generally,
  $\psi(\bhx,\bhu)$ can be modeled by a feedforward neural
  network~\citep{arandjelovic2017look}. Scalar functions of
  $a(\bhx(\bx))$ and $b(\bhu(\bu))$ express any functions in the
  log-density ratio~\eqref{ratio} that depend only on either $\bx$ or
  $\bu$.  Section~\ref{sec:unified-view} often assumes that there exist
  functions $\psi^{\star}$, $\bhx^{\star}$, $\bhu^{\star}$, $a^{\star}$
  and $b^{\star}$ such that the logarithmic density ratio is universally
  approximated at $\psi=\psi^{\star}$, $\bhx=\bhx^{\star}$,
  $\bhu=\bhu^{\star}$, $a=a^{\star}$ and $b=b^{\star}$ in~\eqref{defi-h}
  as follows:
  \begin{align*}
   \log\frac{p(\bm{x},\bm{u})}{p(\bm{x})p(\bm{u})}
   =\psi^{\star}(\bhx^{\star}(\bm{x}),\bhu^{\star}(\bm{u}))+a^{\star}(\bhx^{\star}(\bx))+b^{\star}(\bhu^{\star}(\bu)).
  \end{align*}
  This universal approximation assumption would be realistic when the
  log-density ratio is a continuous function and neural networks are
  employed for modelling $\psi$, $\bhx$, $\bhu$, $a$ and
  $b$~\citep{hornik1991approximation}. Section~\ref{sec:estimation-methods}
  proposes practical methods to approximate $\psi^{\star}$,
  $\bhx^{\star}$, $\bhu^{\star}$, $a^{\star}$ and $b^{\star}$ from data
  samples.
  
  The approach of using the complementary data is recently becoming more
  popular as in multi-view learning~\citep{li2018survey}.  Furthermore,
  it might not be so expensive to obtain complementary data samples, but
  they are rather generated from a single unlabeled dataset in many
  practical situations. Let us list the following examples:
  \begin{itemize}
   \item Denoising autoencoder makes complementary data samples
	 $\bm{u}(t)$ by injecting noises to input data samples
	 $\bm{x}(t)$~\citep{vincent2010stacked}.
	 
   \item $\bm{x}(t)$ and $\bm{u}(t)$ can be drawn by dividing a single
	 video data into image (i.e., a video frame) and sound data
	 samples at each time index $t$,
	 respectively~\citep{arandjelovic2017look}.
	 
   \item Regarding time series data $\bm{x}(t)$ including video data,
	 past input data is an example of the complementary data (e.g.,
	 $\bm{u}(t)=\bm{x}(t-1)$)~\citep{misra2016shuffle,pmlr-v54-hyvarinen17a}.
	 
   \item Suppose that $\bm{x}(t)$ are image patches extracted from a
	 large single image where the index $t$ conveys some positional
	 information of patches. Then, pairs of image patches at $t$ and
	 $t' (\neq{t})$ can be used as input and complementary data
	 samples, that is,
	 $\bm{u}(t)=\bm{x}(t')$~\citep{noroozi2016unsupervised}.

   \item Clustering labels to input data samples $\bm{x}(t)$ can be used
	 as the complementary data samples
	 $\bm{u}(t)$~\citep{caron2018deep}.

   \item In~\citet{hjelm2019learning}, $\bx(t)$ is an image, while
	 $\bu(t)$ is a smaller image patch extracted from the single
	 image $\bx(t)$.
  \end{itemize} 
  
  Density ratio estimation is useful particularly when neural networks
  are used for $r(\bx,\bu)$ because the density ratio is invariant under
  any invertible transformations or reparametrizations of $\bm{x}$
  and/or $\bm{u}$. This invariant property has been exploited by noise
  contrastive estimation as well~\citep{Gutmann2012a}.  In addition to
  the practical usefulness, in this paper, we theoretically show that
  three frameworks for unsupervised representation learning can be
  performed by density ratio estimation.
  \subsection{Contrastive learning and variational estimation of mutual information}
  \label{ssec:CL-MI}
  In order to estimate the representation functions, contrastive
  learning solves a classification problem where class labels are
  automatically generated from unlabelled data. A common setting is
  based on the following two datasets:
  \begin{align}
   \calD_+:=\{(\bm{x}(t),
   \bm{u}(t)\}_{t=1}^{T}\sim{p}(\bx,\bu)\quad\text{vs.}
   \quad\calD_-:=\{(\bm{x}(t),\bm{u}^*(t))\}_{t=1}^{T}\sim{p}(\bx)p(\bu),
   \label{binary-classification}
  \end{align}
  where $\bm{u}^*(t)$ is a $\Du$-dimensional data vector sampled from
  the marginal density of $\bm{u}$, and thus independent to
  $\bm{x}(t)$. In practice, $\bm{u}^*(t)$ can be generated by randomly
  shuffling $\bm{u}(t)$ with respect to $t$ under the
  i.i.d. assumption. Interestingly, the random shuffling has been
  heuristically used in a number of previous
  works~\citep{misra2016shuffle,Lee_2017_ICCV}.  One of the most popular
  objective functions in contrastive learning is the following cross
  entropy for binary classification used in logistic regression:
  \begin{align}
   \Jlr(r):=-\Exu\left[
   \log\frac{e^{r(\bX,\bU)}}{1+e^{r(\bX,\bU)}}
   \right]-\Extu\left[\log\frac{1}{1+e^{r(\bX,\bU)}}\right],
   \label{logistic}
  \end{align}
  where $\Exu$ and $\Extu$ denote the expectations over $p(\bx,\bu)$ and
  $p(\bx)p(\bu)$, respectively. The representation functions $\bhx(\bx)$
  and $\bhu(\bu)$ in $r(\bx,\bu)$ can be estimated by minimizing the
  empirical version of $\Jlr(r)$. Logistic regression has been
  previously used to estimate a density
  ratio~\citep{sugiyama2012density-ratio}, and the minimizer of
  $\Jlr(r)$ with respect to $r$ is equal to
  \begin{align*}
   \log\frac{p(\bx,\bu)}{p(\bx)p(\bu)}
  \end{align*}
  up to a constant. Intuitively, minimizing $\Jlr(r)$ enables to
  well-capture statistical dependencies between $\bx$ and $\bu$ by
  contrasting $\calD_+$ with $\calD_-$, and thus to estimate data
  representations $\bhx(\bx)$ and $\bhu(\bu)$ having high mutual
  information. We theoretically justify this intuition, and clarifies
  when contrastive learning based on $\Jlr(r)$ can be considered to
  maximize mutual information.
  
  \emph{Infomax} is a classical framework for unsupervised
  representation learning, and advocates maximizing mutual information
  (MI) between input data and its
  representation~\citep{linsker1989application,bell1995information}.
  However, in practice, MI usually requires density estimation, which
  makes it hard to apply neural networks because of the notorious
  partition function problem. Recent promising approach employs
  complementary data and alternatively maximizes variational
  lower-bounds of MI between input and complementary data. For instance,
  the following lower bounds are often employed:
  \begin{align}
   I(\calX,\calU)\geq&\Exu[r(\bX,\bU)]-\Extu[e^{r(\bX,\bU)-1}]
   &&\text{\citep{nguyen2008estimating,sugiyama2008direct}} \label{f-div}
   \\ I(\calX,\calU)\geq&\Exu[r(\bX,\bU)]-\log\Extu[e^{r(\bX,\bU)}]
   &&\text{\citep{ruderman2012tighter,belghazi2018mutual}}, \label{DV}
  \end{align}  
  where $I(\calX,\calU)$ denotes mutual information between $\bx$ and
  $\bu$, and is defined by
  \begin{align*}
   I(\calX,\calU):=\int
   p(\bm{x},\bm{u})\log\frac{p(\bm{x},\bm{u})}{p(\bm{x})
   p(\bm{u})}\intd\bm{x}\intd\bm{u}.
  \end{align*}   
  For other lower-bounds of MI, we refer
  to~\citet{poole2019variational}.  The key advantage is that these
  lower bounds enable us to employ neural networks without any special
  efforts, while it comes at the price for a statistical limitation that
  these lower-bounds may require an exponentially number of samples to
  accurately estimate
  MI~\citep[Theorem~3.1]{pmlr-v108-mcallester20a}. As in logistic
  regression~\eqref{logistic}, the maximizers of both lower bounds
  in~\eqref{f-div} and~\eqref{DV} with respect to $r$ have been shown to
  be equal to
  \begin{align*}
   \log\frac{p(\bx,\bu)}{p(\bx)p(\bu)}
  \end{align*}
  up to constants~\citep{nguyen2008estimating,ruderman2012tighter}.
  Thus, these lower bounds can be used for density ratio estimation. The
  primary purpose of maximizing these lower-bounds is to estimate mutual
  information between $\bx$ and $\bu$, yet it has been believed that
  maximizing these lower bounds leads to maximization of mutual
  information between $\bhx(\bx)$ and $\bhu(\bu)$. Here, we provide a
  theoretically rigorous support to this belief.
  \subsection{Nonlinear independent component analysis} 
  \label{ssec:rev-nonlinear-ICA}
  Nonlinear ICA is a solid framework for unsupervised representation
  learning, and assumes that data $\bm{x}=(x_1,\dots,x_{\Dx})^{\top}$ is
  generated as the following nonlinear mixing of the latent source
  $\bm{s}=(s_1,\dots, s_{\Dx})^{\top}$:
  \begin{align}
   \bm{x}=\bm{f}(\bm{s}), \label{ICA-model}
  \end{align}
  where $\bm{f}(\bm{s})=(f_1(\bm{s}), \dots, f_{\Dx}(\bm{s}))^{\top}$,
  and $\bm{f}$ is an invertible function. The problem is to recover (or
  identify) the latent source components $s_i$. In the case of the
  linear mixing, i.e., $\bm{f}(\bm{s})=\bm{A}\bm{s}$ with an invertible
  matrix $\bm{A}$, the latent source $\bm{s}$ can be recovered up to the
  permutation (i.e., ordering) and scales of $s_1, s_2,\dots, {s}_{\Dx}$
  when they are mutual independent and follow a nonGaussian
  density~\citep{comon1994independent}. However, the problem of
  nonlinear ICA has been proved to be seriously illposed under the same
  condition as the linear case because there exist an infinite number of
  decompositions of a random vector into mutually independent
  variables~\citep{hyvarinen1999nonlinear,pmlr-v97-locatello19a}.
  Nonetheless, a number of methods for nonlinear ICA have been
  previously
  proposed~\citep{tan2001nonlinear,almeida2003misep,blaschke2007independent},
  but most of them lack theoretical guarantees for source recovery and
  it is thus unclear to what extent these methods can recover the source
  components under the nonlinear mixing function $\bm{f}$.

  Recently, novel recovery conditions for nonlinear ICA have been
  established~\citep{sprekeler2014extension,
  hyvarinen2016unsupervised,pmlr-v54-hyvarinen17a,hyvarinen2018nonlinear}.
  The key condition alternative to mutual independence is conditional
  independence of $s_1, s_2,\dots, {s}_{\Dx}$ given some complementary
  data $\bm{u}$. Time contrastive learning (TCL) employs time segment
  labels as complementary data samples, and assumes that the conditional
  density of the source $\bm{s}$ given a time segment label is
  conditionally independent and belongs to an exponential
  family~\citep{hyvarinen2016unsupervised}. Then, the representation
  function $\bhx(\bm{x})$ learned by the multinomial logistic regression
  has been shown to asymptotically correspond to elementwise nonlinear
  functions of the latent source components $s_1,s_2,\dots,{s}_{\Dx}$ up
  to a linear transformation. \citet{hyvarinen2018nonlinear} removed the
  exponential family assumption in TCL and has established recovery
  conditions in terms of more general conditional densities
  (nonexponential family). Then, it was proved that the representation
  function $\bhx(\bx)$ learned by contrastive learning based on
  $\Jlr(r)$ (i.e., logistic regression) is asymptotically equal to the
  latent source components $s_1,s_2,\dots,{s}_{\Dx}$ up to their
  permutation and elementwise invertible functions. Details of the
  recovery conditions in~\citet{hyvarinen2018nonlinear} are discussed in
  Section~\ref{ssec:nonlinear-ICA}. Nonlinear ICA has been applied to
  causal analysis~\citep{monti2019causal,pmlr-v108-wu20b} and transfer
  learning~\citep{pmlr-v119-teshima20a}.

  This paper provides two new recovery proofs for nonlinear ICA, both of
  which include a novel insight: The dimensionality of complementary
  data is an important factor for source recovery. This insight is
  clearly supported by numerical experiments. Furthermore, we propose a
  novel generative model where data is generated as a nonlinear mixing
  of lower-dimensional latent source components and nuisance
  variables. The proposed generative model is more general than the
  one~\eqref{ICA-model} in nonlinear ICA in the sense that the
  conditional independence of the latent source components is no longer
  assumed. Based on the proposed generative model, we establish
  theoretical conditions that complementary data enables us to
  automatically ignore the nuisance variables, and to estimate a
  nonlinear subspace related to only the lower-dimensional latent source
  components.
  \section{Three frameworks for unsupervised representation learning through  density ratio estimation}
 \label{sec:unified-view}
  This section shows that unsupervised representation learning on three
  frameworks can be performed by density ratio estimation, and discusses
   theoretical contributions on each of the
  frameworks. Tables~\ref{tab:notations-section3} and~\ref{tab:summary}
  are a list of notations and summary of theoretical conditions in the
  three frameworks, respectively.
  \renewcommand{\arraystretch}{1.2}
  \begin{table}[t]
   \caption{\label{tab:notations-section3} List of notations mainly used
   in Section~\ref{sec:unified-view}. Wrt is an abbreviation of ``with
   respect to''.}
   \vspace{1mm}
   \begin{tabular}{|p{0.4\textwidth}|p{0.6\textwidth}|}
    \hline
   $\bhx^{\star}, \bhu^{\star}$ & Optimal representation functions of
       $\bhx$ and $\bhu$ \\
   $\psi^{\star}, a^{\star}, b^{\star}$ & Optimal functions of $\psi, a$
and $b$ \\
   $I(\calX,\calU)$ & Mutual information (MI) between $\bm{x}$ and $\bm{u}$\\ 
   $I(\calYx,\calYu)$& MI between $\byx=\bhx(\bm{x})$ and
   $\byu=\bhu(\bm{u})$\\
    $\bhx^{\perp}:\R{\Dx}\to\R{\Dx-\dx}$ & Function such that
       $(\bhx(\bm{x}), \bhx^{\perp}(\bm{x}))$ is invertible.\\
   $\bhu^{\perp}:\R{\Du}\to\R{\Du-\du}$ & Function such that
       $(\bhu(\bm{u}), \bhu^{\perp}(\bm{u}))$ is invertible.\\
   $\bm{s}\in\R{\dx}$ & Vector of latent source components \\
   $\bm{f}:\R{\Dx}\to\R{\Dx}$ & Nonlinear mixing function\\
   $q_i:\R{}\times\R{\Du}\to\R{}$~(Proposition~\ref{theo:general-ICA}) 
    & Exponent of the conditional density $p(s_i|\bm{u})$\\ 
   $q^{\prime}_i, q^{\prime\prime}_i$ & First- and second-order derivatives of $q_i(t,\bm{u})$ wrt $t$\\ 
   $\nabla_{\bm{u}}$ & Gradient wrt $\bm{u}$\\
   $\bm{w}:\R{\Dx}\times\R{\Du}\to\R{2\Dx}$ & 
       $\bm{w}(\bm{v},\bm{u}):=(q^{\prime}_1(v_1,\bm{u}),\dots,q^{\prime\prime}_{\Dx}(v_{\Dx},\bm{u}))^{\top}$ \\
   $q_i:\R{}\times\R{}\to\R{}$~(Theorem~\ref{prop:general-PCL}) & Exponent of 
       the conditional density $p(s_i|\bm{u})$\\ 
   $\alpha^{1}_i, \alpha^{2}_i:\R{}\to\R{}$ & 
       $\alpha_i^{1}(v):=\frac{\partial^2{q}_i(v,r)}{\partial{v}\partial{r}}\Bigr|_{r=\hq{i}(\bu_1)}$
       and
       $\alpha_i^{2}(v):=\frac{\partial^2{q}_i(v,r)}{\partial{v}\partial{r}}\Bigr|_{r=\hq{i}(\bu_2)}$\\
   $\bm{\alpha}:\R{\Dx}\to\R{\Dx}$ & 
       $\bm{\alpha}(\bm{v}):=\left(\frac{\alpha_1^{2}(v_1)}{\alpha_1^{1}(v_{1})},\dots,\frac{\alpha_{\Dx}^{2}(v_{\Dx})}{\alpha_{\Dx}^{1}(v_{\Dx})}\right)^{\top}$\\
    $\bm{n}\in\R{\Dx-\dx}$ & Vector of nuisance variables \\
   \hline
   \end{tabular}
   \begin{center}
    \caption{\label{tab:summary} Summary and comparison of theoretical
    conditions in maximization of MI, nonlinear ICA and nonlinear
    subspace estimation.  Gen. model and source cond. are abbreviations
    of generative model and source condition, respectively. ``-'' means
    that the condition is not required. See the main text for more
    details.}
    \vspace{1mm}
    \begin{tabular}{c|c|c|c}
     \hline & Maximization of MI & Nonlinear ICA & Nonlinear subspace \\
    \hline Form of $\psi^{\star}$ &
    $\psi^{\star}(\bhx^{\star}(\bx),\bhu^{\star}(\bm{u}))$
	 & \begin{tabular}{c}
	    $\sum_{i=1}^{\Dx}\psi^{\star}_i(\hx{i}^{\star}(\bm{x}),\bhu^{\star}(\bm{u}))$ \\
	    or~$\sum_{i=1}^{\Dx}\psi^{\star}_i(\hx{i}^{\star}(\bm{x}),\hu{i}^{\star}(\bm{u}))$
	   \end{tabular}
     & $\psi^{\star}(\bhx^{\star}(\bx),\bhu^{\star}(\bm{u}))$\\
     \hline Dim. assump. & $\dx\leq\Dx$ and $\du\leq\Du$ 
     & \begin{tabular}{c}
	$\dx=\Dx$ and  \\ $2\Dx\leq\Du$ or $\Dx\leq\Du$ 
	\end{tabular} 
	     & $\dx\leq\Dx$ and $\dx\leq\du$ \\
     \hline Gen. model & - & $\bm{x}=\bm{f}(\bm{s})$
	     & $\bm{x}=\bm{f}(\bm{s},\bm{n})$\\
     \hline Source cond. & - & $s_i\perp{s}_j|\bm{u}~(i\neq{j})$ &
		 $\bm{s},\bm{u}\perp\bm{n}$ and $\bm{s}\not\perp\bm{u}$\\
     \hline
       \end{tabular}
   \end{center}
  \end{table}
  \subsection{Maximization of mutual information}
  \label{ssec:maxMI}
  Maximization of mutual information
  (MI)~\citep{barlow1961possible,linsker1989application,bell1995information}
  is a classical yet recently retrieved framework for unsupervised
  representation learning combined with the recent development of deep
  neural
  networks~\citep{hjelm2019learning,tschannen2019mutual}.
  Density ratio estimation seems not to be strongly related, but our
  analysis implies that it is essential for maximization of MI.
  
  Let us re-denote the representation functions of $\bm{x}$ and $\bm{u}$
  by
  \begin{align*}
   \byx:=\bhx(\bm{x})\quad \text{and}\quad \byu:=\bhu(\bm{u}),
  \end{align*}
  respectively. The goal of maximization of MI is to find $\bhx$ and
  $\bhu$, which maximize MI between $\byx$ and $\byu$ defined by
  \begin{align*}
   I(\calYx,\calYu):=\int
   p(\byx,\byu)\log\frac{p(\byx,\byu)}{p(\byx)p(\byu)}\intd\byx\intd\byu.
  \end{align*}  
  Data processing inequality shows that $I(\calYx,\calYu)$ is a lower
  bound of MI between $\bm{x}$ and $\bm{u}$, i.e.,
  \begin{align}
   I(\calX,\calU)\geq I(\calYx,\calYu). \label{ieq-MI-H}
  \end{align}
  Inequality~\eqref{ieq-MI-H} indicates that any reparametrizations of
  $\bm{x}$ and $\bm{u}$ never exceed $I(\calX,\calU)$, and
  $I(\calYx,\calYu)$ is maximized at $I(\calX,\calU)$ if there exist
  such representation functions $\bhx$ and $\bhu$.
  
  The following theorem proved in Appendix~\ref{app:maxMIproof}
  clarifies how and when contrastive learning and variational MI
  estimation can be regarded as performing maximization of MI, and
  establishes conditions on which $I(\calYx,\calYu)=I(\calX,\calU)$,
  i.e., $I(\calYx,\calYu)$ is maximized:
  \begin{theorem}
   \label{prop:maxMI} We make the following assumptions:
   \begin{enumerate}[(A1)]
    \item[(A1)] $p(\bm{x},\bu)>0$, $p(\bm{x})>0$ and $p(\bm{u})>0$.
	  
    \item[(A2)] There exist some functions
		$\bhx^{\perp}:\R{\Dx}\to\R{\Dx-\dx}$ and
		$\bhu^{\perp}:\R{\Du}\to\R{\Du-\du}$ such that
		\begin{align*}
		 \left(
		 \begin{array}{c}
		  \bhx(\bx)\\ \bhx^{\perp}(\bx)
		 \end{array}\right)\in\R{\Dx}
		 \quad\text{and}\quad
		 \left(
		  \begin{array}{c}
		   \bhu(\bu)\\ \bhu^{\perp}(\bu)
		  \end{array}\right)\in\R{\Du}
		\end{align*}
		are both invertible\footnote{Invertibility means that
		there exist $\bm{g}_{\mathrm{x}}$ and
		$\bm{g}_{\mathrm{u}}$ such that
		$\bx=\bm{g}_{\mathrm{x}}(\byx,\byx^{\perp})$ and
		$\bu=\bm{g}_{\mathrm{u}}(\byu,\byu^{\perp})$ where
		$\byx^{\perp}:=\bhx^{\perp}(\bm{x})$ and
		$\byu^{\perp}:=\bhu^{\perp}(\bm{u})$.}.

    \item[(A3)] There exist functions $\psi^{\star}$, $\bhx^{\star}$,
		$\bhu^{\star}$, $a^{\star}$ and $b^{\star}$ such that
		the following equation holds:
		\begin{align}
		 \log\frac{p(\bm{x},\bm{u})}{p(\bm{x})p(\bm{u})}
		 =\psi^{\star}(\bhx^{\star}(\bx),\bhu^{\star}(\bm{u}))
		 +a^{\star}(\bhx^{\star}(\bx))+b^{\star}(\bhu^{\star}(\bu)).
		 \label{univ-approx-MI}
		\end{align}
   \end{enumerate}
   Then, $I(\calYx,\calYu)=I(\calX,\calU)$ at $\bhx=\bhx^{\star}$, and
   $\bhu=\bhu^{\star}$.  Conversely, suppose that
   $I(\calYx,\calYu)=I(\calX,\calU)$ at $\bhx=\bhx^{\star}$ and
   $\bhu=\bhu^{\star}$ under Assumptions~(A1-2). Then, there exist
   functions $\psi^{\star}$, $a^{\star}$ and $b^{\star}$ such that
   \eqref{univ-approx-MI} holds.
  \end{theorem}    
  Theorem~\ref{prop:maxMI} implies that $I(\calYx,\calYu)$ can be
  maximized through density ratio estimation, and thus motivates us to
  develop practical methods for maximization of MI through density ratio
  estimation. Eq.\eqref{univ-approx-MI} is inspired by \emph{sufficient
  dimension
  reduction}~\citep{li1991sliced,cook1998regression,fukumizu2004dimensionality},
  which is a solid framework for \emph{supervised} dimensionality
  reduction and whose goal is to find an informative lower-dimensional
  subspace to the output variable based on the conditional independence
  condition. As shown in the proof of Theorem~\ref{prop:maxMI},
  \eqref{univ-approx-MI} can be rewritten as the following conditional
  independence conditions:
  \begin{align*}
   \bm{u}\perp\bm{x}~|~\byx=\bhx^{\star}(\bm{x})\quad\text{and}\quad
   \bm{x}\perp\bm{u}~|~\byu=\bhu^{\star}(\bm{u}).
  \end{align*}
  The conditional independence between $\bm{u}$ and $\bm{x}$ given
  $\byx=\bhx^{\star}(\bm{x})$ implies that the lower-dimensional
  representation $\bhx^{\star}(\bm{x})$ has the same amount of
  information for $\bm{u}$ as the original input data $\bm{x}$. The same
  implication holds the conditional independence between $\bm{u}$ and
  $\bm{x}$ given $\byu=\bhu^{\star}(\bm{u})$ as well. Thus, accurately
  estimating the density ratio would yield representations of $\bm{u}$
  and $\bm{x}$ possibly with minimum information loss.

  An interesting point of Theorem~\ref{prop:maxMI} is that
  $I(\calYx,\calYu)=I(\calX,\calU)$ conversely implies the density ratio
  equation~\eqref{univ-approx-MI}, and is useful for understanding when
  contrastive learning does not perform maximization of MI.  In
  Section~\ref{ssec:CL-MI}, we suppose that $\bm{u}^*(t)$ are drawn from
  the marginal density $p(\bu)$ in~\eqref{binary-classification}, but in
  practice, $\bm{u}^*(t)$ are often taken from another
  dataset~\citep{arandjelovic2017look}\footnote{For instance,
  in~\citet{arandjelovic2017look}, the positive pairs of input and
  complimentary data samples are video frames and audio clips in the
  same video that overlap in time, while the negative pairs are randomly
  extracted from two different videos. Thus, the marginal density of the
  complimentary data (i.e., video frame or audio clip) can be different
  in the positive and negative data.} whose probability density is
  different from the marginal density $p(\bm{u})$ and denoted by
  $\tilde{p}(\bu)$. Then, by assuming that the $\bm{u}^*$ are
  independent to $\bx$, contrastive learning related to density ratio
  estimation yields an estimate of
  \begin{align*}
   \log\frac{p(\bx,\bu)}{p(\bx)\tilde{p}(\bu)}
  \end{align*}
  up to a constant, and \eqref{univ-approx-MI} is never fulfilled. Thus,
  contrastive learning based on another marginal density
  $\tilde{p}(\bu)$ might not be regarded as maximizing
  $I(\calYx,\calYu)$ in general. On the other hand, as long as
  $\bm{u}^*(t)$ are drawn form the marginal density $p(\bu)$,
  Theorem~\ref{prop:maxMI} would be a direct support to the belief that
  contrastive learning and variational MI estimation can be considered
  to perform maximization of MI because the popular objective functions
  are related to density ratio estimation as reviewed in
  Section~\ref{ssec:CL-MI}.

  Another simple practical point of Theorem~\ref{prop:maxMI} is the form
  of the right-hand side on~\eqref{univ-approx-MI}: The right-hand side
  indicates that all terms have to be functions of $\bhu^{\star}$ and/or
  $\bhx^{\star}$. Thus, even when $a^{\star}(\bhx^{\star}(\bx))$ is
  simply replaced with $a(\bx)$ (i.e., $a$ is not a function of
  $\bhx^{\star}$), Theorem~\ref{prop:maxMI} does not hold.
  \subsection{Nonlinear ICA: A new insight for source recovery}
  \label{ssec:nonlinear-ICA}
  Here, we perform two theoretical analyses for source recovery in
  nonlinear ICA with density ratio estimation
  (Proposition~\ref{theo:general-ICA} and
  Theorem~\ref{prop:general-PCL}). These analyses shed light on a novel
  insight that the dimensionality of complementary data is an important
  factor for source recovery, which has not been revealed in previous
  work of nonlinear ICA. Furthermore, Theorem~\ref{prop:general-PCL} can
  be regarded as a generalization of Theorem~1
  in~\citet{pmlr-v54-hyvarinen17a}.
  \paragraph{Importance of the dimensionality of complementary data:}
  As reviewed in Section~\ref{ssec:rev-nonlinear-ICA}, nonlinear ICA
  assumes that input data $\bm{x}=(x_1,x_2,\dots,x_{\Dx})^{\top}$ is
  generated as a nonlinear mixing of the latent source components $s_1,
  s_2, \dots, s_{\Dx}$ as follows:
  \begin{align*}
   \bm{x}=\bm{f}(\bm{s}),
  \end{align*}
  where $\bm{s}=(s_1,s_2,\dots,s_{\Dx})^{\top}$ and
  $\bm{f}:\R{\Dx}\to\R{\Dx}$ is assumed to be invertible.  Then, the
  goal is to recover the latent source components $s_i$.  To this end,
  recent work of nonlinear ICA employs complementary data $\bm{u}(t)$ in
  addition to input data $\bm{x}(t)$. For instance, by regarding
  $\bx(t)$ as time series data at the time index $t$,
  \citet{pmlr-v54-hyvarinen17a} use past input data as $\bm{u}(t)$
  (e.g., $\bm{u}(t)=\bm{x}(t-1)$).

  We first establish the following theorem showing that the latent
  source components can be recovered up to their permutation (i.e.,
  ordering) and elementwise invertible functions, and that density ratio
  estimation plays an important role in nonlinear ICA as well:
  \begin{proposition}
   \label{theo:general-ICA} Suppose that $\dx=\Dx$. We further make the
   following assumptions:
   \begin{enumerate}[({B}1)]
    \item The latent source components $s_i$ are conditionally
	  independent given $\bm{u}$. More specifically, the conditional
	  density of $\bm{s}$ given $\bm{u}$ takes the following form:
	  \begin{align*}
	   \log p(\bm{s}|\bm{u})=\sum_{i=1}^{\Dx} q_i(s_i,\bm{u})
	  -\log{Z}(\bu),
	  \end{align*}
	  where ${Z}(\bu)$ denotes the partition function and $q_i$ are
	  differentiable functions.
	    
    \item Input data $\bx$ is generated according to~\eqref{ICA-model}
	  where the mixing function $\bm{f}$ is invertible.

    \item Dimensionality of complementary data is twice larger than or
	  twice as large as input data, i.e., $2\Dx\leq\Du$.
	  
    \item There exists a single point $\bu_1$ such that the rank of
	  $\nabla_{\bm{u}}\bm{w}(\bm{v},\bm{u})\in\R{2\Dx\times\Du}$ at
	  $\bu=\bu_1$ is $2\Dx$ for all
	  $\bm{v}=(v_1,v_2,\dots,v_{\Dx})^{\top}$ where
	  $\nabla_{\bm{u}}$ denotes the differential operator with
	  respect to $\bu$, and
	  \begin{align*}
	   \bm{w}(\bm{v},\bm{u}):=\left(q^{\prime}_1(v_1,\bm{u})
	   ,\dots, q^{\prime}_{\Dx}(v_{\Dx},\bm{u}),
	   q^{\prime\prime}_1(v_1,\bm{u}),\dots,
	   q^{\prime\prime}_{\Dx}(v_{\Dx},\bm{u})\right)^{\top}\in\R{2\Dx},
	  \end{align*}
	  with $q^{\prime}_i(t,\bm{u}):=\parder{}{t}q_i(t,\bm{u})$ and
	  $q^{\prime\prime}_i(t,\bm{u}):=\frac{\partial^2}{\partial t^2}
	  q_i(t,\bm{u})$.

    \item There exist functions $\psi^{\star}_i$, $\bhx^{\star}$,
	  $\bhu^{\star}$, $a^{\star}$ and $b^{\star}$ such that the
	  following equation holds:
	  \begin{align}
	   \log\frac{p(\bm{x},\bm{u})}{p(\bm{x})p(\bm{u})} =
	   \sum_{i=1}^{\Dx}[\psi^{\star}_i(\hx{i}^{\star}(\bm{x}),\bhu^{\star}(\bm{u}))]
	   +a^{\star}(\bhx^{\star}(\bx))+b^{\star}(\bhu^{\star}(\bu)),
	   \label{ICA-universal}
	  \end{align}
	  where $\psi_i^{\star}$ are differentiable functions and
	  $\bhx^{\star}(\bm{x}):=(\hx{1}^{\star}(\bm{x}),\dots,\hx{\Dx}^{\star}(\bm{x}))^\top$
	  is invertible.
   \end{enumerate}
   Then, under Assumptions~(B1-5), the representation function
   $\bhx^{\star}(\bm{x})$ is equal to $\bm{s}$ up to a permutation and
   elementwise invertible functions.
  \end{proposition}
  The proof is given in Appendix~\ref{app:general-ICA}. We essentially
  followed the proof of Theorem~1 in~\citet{hyvarinen2018nonlinear} and
  derived the same conclusion. Here, the main difference is
  Assumptions~(B3-4), which give a new insight for source
  recovery. \citet{hyvarinen2018nonlinear} adopted an alternative
  assumption to Assumptions~(B3-4) called the \emph{assumption of
  variability}, which assumes that there exist $2\Dx+1$ points,
  $\bm{u}_0,\bm{u}_1,\dots,\bm{u}_{2\Dx}$, such that the following
  $2\Dx$ vectors are linearly independent:
  \begin{center}
   $\bm{w}(\bm{v},\bm{u}_j)-\bm{w}(\bm{v},\bm{u}_0)$ for
   $j=1,\dots,2\Dx$ are linearly independent for all $\bm{v}$.
  \end{center}          
  The dimensionality assumption $2\Dx\leq\Du$ in Assumption~(B3) would
  correspond to the $2\Dx$ vectors in the assumption of variability,
  but, in contrast, sheds light on a novel insight: The dimensionality
  of complementary data is an important factor for source recovery,
  which has not been revealed in previous work of nonlinear ICA. In
  fact, we numerically demonstrate that the accuracy of source recovery
  clearly depends on the dimensionality of complementary data in
  Section~\ref{sssec:high-aug-var}. Furthermore, Assumption~(B3) is
  practically useful because it can be checked very easily. Regarding
  Assumption~(B4), it implies that the complementary data $\bm{u}$ is
  clearly dependent to $\bm{s}$: When the $i$-th element in $\bm{u}$ is
  independent to $\bm{s}$ and $2\Dx=\Du$ (i.e.,
  $\nabla_{\bm{u}}\bm{w}(\bm{v},\bm{u})$ is a square matrix), the $i$-th
  column in $\nabla_{\bm{u}}\bm{w}(\bm{v},\bm{u})$ is the zero vector,
  and thus the rank assumption in Assumption~(B4) is never satisfied.
  
  According to~\citet{hyvarinen2018nonlinear}, the assumption of
  variability implies that the underlying conditional density
  $p(\bm{s}|\bm{u})$ has to be diverse and complex to recover the source
  components. As in Theorem~2 in~\citet{hyvarinen2018nonlinear}, we show
  that Assumption~(B4) includes the same implication under the following
  exponential family:
  \begin{align}
   \log p(\bm{s}|\bm{u})=\sum_{i=1}^{\Dx}\sum_{k=1}^K
   \lambda_{ik}(\bm{u})q_{ik}(s_i) -\log Z(\bu), \label{exp-family}
  \end{align}
  where $\lambda_{ik}$ and $q_{ik}$ are some scalar functions and $K$ is
  a positive integer.  Appendix~\ref{app:rank-inequality} proves that
  \begin{itemize}
   \item When $K=1$,
	 $\rank(\nabla_{\bm{u}}\bm{w}(\bm{v},\bm{u}))\leq\Dx$ where
	 $\rank(\cdot)$ denotes the rank of a matrix.

   \item When $K>1$,
	 $\rank(\nabla_{\bm{u}}\bm{w}(\bm{v},\bm{u}))\leq2\Dx$
  \end{itemize}
  Thus, when $K=1$, the $2\Dx$ rank assumption in Assumption~(B4) is
  never fulfilled. This implies that in order to recover the source
  components, the conditional density has to be diverse and complex such
  as the exponential family~\eqref{exp-family} with a relatively large
  mixture number $K$.
  
  \paragraph{A milder dimensionality assumption:}
  The dimensionality assumption $2\Dx\leq\Du$ in Assumption~(B3) might
  be strong because we need to have relatively high-dimensional
  complementary data. However, this assumption can be relaxed by
  restricting the underlying conditional density of $\bm{s}$ given
  $\bu$. The following theorem is based on a milder dimensionality
  assumption than Assumption~(B3), and can be seen as a generalization
  of Theorem~1 in~\citet{pmlr-v54-hyvarinen17a}:
  \begin{theorem}
   \label{prop:general-PCL} Suppose that $\dx=\du=\Dx$. We make the
   following assumptions:
   \begin{enumerate}[({B$'$}1)]
    \item The latent source components $s_i$ are conditionally
	  independent given $\bm{u}$, and the conditional density of
	  $\bm{s}$ given $\bm{u}$ takes the following form:
	  \begin{align}
	   \log p(\bm{s}|\bm{u}) =\sum_{i=1}^{\Dx}q_i(s_i,
	   \lambda_{i}(\bu))-\log{Z}(\bu), \label{cond-dist-general-PCL}
	  \end{align}
	  where $q_i$ and $\lambda_{i}$ are differentiable functions for
	  $i=1,\dots,\Dx$.
	  
    \item Input Data $\bx$ is generated according to~\eqref{ICA-model}
	  where the mixing function $\bm{f}$ is invertible.
	  
    \item Dimensionality of complementary data is larger than or equal
	  to input data, i.e., $\Dx\leq\Du$.
	  
    \item There exist two points, $\bu_1$ and $\bu_2$, such that
	  $\alpha_i^{1}(v):=\frac{\partial^2{q}_i(v,r)}{\partial{v}\partial{r}}\Bigr|_{r=\hq{i}(\bu_1)}\neq0$
	  and
	  $\alpha_i^{2}(v):=\frac{\partial^2{q}_i(v,r)}{\partial{v}\partial{r}}\Bigr|_{r=\hq{i}(\bu_2)}\neq{0}$
	  for all $i$ and $v$.

    \item There exist $\Dx$ points,
	  $\bm{v}_1,\bm{v}_2,\dots,\bm{v}_{\Dx}$, such that the $\Dx$
	  vectors $\bm{\alpha}(\bm{v}_1)$, $\bm{\alpha}(\bm{v}_2)$,
	  \dots, $\bm{\alpha}(\bm{v}_{\Dx})$ are linearly independent
	  where with $\bm{v}:=(v_1,v_2,\dots,v_{\Dx})^{\top}$,
	  \begin{align*}
	   \bm{\alpha}(\bm{v}):=\left(\frac{\alpha_1^{2}(v_1)}{\alpha_1^{1}(v_1)},
	   \frac{\alpha_2^{2}(v_2)}{\alpha_2^{1}(v_2)},\dots,\frac{\alpha_{\Dx}^{2}(v_{\Dx})}{\alpha_{\Dx}^{1}(v_{\Dx})}\right)^{\top}\in\R{\Dx}.
	  \end{align*}

    \item There exist functions $\psi^{\star}_i$, $\bhx^{\star}$,
	  $\bhu^{\star}$, $a^{\star}$ and $b^{\star}$ such that the
	  following equation holds:
	  \begin{align}
	   \log \frac{p(\bm{x},\bm{u})}{p(\bm{x})p(\bm{u})} =
	   \sum_{i=1}^{\Dx}
	   [\psi^{\star}_i(\hx{i}^{\star}(\bm{x}),\hu{i}^{\star}(\bm{u}))]
	   +a^{\star}(\bhx^{\star}(\bx))+b^{\star}(\bhu^{\star}(\bu)), 
	   \label{ICA-universal2}
	  \end{align}
	  where $\psi_i^{\star}$ are differentiable functions,
	  $\bhx^{\star}(\bx):=(\hx{1}^{\star}(\bx),\dots,\hx{\Dx}^{\star}(\bx))^\top$
	  is invertible, and
	  $\bhu^{\star}(\bm{x}):=(\hu{1}^{\star}(\bu),\dots,\hu{\Dx}^{\star}(\bu))^\top$.

   \end{enumerate}
   Then, under Assumptions~(B$'$1-6), the representation function
   $\bhx^{\star}(\bm{x})$ is equal to $\bm{s}$ up to a permutation and
   elementwise invertible functions.
  \end{theorem}
  The proof is given in Appendix~\ref{app:proof-general-PCL}. The key
  point in Theorem~\ref{prop:general-PCL} is that the dimensionality
  assumption $\Dx\leq\Du$~in Assumption~(B$'$3) is milder than
  $2\Dx\leq\Du$ in Assumption (B3). However, this milder assumption
  comes at a cost of making the conditional density $p(\bm{s}|\bm{u})$
  less general than Proposition~\ref{theo:general-ICA} and of
  restricting the form of the right-hand side on~\eqref{ICA-universal2}:
  The conditional density $p(\bm{s}|\bm{u})$ is restricted into a form
  of pairwise combinations of $s_i$ and $\lambda_{i}(\bu)$ for
  $i=1\dots,\Dx$ in~\eqref{cond-dist-general-PCL}, and the right-hand
  side on~\eqref{ICA-universal2} also takes a pairwise form of
  $\hx{i}^{\star}(\bm{x})$ and $\hu{i}^{\star}(\bm{u})$.  
  
  Similarly as Assumption~(B4) in Proposition~\ref{theo:general-ICA},
  Assumption~(B$'$5) also implies that the underlying conditional
  density~\eqref{cond-dist-general-PCL} is diverse and complex. Indeed,
  when $p(\bm{s}|\bu)$ belongs to the exponential
  family~\eqref{exp-family} in $K=1$, the ratio
  $\frac{\alpha_i^{2}(v_i)}{\alpha_i^{1}(v_i)}$ is equal to a constant
  $\frac{\lambda_{i1}(\bu_2)}{\lambda_{i1}(\bu_1)}$ for all $i$. Then,
  $\bm{\alpha}(\bm{v})$ is a constant vector, and cannot be linearly
  independent over the $\Dx$ points. Thus, Assumption~(B$'$5) is never
  satisfied under the exponential family~\eqref{exp-family} in $K=1$.

  Theorem~\ref{prop:general-PCL} generalizes Theorem~1
  in~\citet{pmlr-v54-hyvarinen17a}: By regarding $\bm{x}(t),
  t=1,\dots,T$ as time series data at the time index $t$, Theorem~1
  in~\citet{pmlr-v54-hyvarinen17a} is a special case of
  Theorem~\ref{prop:general-PCL} where $\bm{u}(t)=\bm{x}(t-1)$ (thus,
  implicitly suppose $\Dx=\Du$ and Assumption~(B$'$3) is satisfied),
  $\bhq(\bu)=\bu$, and $\bhu=\bhx$ (e.g., the same neural architecture
  with weight sharing). This generalization is not straightforward
  because we had to derive a new lemma (Lemma~\ref{lem:diagonals}), and
  thus the proof is substantially different. Furthermore, by this
  generalization, Theorem~\ref{prop:general-PCL}, again, reveals that
  the dimensionality of complementary data is an important factor, which
  has not been seen in Theorem~1 of~\citet{pmlr-v54-hyvarinen17a}.

  Finally, we note a subtle difference of
  $\psi(\bhx(\bm{x}),\bhu(\bm{u}))$ in nonlinear ICA. Nonlinear ICA
  requires us to slightly modify $\psi(\bhx(\bm{x}),\bhu(\bm{u}))$ as
  follows: For Proposition~\ref{theo:general-ICA},
  \begin{align}
   \psi(\bhx(\bm{x}),\bhu(\bm{u}))
   =\sum_{i=1}^{\Dx}\psi_i(\hx{i}(\bm{x}),\bhu(\bm{u})),
   \label{prop-ICA}
  \end{align}
  while in Theorem~\ref{prop:general-PCL},
  \begin{align}
   \psi(\bhx(\bm{x}),\bhu(\bm{u}))
   =\sum_{i=1}^{\Dx}[\psi_i(\hx{i}(\bm{x}),\hu{i}(\bm{u}))].
   \label{theo-ICA}
  \end{align}
  In contrast with maximization of mutual information
  (Theorem~\ref{prop:maxMI}) as well as nonlinear subspace estimation
  (Theorem~\ref{theo:manifold}), $\psi(\bhx(\bm{x}),\bhu(\bm{u}))$
  in~\eqref{prop-ICA} (or in~\eqref{theo-ICA}) is expressed as the sum
  of elementwise functions with respect to $\bhx(\bm{x})$ (or
  $\bhx(\bm{x})$ and $\bhu(\bm{u})$). Thus, in practice, we need to
  slightly modify the form of $\psi(\bhx(\bm{x}),\bhu(\bm{u}))$
  to~\eqref{prop-ICA} or~\eqref{theo-ICA} when performing nonlinear ICA.
  \subsection{Nonlinear subspace estimation with complementary data}
  Here, we first propose a new generative model where input data is
  generated as a nonlinear mixing of lower-dimensional latent source
  components and nuisance variables. Then, we establish some theoretical
  conditions to estimate a lower-dimensional nonlinear subspace of the
  latent source components only.
    \paragraph{A new generative model:} 
    Let us consider the following novel generative model for input data
    $\bm{x}\in\R{\Dx}$:
    \begin{align}
     \bm{x}=\bm{f}(\bm{s},\bm{n}), \label{generative-model2}
    \end{align}
    where $\bm{f}$ is an invertible nonlinear mixing function,
    $\bm{s}\in\R{\dx}$ and $\bm{n}\in\R{\Dx-\dx}$. We further assume
    that
    \begin{itemize}
     \item Source components $s_i$ are statistically independent to
	   nuisance variables $\bm{n}$, i.e., $\bm{s}\perp\bm{n}$

     \item Complementary data is supposed to be available such that
	   $\bm{u}\perp\bm{n}$, while $\bm{u}$ depends on $\bm{s}$,
	   i.e., $\bm{s}\not\perp\bm{u}$ where $\not\perp$ denotes
	   statistical dependence.
    \end{itemize}
    Based on the observations of $\bm{x}$ and $\bu$, the goal is to
    estimate a lower-dimensional nonlinear subspace of only $\bm{s}$,
    which is separated from $\bm{n}$. In contrast with nonlinear ICA, we
    do not necessarily assume that $s_i$ are conditionally independent
    given $\bm{u}$. Thus, \eqref{generative-model2} is more general than
    the generative model~\eqref{ICA-model} in nonlinear ICA in the sense
    that the conditional independence of $s_i$ given $\bu$ is no longer
    assumed.
    
    The new generative model~\eqref{generative-model2} can be motivated
    by many practical situations. For example, time-series data such as
    brain signals~\citep{dornhege2007toward} could be observed as a
    mixture of stationary and nonstationary components, and the
    nonstationarity could be important for a range of tasks such as
    change detection~\citep{blythe2012feature}.  For this example, by
    using the past data in time series data as complementary data (e.g.,
    $\bu(t)=\bx(t-1)$), we may estimate a subspace of the useful
    nonstationary components. Another example is image data.  The pixel
    values around the center of an image usually depends on surrounding
    pixels, while the pixels around the corners of the image are often
    almost independent to the other pixels (e.g., MNIST images). By
    using surrounding pixels as the complementary data, it would be very
    informative to estimate some lower-dimensional subspace for image
    data, which constitutes the fundamental part of image data.
    \paragraph{Estimating a lower-dimensional subspace of the source components:}
    First of all, it is important to understand whether we can estimate
    a lower-dimensional subspace of the latent sources $\bm{s}$, which
    is separated from nuisance variables $\bm{n}$. The following theorem
    gives conditions to estimate such a subspace as a (vector-valued)
    function of $\bm{s}$ only:
    \begin{theorem}
     \label{theo:manifold} Assume that
     \begin{enumerate}[(C1)]
      \item[(C1)] Input data $\bm{x}$ is generated according
		  to~\eqref{generative-model2} where the mixing function
		  $\bm{f}$ is invertible, $\bm{s}\perp\bm{n}$,
		  $\bm{s}\not\perp\bm{u}$ and $\bu\perp\bm{n}$.

      \item[(C2)] There exist functions $\psi^{\star}$, $\bhx^{\star}$,
		  $\bhu^{\star}$, $a^{\star}$ and $b^{\star}$ such that
		  the following equation holds:
		  \begin{align}
		   \log\frac{p(\bm{x},\bm{u})}{p(\bm{x})p(\bm{u})}
		   =\psi^{\star}(\bhx^{\star}(\bm{x}),\bhu^{\star}(\bm{u}))
		   +a^{\star}(\bhx^{\star}(\bx))+b^{\star}(\bhu^{\star}(\bu)),
		   \label{manifold-universal}
		  \end{align}		
		  where $\bhx^{\star}$ is surjective.
		  	    
      \item[(C3)] Dimensionality of $\bhx^{\star}(\bx)$ is smaller
		  than or equal to $\bhu^{\star}(\bu)$, i.e.,
		  $\dx\leq\du$.
		  		  
      \item[(C4)] There exists at least a single point $\bu_1$ such
		  that the rank of
		  $\nabla_{\bm{v}}\nabla_{\bm{u}}\psi^{\star}(\bm{v},\bhu^{\star}(\bu))\in\R{\Du\times\dx}$
		  is $\dx$ at $\bu=\bu_1$ for all $\bm{v}\in\R{\dx}$
		  where $\bm{v}:=\bhx^{\star}(\bx)$.
     \end{enumerate}     
     Under Assumptions~(C1-4), the representation function
     $\bhx^{\star}(\bx)$ is a nonlinear function of only $\bm{s}$.
    \end{theorem}            
    The proof is given in Appendix~\ref{app:manifold}. The interesting
    point of Theorem~\ref{theo:manifold} is that complementary data
    $\bm{u}$ enables us to automatically ignore the nuisance variables
    $\bm{n}$, and Theorem~\ref{theo:manifold} implies that a
    lower-dimensional nonlinear subspace only for $\bm{s}$ can be
    estimated through density ratio estimation. Assumption~(C3) is a
    necessary condition that the rank of
    $\nabla_{\bm{v}}\nabla_{\bm{u}}\psi^{\star}(\bm{v},\bhu^{\star}(\bu))$
    is $\dx$: By defining $\bm{r}:=\bhu^{\star}(\bu)$,
    \begin{align*}
     \rank(\nabla_{\bm{v}}\nabla_{\bm{u}}\psi^{\star}(\bm{v},\bhu^{\star}(\bu)))
     =\rank(\bJ_{\bm{r}}(\bu)^{\top}
     [\nabla_{\bm{v}}\nabla_{\bm{r}}\psi^{\star}(\bm{v},\bm{r})])
     \leq\min(\dx,\du),
    \end{align*}
    where $\bJ_{\bm{r}}(\bu):=\nabla_{\bu}\bm{r}\in\R{\du\times\Du}$ is
    the Jacobian of $\bm{r}$ at $\bu$ and
    $\nabla_{\bm{v}}\nabla_{\bm{r}}\psi^{\star}(\bm{v},\bm{r})\in\R{\du\times\dx}$.
    Thus, it must be $\dx\leq\du$ to satisfy Assumption~(C4).

    In order to understand the implication of Assumption~(C4), we derive
    the following equation from~\eqref{manifold-universal} in
    Appendix~\ref{app:two-grads}:
    \begin{align}
     \bJ_{\bm{v}}(\bm{s})[\nabla_{\bm{s}}\nabla_{\bu}\log
     p(\bm{s}|\bu)\bigr|_{\bm{u}=\bm{u}_1}]
     =\nabla_{\bm{v}}\nabla_{\bu}\psi^\star(\bm{v},\bu)\bigr|_{\bm{u}=\bm{u}_1},
    \label{two-grads}
    \end{align}
    where
    $\bJ_{\bm{v}}(\bm{s}):=\nabla_{\bm{s}}\bm{v}\in\R{\dx\times\dx}$ is
    the Jacobian of $\bm{v}=\bhx^{\star}(\bm{f}(\bm{s},\bm{n}))$ at
    $\bm{s}$. According to Lemma~\ref{lem:rank} and Assumption~(C4),
    \eqref{two-grads} ensures that the rank of
    $\nabla_{\bm{s}}\nabla_{\bu}\log
    p(\bm{s}|\bu)\bigr|_{\bm{u}=\bm{u}_1}$ is $\dx$.  Next, as discussed
    in nonlinear ICA (Section~\ref{ssec:nonlinear-ICA}), we investigate
    whether or not the rank of $\nabla_{\bm{s}}\nabla_{\bu}\log
    p(\bm{s}|\bu)\bigr|_{\bu=\bu_1}$ is $\dx$ under the following
    exponential family:
    \begin{align}
     \log p(\bm{s}|\bm{u})=\sum_{j=1}^{D}\lambda_{j}(\bm{u})q_j(\bm{s})
     -\log Z(\bu), \label{exp-family2}
    \end{align}
    where $\lambda_{j}$ and $q_j$ are some scalar functions.  Then, we
    have
    \begin{align*}
     \nabla_{\bm{s}}\nabla_{\bm{u}}\log p(\bm{s}|\bm{u}) =\sum_{j=1}^{D}
     \left\{\nabla_{\bm{u}}\lambda_{j}(\bm{u})\right\}
     \left\{\nabla_{\bm{s}}q_j(\bm{s})\right\}^\top,
    \end{align*}   
    which indicates that by the rank factorization theorem,
    \begin{align*}
     \rank(\nabla_{\bm{s}}\nabla_{\bm{u}}\log p(\bm{s}|\bm{u})) 
     \leq\min(\dx,D)
    \end{align*}
    where we used $\dx\leq\du\leq\Du$ from Assumption~(C3).  Thus, when
    $D<\dx$, Assumption~(C4) is not fulfilled.  As in nonlinear ICA,
    Assumption~(C4) possibly implies that in order to estimate a
    nonlinear subspace of $\bm{s}$, $p(\bm{s}|\bm{u})$ has to be
    sufficiently complex such as the exponential
    family~\eqref{exp-family2} with relatively large $D$ ($>\dx$).

    Compared with nonlinear ICA in Section~\ref{ssec:nonlinear-ICA},
    Theorem~\ref{theo:manifold} stands on a more general setting in the
    sense that the elements in $\bm{s}$ are not necessarily assumed to
    be conditionally independent. Furthermore, the dimensionality
    condition $\dx\leq\du$ in Assumption~(C3) is milder than
    Assumption~(B3) in Proposition~\ref{theo:general-ICA} and
    Assumption~(B$'$3) in Theorem~\ref{prop:general-PCL} because $\dx$
    and $\du$ are the dimensionalities of the representation functions
    $\bhx^{\star}$ and $\bhu^{\star}$ and assumed to be smaller than
    $\Dx$ and $\Du$, respectively.  However, these milder conditions
    come at a price for losing the recover of each latent source
    component $s_i$: Nonlinear ICA recovers each component $s_i$ up to a
    permutation and elementwise invertible  function, while
    Theorem~\ref{theo:manifold} only guarantees that a nonlinear
    function of $\bm{s}$ can be estimated.
    
    A similar generative model as~\eqref{generative-model2} was proposed
    in multi-view learning~\citep{gresele2019incomplete}, which
    considers that two data, $\bm{x}^{(1)}$ and $\bm{x}^{(2)}$, are
    generated as
    \begin{align}
     \bm{x}^{(1)}&=\bm{f}^{(1)}(\bm{s})\quad\text{and}\quad
     \bm{x}^{(2)}=\bm{f}^{(2)}(\bm{g}(\bm{s},\bm{n})),
     \label{generative-model3}
    \end{align}
    where $\bm{f}^{(1)}$ and $\bm{f}^{(2)}$ are the mixing functions,
    and $\bm{g}$ denotes an elementwise vector-valued function. The generative model~\eqref{generative-model3}
    assumes that $\bm{s}\perp\bm{n}$, $s_i\perp{s}_j$ and
    $n_i\perp{n}_j$ for $i\neq{j}$ where $n_i$ denotes the $i$-th
    element in $\bm{n}$.  Then, \citet{gresele2019incomplete} proved
    that each component $s_i$ can be recovered up to elementwise
    invertible functions. In contrast, the assumptions in our generative
    model~\eqref{generative-model2} can be more general because it is
    not assumed that the elements both in $\bm{s}$ and $\bm{n}$ are
    independent, i.e., $s_i\not\perp{s}_j$ and $n_i\not\perp{n}_j$ for
    $i\neq{j}$ in our generative model.
 \section{Estimation methods}
 \label{sec:estimation-methods}
 Section~\ref{sec:unified-view} theoretically showed that density ratio
 estimation plays a key role in the three frameworks, and motivates us
 to develop practical methods for unsupervised representation learning
 through density ratio estimation. This section proposes two practical
 methods to estimate the log-density ratio based on neural networks.
 The first method employs the $\gamma$-cross
 entropy~\citep{fujisawa2008robust}, which is a robust variant of the
 standard cross entropy against outliers. For nonlinear ICA, the second
 one employs the Donsker-Varadhan variational estimation for mutual
 information~\citep{ruderman2012tighter,belghazi2018mutual}. After
 describing the proposed methods, we investigate the outlier-robustness
 of the proposed methods. A list of notations mainly used in this
 section is given in Table~\ref{notations-estimation}.
 \begin{table}
  \caption{\label{notations-estimation} List of notations used in
  Section~\ref{sec:estimation-methods}.}
  \vspace{1mm}
  \begin{tabular}{|p{0.2\textwidth}|p{0.8\textwidth}|}
   \hline
   $\Exu$ & Expectation over $p(\bx,\bu)$\\
   $\Extu$ & Expectation over $p(\bx){p}(\bu)$\\
   $\Ex$ & Expectation over $p(\bx)$\\
   $\Eu$ & Expectation over $p(\bu)$\\
   $\Jga$ & $\gamma$-cross entropy\\
   $\wJga,~\tJga$ & Empirical approximations of $\Jga$\\
   $\Jdv$ &  Donsker-Varadhan variational lower-bound of MI\\
   $\wJdv,~\tJdv$ & Empirical approximations of $\Jdv$\\
   $\bm{u}_{\mathrm{p}}(t)$ & Random permutation of $\bm{u}(t)$ with respect to $t$\\ 
   $\bar{p}(\bm{x},\bm{u})$ & Contaminated joint probability density function of $\bm{x}$ and $\bm{u}$\\
   $\bar{p}(\bm{x})$ & Contaminated marginal probability density function of $\bm{x}$\\
   $\bar{p}(\bm{u})$ & Contaminated marginal probability density function of $\bm{u}$\\
   $\delta(\bm{x},\bm{u})$ & Joint probability density function 
for outliers \\
   $\delta(\bm{x}),~\delta(\bm{u})$ & Marginal probability density functions for outliers \\   
   $\epsilon$ & Contamination ratio ($0\leq\epsilon\leq{1}$) \\
   $\bar{\bm{x}},~\bar{\bm{u}}$ &  Outlier points \\
   $\delta_{\bar{\bx}}(\bx)$ &  Dirac delta function with a  point mass at $\bar{\bx}$\\
   $\delta_{\bar{\bu}}(\bu)$ & 
   Dirac delta function with a  point mass at $\bar{\bu}$\\
   $r_{\vtheta}(\bx,\bu)$ & Model parametrized by $\vtheta$\\
   $\bm{g}_{\vtheta}(\bm{x},\bm{u})$ & Gradient of $r_{\vtheta}(\bx,\bu)$ with respect to $\vtheta$\\
   $\vtheta^{\star}$ & Parameters optimized over the (noncontaminated) densities \\
   $\vtheta_{\epsilon}$ & Parameters optimized over the contaminated densities \\
   $\IFdv$ & Influence function over $\Jdv$\\
   $\Vdv$ & $\Vdv:=\Exu[\bm{g}_{\vtheta^{\star}}(\bm{X},\bm{U})
       \bm{g}_{\vtheta^{\star}}(\bm{X},\bm{U})^{\top}]
       -\Exu[\bm{g}_{\vtheta^{\star}}(\bm{X},\bm{U})]
       \Exu[\bm{g}_{\vtheta^{\star}}(\bm{X},\bm{U})]^{\top}$\\
   $\IFga$ & Influence function over $\Jga$\\
   $\bm{V}_{\gamma}$ & $\bm{V}_{\gamma}:=\Exu[\eta(\bX,\bU)^{\frac{\gamma}{1+\gamma}}S(\bX,\bU)^{\frac{1}{1+\gamma}}\bm{g}_{\vtheta^{\star}}(\bX,\bU)\bm{g}_{\vtheta^{\star}}(\bX,\bU)^{\top}]$\\ 
   $S(\bm{x},\bm{u})$ & $S(\bm{x},\bm{u}):=\frac{1}{1+e^{(1+\gamma)r_{\vtheta^{\star}}(\bm{x},\bm{u})}}$\\
   $\eta(\bm{x},\bm{u})$ & $\eta(\bm{x},\bm{u}):=S(\bm{x},\bm{u})(1-S(\bm{x},\bm{u}))$ \\
   $\bExu$ & Expectation over $\bar{p}(\bx,\bu)$\\
   $\bExtu$ & Expectation over $\bar{p}(\bx)\bar{p}(\bu)$\\
   $\bJga$ & $\gamma$-cross entropy over $\bar{p}(\bm{x},\bu)$ and $\bar{p}(\bm{x})\bar{p}(\bu)$\\
   \hline
  \end{tabular}
 \end{table}

  \subsection{Robust representation learning based on the $\gamma$-cross entropy}
  \label{ssec:RCL-gamma}
  The first method is based on the following $\gamma$-cross entropy for
  binary classification~\citep{fujisawa2008robust,hung2018robust}. Let
  us recall the following two datasets:
  \begin{align*}
   \calD_+:=\{(\bm{x}(t),
   \bm{u}(t)\}_{t=1}^{T}\sim{p}(\bx,\bu)\quad\text{vs.}
   \quad\calD_-:=\{(\bm{x}(t),\bm{u}^*(t))\}_{t=1}^{T}\sim{p}(\bx)p(\bu).
  \end{align*}
  By assigning class labels $y=0$ and $y=1$ to $\calD_+$ and $\calD_-$
  respectively, the $\gamma$-cross entropy for posterior probability
  estimation can be formulated as
  \begin{align*}
   \Jga(f_+,f_-)&:=-\frac{1}{\gamma}\log
   \left[\iint\sum_{y=0}^{1}
   p(y,\bx,\bu)\left(
   \frac{\{f_+(\bx,\bu)^{\gamma+1}\}^{1-y}\{f_-(\bx,\bu)^{\gamma+1}\}^{y}}
   {f_+(\bx,\bu)^{\gamma+1}+f_-(\bx,\bu)^{\gamma+1}}\right)^{\frac{\gamma}{\gamma+1}}
   \intd\bx\intd\bu\right],\\
   &=-\frac{1}{\gamma}\log\left[
   p(y=0)\Exu\left[
   \left(\frac{f_+(\bX,\bU)^{\gamma+1}}
   {f_+(\bX,\bU)^{\gamma+1}+f_-(\bX,\bU)^{\gamma+1}}\right)^{\frac{\gamma}{\gamma+1}}\right]\right.\\
   &\qquad\qquad\left.+p(y=1)\Extu\left[\left(\frac{f_-(\bX,\bU)^{\gamma+1}}
   {f_+(\bX,\bU)^{\gamma+1}+f_-(\bX,\bU)^{\gamma+1}}\right)^{\frac{\gamma}{\gamma+1}}
   \right]\right],
  \end{align*}
  where $f_+$ and $f_-$ are models for posterior probabilities, $p(y=0)$
  and $p(y=1)$ are class probabilities, and we used the following
  relation based on the datasets $\calD_+$ and $\calD_-$:
  \begin{align*}
   p(\bx,\bu|y=0)=p(\bx,\bu)\quad\text{and}\quad
   p(\bx,\bu|y=1)=p(\bx)p(\bu).
  \end{align*}
  By denoting $\log\frac{f_+(\bx,\bu)}{f_-(\bx,\bu)}$ by $r(\bx,\bu)$
  and assuming symmetric class probabilities (i.e.,
  $p(y=0)=p(y=1)=\frac{1}{2}$), the $\gamma$-cross entropy can be
  written as
  \begin{align}
   \Jga(r)&:=-\frac{1}{\gamma}\log\left[\Exu\left[
  \left(\frac{e^{(\gamma+1)r(\bX,\bU)}} {1+e^{(\gamma+1)r(\bX,\bU)}}
   \right)^{\frac{\gamma}{\gamma+1}}\right]
   +\Extu\left[\left(\frac{1}{1+e^{(\gamma+1)r(\bX,\bU)}}
   \right)^{\frac{\gamma}{\gamma+1}}\right]\right],  \label{binary}
  \end{align}
  where the term related to $p(y=0)$ and $p(y=1)$ is omitted because it
  is irrelevant to estimation of a model $r$. Since $r(\bx,\bu)$ is a
  model for the log-ratio of posterior probabilities, $\Jga(r)$ is
  minimized at
  \begin{align*}
   \log\frac{p(y=0|\bx,\bu)}{p(y=1|\bx,\bu)}=
   \log\frac{p(\bx,\bu|y=0)p(y=0)}{p(\bx,\bu|y=1)p(y=1)}
   =\log\frac{p(\bx,\bu)}{p(\bx)p(\bu)}.
  \end{align*}  
  Thus, the $\gamma$-cross entropy~\eqref{binary} can be used for
  density ratio estimation.  As proven in~\citet{fujisawa2008robust},
  the cross entropy in logistic regression can be obtained in the limit
  of the $\gamma$-cross entropy as follows:
  \begin{align*}
   \lim_{\gamma\to{0}}\Jga(r)=\Jlr(r),
  \end{align*}
  indicating $\Jga(r)$ can be regarded as a generalization of the cross
  entropy in logistic regression. The remarkable property of $\Jga(r)$
  is robustness against outliers. The positive parameter $\gamma$
  controls the robustness, and a larger value of $\gamma$ tends to be
  more robust to outliers.  We theoretically characterize the robustness
  of the $\gamma$-cross entropy in the context of density ratio
  estimation later.  On the other hand, as discussed
  in~\citet{fujisawa2008robust}, there seems to exist a trade-off
  between robustness and efficiency, indicating that outlier-robustness
  for the $\gamma$-cross entropy may come at a price of making
  estimation sample-inefficient.  In fact, the experimental results in
  Section~\ref{sssec:eff-robust} partially demonstrate this trade-off:
  The proposed method based on the $\gamma$-cross entropy is the most
  outlier-robust method, but it is not the best in terms of
  sample-efficiency.

  In practice, we empirically approximate $\Jga(r)$ as
  \begin{align*}
   \wJga(r):=-\frac{1}{\gamma}\log\left[\frac{1}{T}\sum_{t=1}^T\left\{
   \left(\frac{e^{(\gamma+1)r(\bm{x}(t),\bm{u}(t))}}
   {1+e^{(\gamma+1)r(\bm{x}(t),\bm{u}(t))}}
   \right)^{\frac{\gamma}{\gamma+1}}
   +\left(\frac{1}{1+e^{(\gamma+1)r(\bm{x}(t),\bup(t))}}
   \right)^{\frac{\gamma}{\gamma+1}}\right\}\right],
  \end{align*}  
  where $\bu_{\mathrm{p}}(t)$ denotes a random permutation of
  $\bm{u}(t)$ with respect to $t$. Another empirical approximation is
  also possible as
  \begin{align*}
   \tJga(r):=-\frac{1}{\gamma}\log\left[\frac{1}{T}
   \sum_{t=1}^T\left(\frac{e^{(\gamma+1)r(\bm{x}(t),\bm{u}(t))}}
   {1+e^{(\gamma+1)r(\bm{x}(t),\bm{u}(t))}}
   \right)^{\frac{\gamma}{\gamma+1}}
   +\frac{1}{T^2}\sum_{t=1}^T\sum_{t'=1}^T
   \left(\frac{1}{1+e^{(\gamma+1)r(\bm{x}(t),\bm{u}(t'))}}
   \right)^{\frac{\gamma}{\gamma+1}}\right].
  \end{align*}  
  These empirical objective functions are minimized with a minibatch
  stochastic gradient method in  Section~\ref{sec:exp}.
  \subsection{Nonlinear ICA with the Donsker-Varadhan variational estimation}
  \label{ssec:RCL-DV}
  Our second method is based on variational estimation of mutual
  information~\citep{ruderman2012tighter,belghazi2018mutual}, and
  employs the negative of the lower-bound in~\eqref{DV} as the objective
  function:
  \begin{align}
   \Jdv(r):=-\Exu[r(\bX,\bU)]+\log(\Extu[e^{r(\bX,\bU)}]).
   \label{dv-obj}
  \end{align}    
  As proved in~\citet{banerjee2006bayesian}
  and~\citet{belghazi2018mutual}, $\Jdv(r)$ is minimized at
  $\log\frac{p(\bm{x},\bm{u})}{p(\bm{x})p(\bm{u})}$ up to some constant,
  and thus can be used for density ratio estimation. $\Jdv(r)$ has been
  already employed in the context of maximization of mutual
  information~\citep{hjelm2019learning}. Here, our contribution is to
  apply $\Jdv(r)$ to nonlinear ICA, and we numerically demonstrate its
  usefulness.
  
  The objective function $\Jdv(r)$ has been independently derived in
  terms of density ratio estimation based on the
  KL-divergence~\citep{sugiyama2008direct,tsuboi2009direct}. Since
  $r(\bx,\bu)$ is a model for $\log\frac{p(\bx,\bu)}{p(\bx)p(\bu)}$, the
  following model $p_m(\bx,\bu)$ can be regarded as a density model for
  $p(\bx,\bu)$:
  \begin{align*}
   p_m(\bx,\bu):=\frac{p(\bx)p(\bu)e^{r(\bx,\bu)}}{\Extu[e^{r(\bX,\bU)}]},
  \end{align*}
  where the denominator ensures that $p_m(\bx,\bu)$ is normalized to be
  one.  Then, the KL-divergence between $p(\bx,\bu)$ and $p_m(\bx,\bu)$
  is given by
  \begin{align}
   \mathrm{KL}[p(\bx,\bu)||p_m(\bx,\bu)]
   &=\iint{p(\bx,\bu)}\log\frac{p(\bx,\bu)}{p_m(\bx,\bu)}
   \intd\bx\intd\bu \nonumber\\&=I(\calX,\calU)
   \underbrace{-\Exu[r(\bX,\bU)]+\log\left(
   \Extu[e^r(\bX,\bU)]\right)}_{=\Jdv(r)}.
   \label{KL-DV}
  \end{align}
  Eq.\eqref{KL-DV} shows that minimizing
  $\mathrm{KL}[p(\bx,\bu)||p_m(\bx,\bu)]$ is equal to minimizing
  $\Jdv(r)$ because the mutual information $I(\calX,\calU)$ on the
  right-hand side is a constant with respect to $r$.  As shown
  in~\citet{kanamori2010theoretical}, the density ratio estimator based
  on the KL-divergence could be more accurate than the estimator based
  on the cross entropy $\Jlr$ (i.e., logistic regression) under the
  miss-specified setting where the true density ratio is not necessarily
  included in the function class of a model $r$. In fact, we
  experimentally demonstrate that the nonlinear ICA method based on
  $\Jdv$ performs better than an existing method based on $\Jlr$ when
  the number of data samples is small.
  
  In practice, we empirically approximate $\Jdv(r)$ as
  \begin{align*}
   \wJdv(r):=-\frac{1}{T}\sum_{t=1}^T r(\bm{x}(t),\bm{u}(t))
   +\log\left(\frac{1}{T}\sum_{t=1}^Te^{r(\bm{x}(t),\bup(t))}\right),
  \end{align*}
  or
  \begin{align*}
   \tJdv(r):=-\frac{1}{T}\sum_{t=1}^T r(\bm{x}(t),\bm{u}(t))
   +\log\left(\frac{1}{T^2}\sum_{t=1}^T
   \sum_{t'=1}^Te^{r(\bm{x}(t),\bm{u}(t'))}\right).
  \end{align*}
  Section~\ref{sec:exp} uses both approximations in numerical
  experiments with a mini-batch stochastic gradient method.
  \subsection{Theoretical analysis for outlier-robustness}
  \label{ssec:analysis-outlier}
  Here, we investigate outlier-robustness of estimators based on $\Jga$
  and $\Jdv$. To this end, we consider the following two scenarios of
  contamination by outliers:
  \begin{itemize}
   \item \emph{Contamination model~1}: Given input data $\bm{x}$,
	 complementary data $\bu$ is conditionally contaminated by
	 outliers as follows:
	 \begin{align}
	  \bar{p}(\bu|\bx):=(1-\epsilon)p(\bu|\bx)+\epsilon\delta(\bu|\bx),
	  \label{cond-contamination}
	 \end{align}
	 where $\bar{p}(\bu|\bx)$ is the contaminated conditional
	 density of $\bu$ given $\bx$, $\delta(\bu|\bx)$ denotes the
	 conditional density for outliers, and $0\leq\epsilon<1$ denotes
	 the contamination ratio. By assuming that input data $\bm{x}$
	 is \emph{non}contaminated, \eqref{cond-contamination} leads to
	 the following contaminated joint and marginal densities:
	 \begin{align*}
	  \bar{p}(\bx,\bu)&:=(1-\epsilon)p(\bx,\bu)+\epsilon\delta(\bu|\bx){p}(\bx),\\
	  \quad\bar{p}(\bx):=p(\bx)\quad&\text{and}\quad
	  \bar{p}(\bu):=(1-\epsilon)p(\bu)+\epsilon\int\delta(\bu|\bx)p(\bx)\intd\bx,
	 \end{align*}
	 where $\bar{p}(\bx,\bu)$ is the contaminated joint density, and
	 $\bar{p}(\bx)$ and $\bar{p}(\bu)$ denote the contaminated
	 marginal densities of $\bx$ and $\bu$, respectively.
	 
   \item \emph{Contamination model~2}: Input data $\bm{x}$ and
	 complementary data $\bu$ are jointly contaminated by outliers:
	 \begin{align}
	  \bar{p}(\bx,\bu):=(1-\epsilon)p(\bx,\bu)+\epsilon\delta(\bx,\bu),
	  \label{joint-contamination}
	 \end{align}
	 where $\delta(\bx,\bu)$ denotes the joint density for outliers.
	 Eq.\eqref{joint-contamination} leads to the following
	 contaminated marginal densities:
	 \begin{align*}
	  \bar{p}(\bx):=(1-\epsilon)p(\bx)+\epsilon\delta(\bx)
	  \quad\text{and}\quad
	  \bar{p}(\bu):=(1-\epsilon)p(\bu)+\epsilon\delta(\bu),
	 \end{align*}
	 where 
	 \begin{align*}
	  \delta(\bx):=\int\delta(\bx,\bu)\intd\bu\quad\text{and}\quad
	 \delta(\bu):=\int\delta(\bx,\bu)\intd\bx.
	 \end{align*}
  \end{itemize}
  The fundamental difference between these contamination scenarios is
  whether or not input data $\bx$ is contaminated by outliers. Next, we
  perform two analyses for outlier-robustness: One analysis is performed
  under the condition that the contamination ratio $\epsilon$ is small,
  while another one does \emph{not} necessarily assume that $\epsilon$
  is small and thus heavy contamination of outliers is also within the
  scope of the analysis.
   \subsubsection{Influence function analysis}
   \label{ssec:IFanalysis}
   \begin{table}[t]
    \begin{center}
     \caption{\label{tab:IF_summary} Comparison of contamination models
     in influence function analysis (Propositions~\ref{prop:IF}
     and~\ref{prop:IFga}). $\delta_{\bar{\bx}}(\bx)$ and
     $\delta_{\bar{\bu}}(\bu)$ denote the Delta functions having point
     masses at $\bar{\bx}$ and $\bar{\bu}$, respectively.}
     \vspace{1mm}
     \begin{tabular}{c|c}
     \hline 
      Contamination model~1 & Contamination model~2 \\
      \hline      
      \begin{tabular}{l} 
       $\bar{p}(\bx,\bu)=(1-\epsilon)p(\bx,\bu)+\epsilon\delta_{\bar{\bu}}(\bu)p(\bx)$\\
       ${~~~~~}\bar{p}(\bx)=p(\bx)$\\
       ${~~~~~}\bar{p}(\bu)=(1-\epsilon)p(\bu)+\epsilon\delta_{\bar{\bu}}(\bu)$
      \end{tabular}
      & \begin{tabular}{l} 
	 $\bar{p}(\bx,\bu)=(1-\epsilon)p(\bx,\bu)+\epsilon\delta_{\bar{\bx}}(\bx)\delta_{\bar{\bu}}(\bu)$ \\
       ${~~~~~}\bar{p}(\bx)=(1-\epsilon)p(\bx)+\epsilon\delta_{\bar{\bx}}(\bx)$\\
       ${~~~~~}\bar{p}(\bu)=(1-\epsilon)p(\bu)+\epsilon\delta_{\bar{\bu}}(\bu)$
      \end{tabular}	 
      \\ \hline
     \end{tabular}
    \end{center}
   \end{table} 
   Here, we perform influence function
   analysis~\citep{hampel2011robust}, which is an established tool in
   robust statistics and assumes that the densities for outliers are
   given as follows:
   \begin{align}
    \delta(\bx,\bu)=\delta_{\bar{\bx}}(\bx)\delta_{\bar{\bu}}(\bu), 
    \quad\delta(\bx)=\delta_{\bar{\bx}}(\bx)\quad\text{and}\quad
    \delta(\bu)=\delta_{\bar{\bu}}(\bu), 
    \label{IF-contamination-density}
   \end{align}
   where $\delta_{\bar{\bx}}(\bx)$ and $\delta_{\bar{\bu}}(\bu)$ are the
   Dirac delta functions having point masses at $\bar{\bx}$ and
   $\bar{\bu}$, respectively.  Eq.\eqref{IF-contamination-density}
   implies that $\delta(\bu|\bx)=\delta_{\bar{\bu}}(\bu)$ and
   $\int\delta(\bu|\bx)p(\bx)\intd\bx=\delta_{\bar{\bu}}(\bu)$ in the
   contamination model~1. Table~\ref{tab:IF_summary} is a comparison of
   contamination models~1 and~2 in this analysis.

   In this analysis, we suppose that a model $r_{\vtheta}(\bx,\bu)$ is
   parametrized by $\bm{\theta}$ and has the following properties:
   \begin{itemize}
    \item $r_{\vtheta}(\bx,\bu)$ and its gradient
	  $\bm{g}_{\vtheta}(\bm{x},\bm{u})$ are continuous where
	  $\bm{g}_{\vtheta}(\bm{x},\bm{u}):=\nabla_{\vtheta}r_{\vtheta}(\bx,\bu)$.

    \item $|r_{\vtheta}(\bx,\bu)|\to\infty$ and
	  $\|\bm{g}_{\vtheta}(\bm{x},\bm{u})\|\to\infty$ as
	  $\|\bx\|\to\infty$ and/or $\|\bu\|\to\infty$.
   \end{itemize}   
   These properties are simply used to discuss the robustness of the
   estimators based on the influence function, and satisfied by standard
   models such neural networks with an unbounded activation function
   (e.g., the softplus function). Based on $r_{\vtheta}(\bx,\bu)$, we
   define two estimators as
   \begin{align*}
    \vtheta^{\star}&:=\argmin_{\vtheta}J(r_{\vtheta})
    \quad\text{and}\quad \vtheta_{\epsilon}:=
    \argmin_{\vtheta}\bar{J}(r_{\vtheta}),
   \end{align*}
   where $J$ is some objective function over the \emph{non}contaminated
   densities (e.g., $\Jga$), while $\bar{J}$ is the contaminated version
   of $J$ computed over the contaminated densities $\bar{p}(\bx,\bu)$,
   $\bar{p}(\bx)$ and $\bar{p}(\bu)$ instead of $p(\bx,\bu)$, $p(\bx)$
   and $p(\bu)$. Then, we define the \emph{influence function} as
   \begin{align}
    {\rm IF}(\bar{\bx},\bar{\bu})&= \lim_{\epsilon\to
    0}\frac{\vtheta^{\star}-\vtheta_{\epsilon}}{\epsilon}.
    \label{defi-IF}
   \end{align}
   Eq.\eqref{defi-IF} indicates that $\|{\rm IF}(\bar{\bx},\bar{\bu})\|$
   measures how the estimator $\vtheta^{\star}$ is influenced by small
   contamination of outliers $\bar{\bx}$ and $\bar{\bu}$, and a larger
   $\|{\rm IF}(\bar{\bx},\bar{\bu})\|$ implies stronger influence by
   outliers. Based on the influence function, some robust properties of
   estimators can be characterized: Estimator $\vtheta^{\star}$ is
   called \emph{B-robust} if $\sup_{\bar{\bx},\bar{\bu}}\|{\rm
   IF}(\bar{\bx},\bar{\bu})\|<\infty$~\citep{hampel2011robust}.
   B-robustness guarantees that the influence from outliers is limited.
   A more favorable property is the \emph{redescending
   property}~\citep{hampel2011robust}, which defined as
   $\lim_{\|\bar{\bx}\|,\|\bar{\bu}\|\to\infty} \|{\rm
   IF}(\bar{\bx},\bar{\bu})\|=0$.  The redescending property ensures
   that $\vtheta^{\star}$ has almost no influence from even strongly
   deviated data $\bar{\bx}$ and/or $\bar{\bu}$.

   First, the following proposition shows the forms of the influence
   functions for $\Jdv$:
   \begin{proposition}
    \label{prop:IF} Assume that
    $r_{\vtheta}(\bx,\bu)=\log\frac{p(\bx,\bu)}{p(\bx)p(\bu)}$ at
    $\vtheta=\vtheta^{\star}$. For the contamination model~1, the
    influence function of the estimator $\vtheta^{\star}$ based on
    $\Jdv$ is given by
    \begin{align}
     \IFdv(\bar{\bu}) &=\Vdv^{-1}
     \Ex[\bm{g}_{\vtheta^{\star}}(\bX,\bar{\bu})-
     e^{r_{\vtheta^{\star}}(\bX,\bar{\bu})}\bm{g}_{\vtheta^{\star}}(\bX,\bar{\bu})],
     \label{IFdv1}
    \end{align}
    where
    $\bm{g}_{\vtheta^{\star}}(\bx,\bu)=\nabla_{\bm{\theta}}r_{\vtheta}(\bx,\bu)|_{\vtheta=\vtheta^{\star}}$
    and
    \begin{align}
     \Vdv&:=-\Exu[\bm{g}_{\vtheta^{\star}}(\bm{X},\bm{U})
     \bm{g}_{\vtheta^{\star}}(\bm{X},\bm{U})^{\top}]
     +\Exu[\bm{g}_{\vtheta^{\star}}(\bm{X},\bm{U})]
     \Exu[\bm{g}_{\vtheta^{\star}}(\bm{X},\bm{U})]^{\top}.
     \label{Vdv}
    \end{align}
    Regarding the contamination model~2, the influence function
    is given by
    \begin{align}
     \IFdv(\bar{\bx},\bar{\bu})
     &=\Vdv^{-1}\Big[\bm{g}_{\vtheta^{\star}}(\bar{\bx},\bar{\bu})
     +\Exu[\bm{g}_{\vtheta^{\star}}(\bX,\bU)]\nonumber\\
     &\qquad
     -\Ex[e^{r_{\vtheta^{\star}}(\bX,\bar{\bu})}\bm{g}_{\vtheta^{\star}}(\bX,\bar{\bu})]
     -\Eu[e^{r_{\vtheta^{\star}}(\bar{\bx},\bU)}\bm{g}_{\vtheta^{\star}}(\bar{\bx},\bU)]
     \Big],
     \label{IFdv2}
    \end{align}
    where $\Ex$ and $\Eu$ denote the expectations over $p(\bx)$ and
    $p(\bu)$, respectively.
   \end{proposition} 
   The proof is deferred in Appendix~\ref{app:IF}. The key to interpret
   the influence functions in Proposition~\ref{prop:IF} is unboundedness
   of $r_{\vtheta^{\star}}(\bx,\bu)$ and its gradient
   $\bm{g}_{\vtheta^{\star}}(\bx,\bu)$.  For example, the right-hand
   side of~\eqref{IFdv1} indicates that $\|\IFdv(\bar{\bu})\|$ can be
   unbounded when $r_{\vtheta^{\star}}(\bx,\bar{\bu})\to\infty$ and
   $\|\bm{g}_{\vtheta^{\star}}(\bx,\bar{\bu})\|\to\infty$ as
   $\|\bar{\bu}\|\to\infty$. This means that the estimator based on
   $\Jdv$ is not B-robust, and can be strongly influenced by outliers
   $\bar{\bu}$. The same discussion is applicable to the influence
   function~\eqref{IFdv2} of the contamination model~2. Thus, these
   results imply that estimation based on $\Jdv$ can be sensitive to
   outliers.

   Next, we show the influence functions for the $\gamma$-cross entropy
   $\Jga$ in the following proposition:
   \begin{proposition}
    \label{prop:IFga} Suppose the same assumption holds as
    Proposition~\ref{prop:IF}.  Regarding the contamination model~1, the
    influence function of the estimator $\vtheta^{\star}$ based on
    $\Jga$ is obtained as follows:
    \begin{align}
     \IFga(\bar{\bu})=V_{\gamma}^{-1}
     &\Ex[
     \{S(\bX,\bar{\bu})^{\frac{1}{1+\gamma}}
     -(1-S(\bX,\bar{\bu}))^{\frac{1}{1+\gamma}}\}
     \eta(\bX,\bar{\bu})^{\frac{\gamma}{1+\gamma}}
     \bm{g}_{\vtheta^{\star}}(\bX,\bar{\bu})],
     \label{IFga1}
    \end{align}
    where
    \begin{align}
     V_{\gamma}&:=\Exu[\eta(\bX,\bU)^{\frac{\gamma}{1+\gamma}}
     S(\bX,\bU)^{\frac{1}{1+\gamma}}
     \bm{g}_{\vtheta^{\star}}(\bX,\bU)\bm{g}_{\vtheta^{\star}}(\bX,\bU)^{\top}],\\ 
     S(\bm{x},\bm{u})&:=\frac{1}{1+e^{(1+\gamma)r_{\vtheta^{\star}}(\bm{x},\bm{u})}} 
     \quad\text{and}\quad
     \eta(\bm{x},\bm{u}):=S(\bm{x},\bm{u})(1-S(\bm{x},\bm{u})).	    
     \label{defi-eta}
    \end{align}
    Regarding the contamination model~2, the influence function is given
    by
    \begin{align}
     \IFga(\bar{\bx},\bar{\bu})
     &=V_{\gamma}^{-1}\Big[
     \eta(\bar{\bx},\bar{\bu})^{\frac{\gamma}{1+\gamma}}
     S(\bar{\bx},\bar{\bu})^{\frac{1}{1+\gamma}}\bm{g}_{\vtheta^{\star}}(\bar{\bx},\bar{\bu})\nonumber\\
     &\qquad-\Ex[\eta(\bX,\bar{\bu})^{\frac{\gamma}{1+\gamma}}
     (1-S(\bX,\bar{\bu}))^{\frac{1}{1+\gamma}}\bm{g}_{\vtheta^{\star}}(\bX,\bar{\bu})]
     \nonumber \\
     &\qquad-\Eu[\eta(\bar{\bx},\bU)^{\frac{\gamma}{1+\gamma}}
     (1-S(\bar{\bx},\bU))^{\frac{1}{1+\gamma}}\bm{g}_{\vtheta^{\star}}(\bar{\bx},\bU)]
     \nonumber \\
     &\qquad+\Extu[\eta(\bX,\bU)^{\frac{\gamma}{1+\gamma}}
     (1-S(\bX,\bU))^{\frac{1}{1+\gamma}}\bm{g}_{\vtheta^{\star}}(\bX,\bU)]\Big].
     \label{IFga2}
    \end{align}
   \end{proposition}    
   The proof of Proposition~\ref{prop:IFga} is not given because it is
   almost the same as Proposition~\ref{prop:IF}. In contrast to $\Jdv$,
   the influence functions based on the $\gamma$-cross entropy $\Jga$
   include the weight functions $\eta(\bx,\bu)$ and $S(\bx,\bu)$, and
   $\|\IFga(\bar{\bu})\|$ and $\|\IFga(\bar{\bx},\bar{\bu})\|$ can be
   bounded even when
   $\|\bm{g}_{\vtheta^{\star}}(\bar{\bx},\bar{\bu})\|\to\infty$ as
   $\|\bar{\bx}\|\to\infty$ and/or $\|\bar{\bu}\|\to\infty$.  The
   following corollary characterizes the robustness of the
   $\gamma$-cross entropy and shows that in stark contrast with $\Jdv$,
   the estimator based on $\Jga$ is B-robust for both contamination
   models, and has the redescending property for the contamination
   model~1 under some conditions:
   \begin{corollary}
    \label{coro:robust} Assume that
    \begin{align}
     \sup_{\bx,\bu}\|\eta(\bx,\bu)^{\frac{\gamma}{1+\gamma}}
     \bm{g}_{\vtheta^{\star}}(\bx,\bu)\|&<\infty.
     \label{coro:sup}
    \end{align}    
    Then, the estimator $\vtheta^{\star}$ based on $\Jga$ is B-robust
    both for the contamination models~1 and~2. Another assumption is
    made as follows:
    \begin{align}
     \lim_{\|\bar{\bu}\|\to\infty}
     \|\eta(\bx,\bar{\bu})^{\frac{\gamma}{1+\gamma}}
     \bm{g}_{\vtheta^{\star}}(\bx,\bar{\bu})\|&=0\quad\text{for all $\bx$}.
     \label{coro:limit}
    \end{align}    
    Then, the estimator $\vtheta^{\star}$ based on $\Jga$ has the
    redescending property for the contamination model~1.
   \end{corollary}
   The results are an almost direct consequence of
   Proposition~\ref{prop:IFga}, and thus the detailed calculation is
   omitted.  Assumptions~\eqref{coro:sup} and~\eqref{coro:limit} would
   be mild: By definition~\eqref{defi-eta}, the function $\eta(\bx,\bu)$
   exponentially convergences to zero when
   $r_{\vtheta^{\star}}(\bx,\bu)\to\infty$ as $\|\bx\|\to\infty$ and/or
   $\|\bu\|\to\infty$. Thus, B-robustness would hold under both
   contamination models for standard models (e.g., neural networks). On
   the other hand, the redescending property is shown only for the
   contamination model~1. This comes from the fact that the
   contamination of outliers is more complicated in the contamination
   model~2: Both $\bx$ and $\bu$ are contaminated by outliers in the
   contamination model~2, while $\bu$ is only contaminated in the
   contamination model~1. However, we numerically demonstrate that the
   proposed method based on the $\gamma$-cross entropy is robust against
   outliers even for the contamination model~2.
   
   Proposition~\ref{prop:IF} also reveals the outlier weakness of the
   cross entropy $\Jlr$ used in logistic regression. We recall that the
   $\gamma$-cross entropy $\Jga$ approaches $\Jlr$ as $\gamma\to{0}$.
   Then, it is obvious that Assumptions~\eqref{coro:sup}
   and~\eqref{coro:limit} in Corollary~\ref{coro:robust} are never
   satisfied in the limit of $\gamma\to{0}$ when
   $\|\bm{g}_{\vtheta^{\star}}(\bx,\bu)\|$ is an unbounded function
   (e.g., neural networks). This implies that the estimator based on
   $\Jlr$ is neither B-robust nor redescending, and estimation based on
   $\Jlr$ can be hampered by outliers.

   In short, the analysis based on the influence function implies that
   it is a promising approach to use the $\gamma$-cross entropy for
   density ratio estimation in the presence of outliers even with neural
   networks, while estimation based on $\Jdv$ and $\Jlr$ can be strongly
   influenced by outliers particularly when $r_{\vtheta}(\bx,\bu)$ is
   modelled by using unbounded functions such as neural networks.
   \subsubsection{Strong robustness of the $\gamma$-cross entropy}
   \label{ssec:strong-robust}
   Here, we investigate the robustness of the $\gamma$-cross entropy
   even when the contamination ratio $\epsilon$ is \emph{not}
   necessarily assumed to be small. Unlike influence function analysis,
   the contaminated densities, $\delta(\bx,\bu)$, $\delta(\bx)$ and
   $\delta(\bu)$, are not Dirac delta functions, but rather general
   probability density functions. To clarify the notations, we denote
   the $\gamma$-cross entropy over the contaminated densities as
   \begin{align*}
    \bJga(r):=-\frac{1}{\gamma}\log\left[
    \bExu\left[\left(\frac{e^{(\gamma+1)r(\bX,\bU)}}{1+e^{(\gamma+1)r(\bX,\bU)}}
    \right)^{\frac{\gamma}{\gamma+1}}\right] +\bExtu\left[
    \left(\frac{1}{1+e^{(\gamma+1)r(\bX,\bU)}}
    \right)^{\frac{\gamma}{\gamma+1}}\right]\right],
   \end{align*}
   where $\bExu$ and $\bExtu$ denote the expectations over
   $\bar{p}(\bx,\bu)$ and $\bar{p}(\bx)\bar{p}(\bu)$ in contamination
   models, respectively.

   The remarkable property of the $\gamma$-cross entropy is \emph{strong
   robustness}~\citep{fujisawa2008robust}: The latent bias caused from
   outliers can be small even in the case of heavy contamination (i.e.,
   nonsmall $\epsilon$). To prove the strong robustness in the context
   of density ratio estimation, we recall that $r(\bx,\bu)$ is defined
   by
   \begin{align*}
    r(\bx,\bu)=\psi(\bhx(\bm{x}),\bhu(\bm{u}))+a(\bhx(\bx))+b(\bhu(\bu)).
   \end{align*}
   The following proposition proved in Appendix~\ref{app:strong-robust}
   focuses on the contamination model~1 and establishes certain
   conditions for strong robustness:
   \begin{proposition}
    \label{prop:strong-robust} Let us define a constant $\nu_1$ as    
    \begin{align*}
     \nu_1:=\iint{p}(\bx)\delta(\bu|\bx)
     \left(\frac{e^{(\gamma+1)r(\bx,\bu)}}{1+e^{(\gamma+1)r(\bx,\bu)}}
     \right)^{\frac{\gamma}{\gamma+1}}\hspace{-3mm}\intd\bx\intd\bu
     +\iint{p}(\bx)\delta(\bu)
     \left(\frac{1}{1+e^{(\gamma+1)r(\bx,\bu)}}\right)^{\frac{\gamma}{\gamma+1}}
     \hspace{-3mm}\intd\bx\intd\bu.
    \end{align*}	  
    We suppose the contamination model~1 and make the following
    assumptions:
    \begin{enumerate}[(D1)]
     \item Assume that
	   \begin{align*}
	    &\iint p(\bx)\delta(\bu|\bx)
	    \left(\frac{e^{(\gamma+1)\{\psi(\bhx(\bm{x}),\bhu(\bm{u}))+a(\bhx(\bx))\}/2}}
  	    {e^{-(\gamma+1){b}(\bhu(\bm{u}))}
	    +e^{(\gamma+1)\{\psi(\bhx(\bm{x}),\bhu(\bm{u}))+a(\bhx(\bx))\}}}
	    \right)^{\frac{2\gamma}{\gamma+1}}\hspace{-3mm}
	    \intd\bu\intd\bx<\infty\\
	    &\quad\text{and}~
	    \iint p(\bx)\delta(\bu)
	    \left(\frac{e^{-(\gamma+1){b}(\bhu(\bm{u}))/2}}
  	    {e^{-(\gamma+1){b}(\bhu(\bm{u}))}
	    +e^{(\gamma+1)\{\psi(\bhx(\bm{x}),\bhu(\bm{u}))+a(\bhx(\bx))\}}}
	    \right)^{\frac{2\gamma}{\gamma+1}}\hspace{-3mm}\intd\bu\intd\bx<\infty.
	   \end{align*}
	   
     \item The following integrals are sufficiently small:
     	   \begin{align*}
     	    \iint p(\bx)\delta(\bu|\bx)
	    e^{\gamma\{\psi(\bhx(\bm{x}),\bhu(\bm{u}))+a(\bhx(\bx))\}}
	    \intd\bu\intd\bx
     	    ~~\text{\and}~~
	    \int \delta(\bu) e^{-\gamma{b}(\bhu(\bm{u}))} \intd\bu.
      	   \end{align*}
    \end{enumerate}
    Then, it holds that $\nu_1$ is sufficient small and
    \begin{align}
     \bJga(r)&=\Jga(r)-\frac{1}{\gamma}\log(1-\epsilon)
     +O\left(\frac{\epsilon}{1-\epsilon}\nu_1\right).
     \label{strong-robust}
    \end{align}
   \end{proposition}
   The proof can be found in Appendix~\ref{app:strong-robust}.
   Eq.\eqref{strong-robust} in Proposition~\ref{prop:strong-robust}
   indicates that for the contamination model~1, minimization of
   $\bJga(r)$ is approximately equal to minimization of $\Jga(r)$ over
   \emph{non}contaminated densities under Assumptions~(D1-2) even when
   the contamination ratio $\epsilon$ is not necessarily small (e.g.,
   $\epsilon=0.2$ as in the numerical experiments
   of~\citet[Section~8]{fujisawa2008robust}). This means that we could
   perform density ratio estimation in many practical situations almost
   as if outliers did not exist.

   Assumption~(D1) implies that $\delta(\bu|\bx)$ and $\delta(\bu)$ are
   localized on the domain of $\bu$ and have a fast decay on their tails
   as often seen in densities for outliers: For example, when both
   $\delta(\bu)$ and $\delta(\bu|\bx)$ have finite supports\footnote{The
   support of $\delta(\bu)$ is defined by
   $\{\bu~|~\delta(\bu)\neq0\}$.}, Assumption~(D1) is fulfilled.
   Assumption~(D2) means that outliers drawn from $\delta(\bu|\bx)$ and
   $\delta(\bu)$ exist on the respective tails of
   $e^{\psi(\bhx(\bm{x}),\bhu(\bm{u}))+a(\bhx(\bx))}$ and
   $e^{-b(\bhu(\bm{u}))}$, which can be roughly regarded as models of
   $p(\bu|\bx)$ and $p(\bu)$ respectively because $r(\bx,\bu)$ is
   modelling
   $\log\frac{p(\bx,\bu)}{p(\bx)p(\bu)}=\log{p}(\bu|\bx)-\log{p}(\bu)$.
   Thus, when $\bm{r}(\bx,\bu)$ is near to
   $\log\frac{p(\bx,\bu)}{p(\bx)p(\bu)}$, Assumption~(D2) implies that
   the outliers exist on the tails of $p(\bx,\bu)$ and $p(\bu)$, and
   indeed reflects typical contamination by outliers. Overall,
   Assumptions~(D1-2) would be practically reasonable.
   
   For the contamination model~2, the strong robustness does not hold,
   but the $\gamma$-cross entropy still has a favorable property which
   we call \emph{semistrong robustness}. The detail of the semistrong
   robustness is given in Appendix~\ref{app:semistrong-robust}. In
   addition, we experimentally demonstrate that the proposed method
   based on the $\gamma$-cross entropy is robust against outliers under
   the contamination model~2.   
 \section{Numerical Experiments}
 \label{sec:exp}
  This section numerically investigates how the proposed methods work in
  nonlinear ICA and a downstream task for linear classification, and
  empirically supports the implication of the theoretical analysis for
  nonlinear ICA in Section~\ref{ssec:nonlinear-ICA}.

  \subsection{Nonlinear ICA on artificial data}
  \label{ssec:nonlinearICA-exp}
  Here, we demonstrate the numerical performance of the proposed methods
  in nonlinear ICA and investigate how the dimensionality of
  complementary data affects the source recovery as implied in
  Section~\ref{ssec:nonlinear-ICA}. All experiments suppose $\dx=\Dx$
  according to Proposition~\ref{theo:general-ICA} and
  Theorem~\ref{prop:general-PCL}.
   \subsubsection{Sample efficiency and  outlier robustness}
   \label{sssec:eff-robust}
   We followed the experimental setting of a nonlinear ICA method called
   \emph{permutation contrastive learning} (PCL)
   in~\citet{pmlr-v54-hyvarinen17a}, which is intended for temporally
   dependent data $\bm{x}(t),~t=1,\dots,T$ and supposes the
   complementary data samples are $\bm{u}(t)=\bm{x}(t-1)$ (i.e., past
   input data). First, as in the autoregressive process, the
   ten-dimensional temporally dependent $T$ sources were generated from
   the contaminated density model as
   \begin{align}
    \bar{p}(\bm{s}(t)|\bm{s}(t-1))
    =(1-\epsilon)p(\bm{s}(t)|\bm{s}(t-1))+\epsilon\delta(\bm{s}(t)|\bm{s}(t-1)),
    \label{outlier-model}
   \end{align}
   where $\epsilon$ denotes the outlier ratio, and with $\rho=0.7$ and $\Dx=10$,
   \begin{align}
    p(\bm{s}(t)|\bm{s}(t-1))&\propto
    \prod_{i=1}^{\Dx} \exp\left(-\sqrt{2}|s_i(t)-\rho s_i(t-1)|/(1-\rho^2)\right)
    \label{exp:cond-source}\\  
    \delta(\bm{s}(t)|\bm{s}(t-1))&\propto
    \prod_{i=1}^{\Dx}\exp\left(-\frac{\{s_i(t)+\rho s_i(t-1)\}^2}{2}\right).
    \label{exp:cond-ourlier}
   \end{align}
   Then, input data $\bm{x}$ was generated according to
   $\bx=\bm{f}(\bm{s})$ in~\eqref{ICA-model} where $\bm{f}$ was modelled
   by a multi-layer feedforward neural network with the leaky ReLU
   activation function and random weights. The numbers of all hidden and
   output units were the same as the dimensionality of data (i.e.,
   $\Dx$).  Since complementary data $\bu(t)$ is past input data
   $\bx(t-1)$, the contamination process corresponds to the
   contamination model~2 in Section~\ref{ssec:analysis-outlier} where
   both $\bm{x}(t)$ and $\bm{u}(t)$ are contaminated by outliers.  As
   preprocessing, we performed whitening based on the
   $\gamma$-cross entropy~\citep{chen2013robust}.
   
   We applied the following methods to data samples:
   \begin{itemize}
    \item Permutation contrastive learning
	  (PCL)~\citep{pmlr-v54-hyvarinen17a}: A nonlinear ICA method
	  for temporally dependent sources based on logistic regression
	  whose objective function is an empirical version of $\Jlr$.
	  
    \item DV-PCL: A proposed nonlinear ICA method for temporally
	  dependent sources based on the Donsker-Varadhan variational
	  estimation $\wJdv$ (Section~\ref{ssec:RCL-DV}).
	  
    \item Robust PCL (RPCL): A proposed nonlinear ICA method for
	  temporally dependent sources based on the $\gamma$-cross
	  entropy $\wJga$ (Section~\ref{ssec:RCL-gamma}). The value of
	  $\gamma$ was increased from $0.0$ to $5.0$ during minibatch
	  stochastic gradient at every $100$ epoch where $\gamma=0.0$
	  means to perform the standard logistic regression.
   \end{itemize}
   We used a same model $r(\bx,\bu)$ for all three methods: The
   representation function $\bhx$ was modelled by a multi-layer
   feedforward neural network where the number of hidden units was
   $4\Dx$, but the final layer was $\Dx$. The number of layers was the
   same as $\bm{f}(\bm{s})$ in the data generative
   model~\eqref{ICA-model}.  The activation functions in hidden layers
   were the max-out function~\citep{goodfellow2013maxout} with two
   groups, while the final layer has no activation function. Regarding
   the representation function $\bhu$, the same architecture as $\bhx$
   was employed with parameter
   sharing (i.e., $\bhu=\bhx$). Following~\citet{pmlr-v54-hyvarinen17a}, $r(\bm{x},\bm{u})$
   was modelled by
   \begin{align}
    r(\bm{x},\bm{u})=\sum_{i=1}^{\Dx}
    |a_{i,1}h_{\mathrm{x},i}(\bm{x})+a_{i,2}h_{\mathrm{u},i}(\bm{u})+b_i|
    -(\bar{a}_{i}\hx{i}(\bm{x})+\bar{b}_i)^2+c,
    \label{psi-model}
   \end{align}
   where $a_{i,1}, a_{i,2}, b_i, \bar{a}_i, \bar{b}_i, c$ are parameters
   to be estimated from data.  All parameters were estimated by the Adam
   optimizer for $1,600$ epochs with mini-batch size $256$ and learning
   rate $0.001$. We also applied the $\ell_2$ regularization to the
   weight parameters with the regularization parameter $10^{-4}$.
   
   With the test sources $s^{\mathrm{te}}_i(t)$ without outliers drawn
   from $p(\bm{s}(t)|\bm{s}(t-1))$, the performance was measured by the
   mean absolute correlation as
   \begin{align*}
    \frac{1}{T_{\mathrm{te}}\Dx}\sum_{i=1}^{\Dx}
    \left|\sum_{t=1}^{T_{\mathrm{te}}}
    (\hat{h}_{\mathrm{x},i}(\bx^{\mathrm{te}}(t))-\mu^{\hat{\mathrm{h}}}_i)
    (s^{\mathrm{te}}_{\pi(i)}(t)-\mu^{\mathrm{te}}_{\pi(i)})\right|,
   \end{align*}
   where $\hat{h}_{\mathrm{x},i}$ denotes the learned representation
   function, $T_{\mathrm{te}}=50,000$ denotes the number of test
   samples, and $\mu^{\hat{\mathrm{h}}}_i$ and $\mu^{\mathrm{te}}_i$ are
   the sample means of $\hat{h}_{\mathrm{x},i}(\bx^{\mathrm{te}}(t))$
   and $s^{\mathrm{te}}_i(t)$, respectively. Since estimation of
   nonlinear ICA is indeterminant with respect to permutation (i.e.,
   ordering) of the source components $s_i$, the permutation indices
   $\pi(i)\in\{1,2,\dots,\Dx\}$ were determined so that the absolute
   mean correlation was maximized.
   
\begin{figure}[t]
    \centering
    \subfigure[$L=1$]{\includegraphics[width=.32\textwidth]{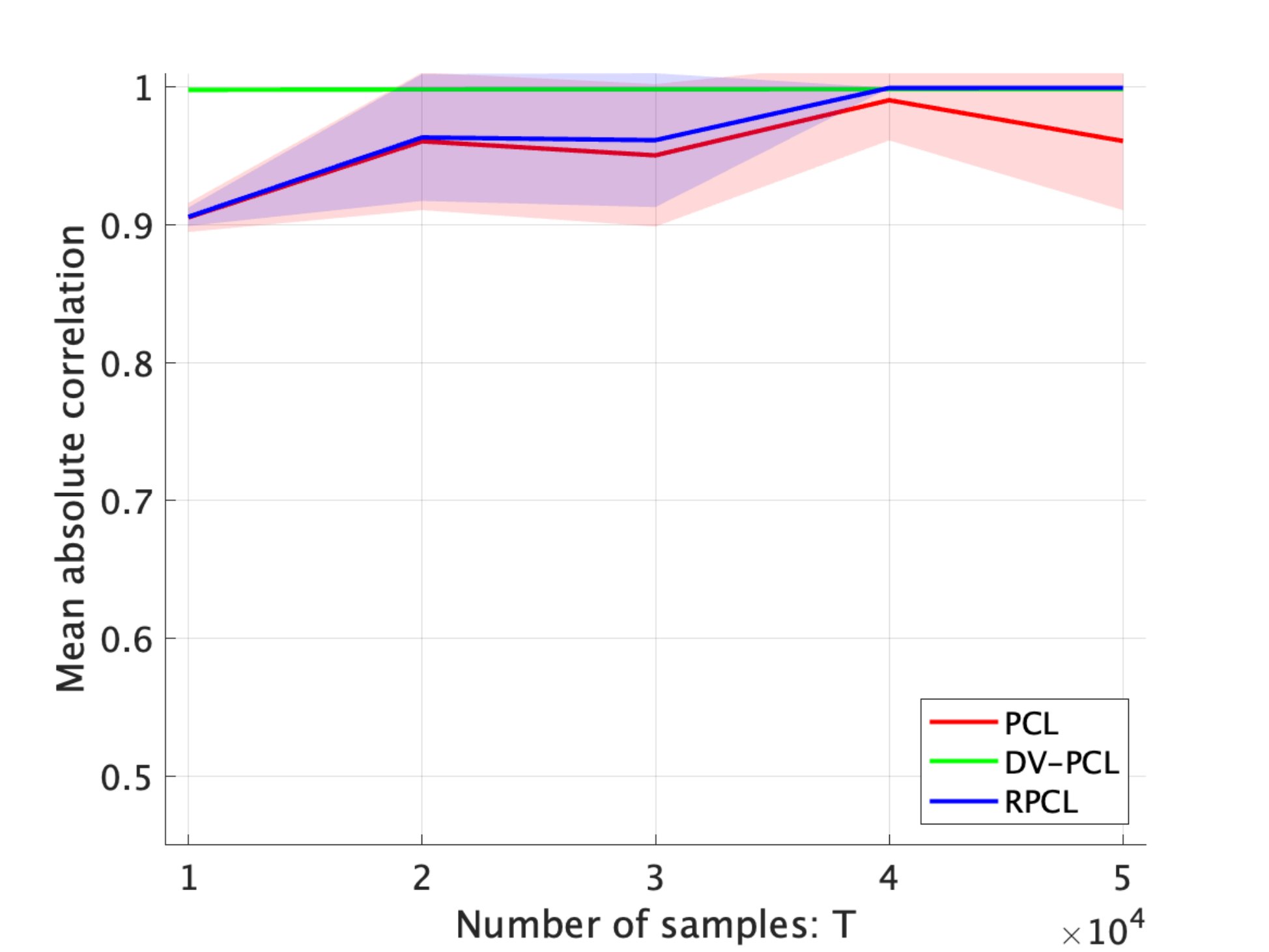}}
    \subfigure[$L=2$]{\includegraphics[width=.32\textwidth]{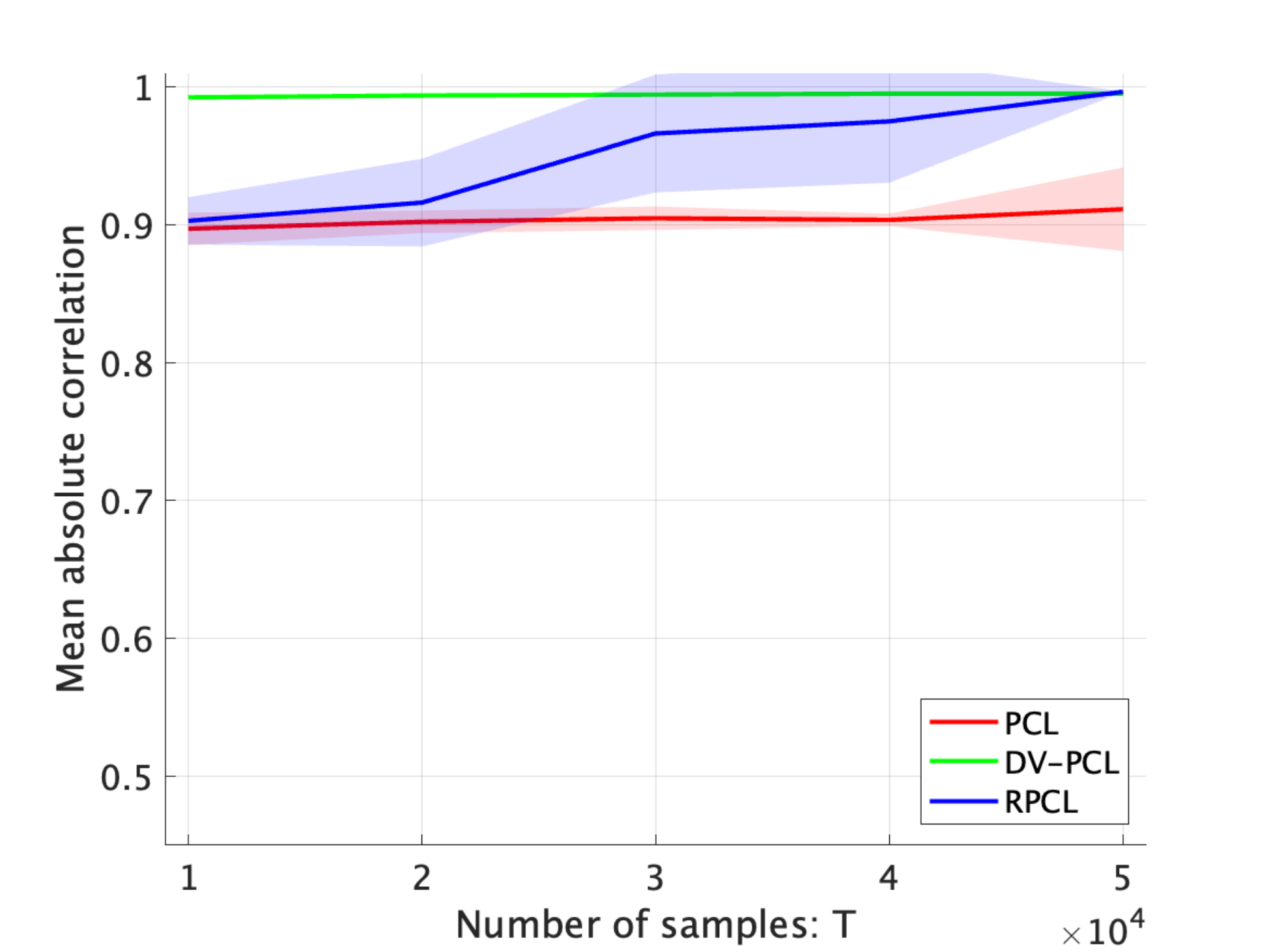}}
    \subfigure[$L=3$]{\includegraphics[width=.32\textwidth]{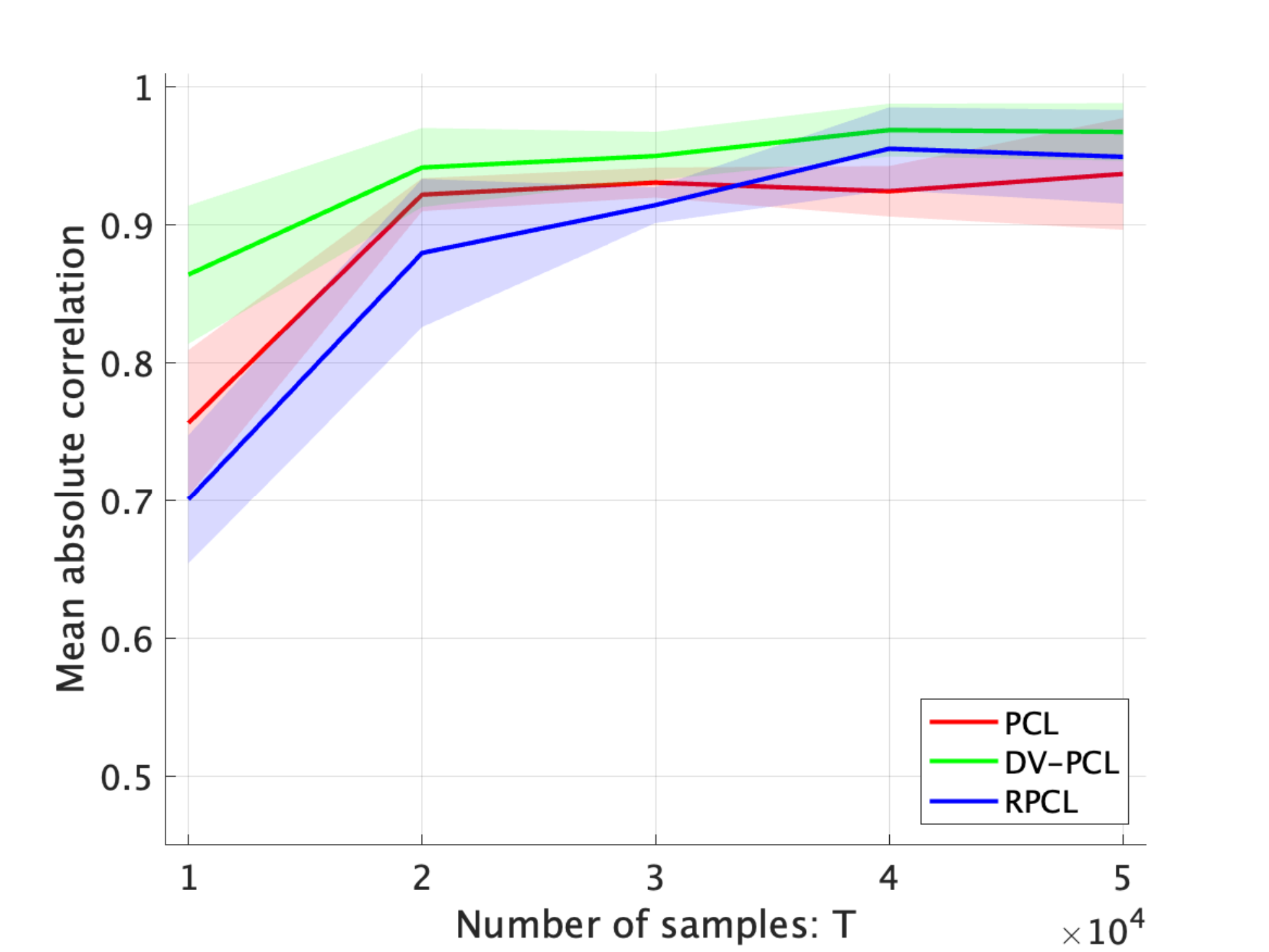}}
    \subfigure[$L=4$]{\includegraphics[width=.32\textwidth]{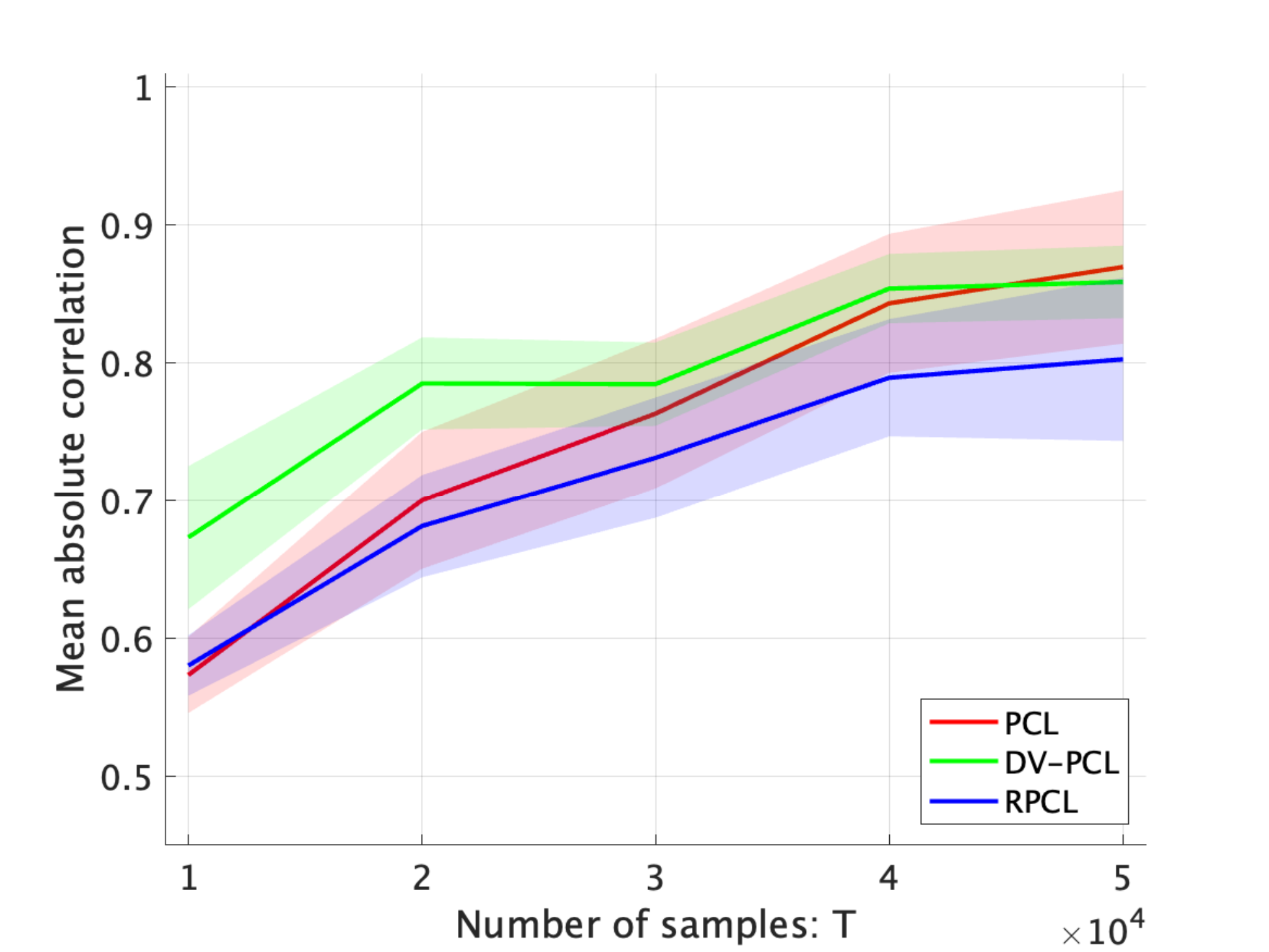}}
    \subfigure[$L=5$]{\includegraphics[width=.32\textwidth]{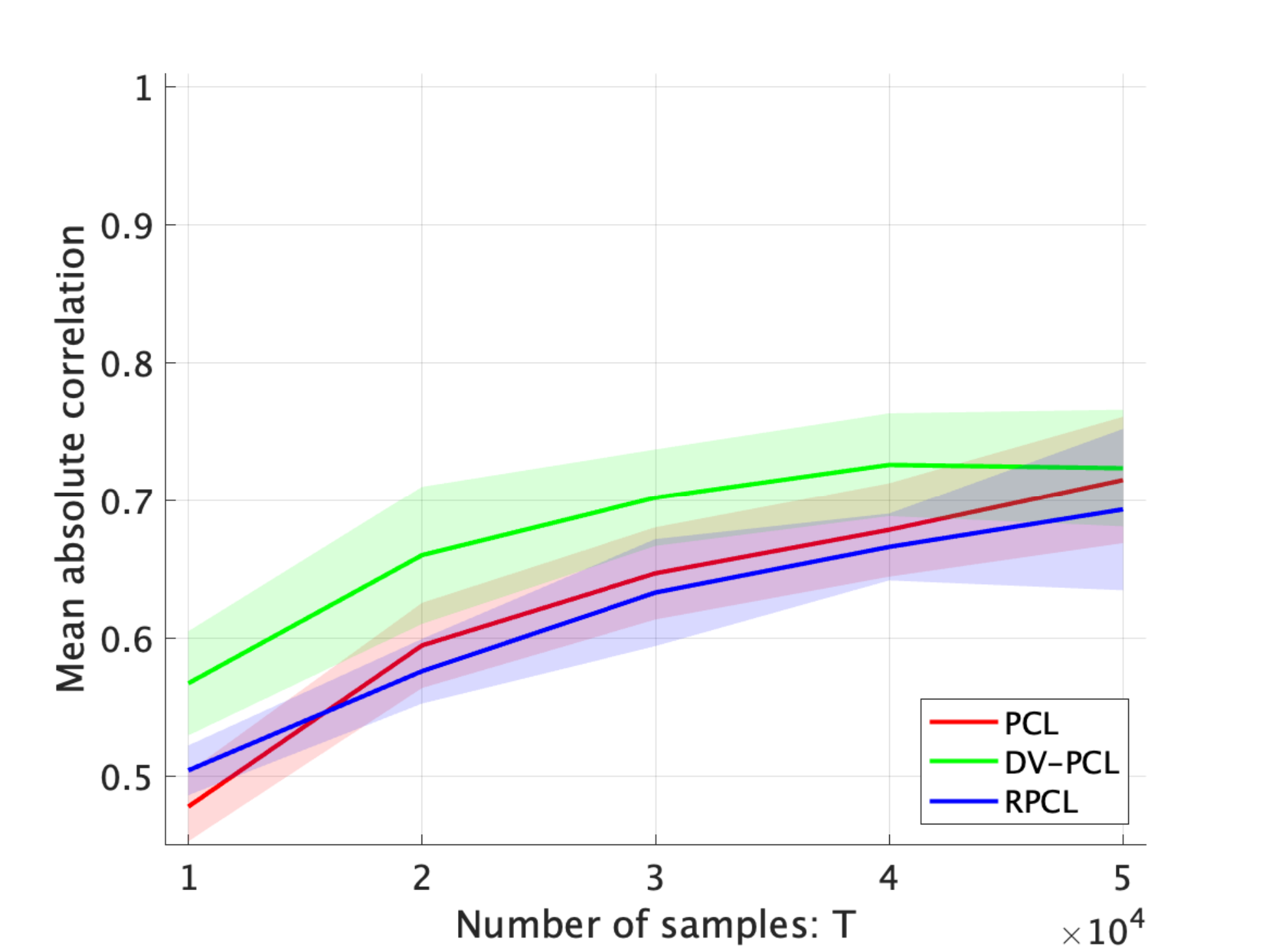}}
    \caption{\label{fig:art-sam} Sample efficiency of nonlinear ICA
    methods in the case of no outliers (i.e., $\epsilon=0$).  The
    averages of the mean absolute correlation were computed over $10$
    runs.  $L$ denotes the number of layers both in the nonlinear mixing
    function $\bm{f}$ in~\eqref{ICA-model} and representation function
    $\bhx$.}
   \end{figure}

   Fig.\ref{fig:art-sam}(a-c) investigate the sample efficiency of each
   method when $\epsilon=0$ (i.e., no outliers) and the numbers of
   layers are one, two and three, respectively. When the number of
   samples is large, all methods well-recover the source components and
   work similarly. However, when the number of samples gets smaller,
   DV-PCL tends to perform better than PCL and RPCL.  This presumably
   because the objective function $\Jdv$ in DV-PCL is based on the
   KL-divergence, and the density ratio estimator based on the
   KL-divergence would be accurate as implied
   in~\citet{kanamori2010theoretical}. The same tendency can be seen
   when the numbers of layers are four (Fig.\ref{fig:art-sam}(d)) and
   five (Fig.\ref{fig:art-sam}(e)).
      
\begin{figure}[t]
    \centering
    \subfigure[$L=1$]{\includegraphics[width=.32\textwidth]{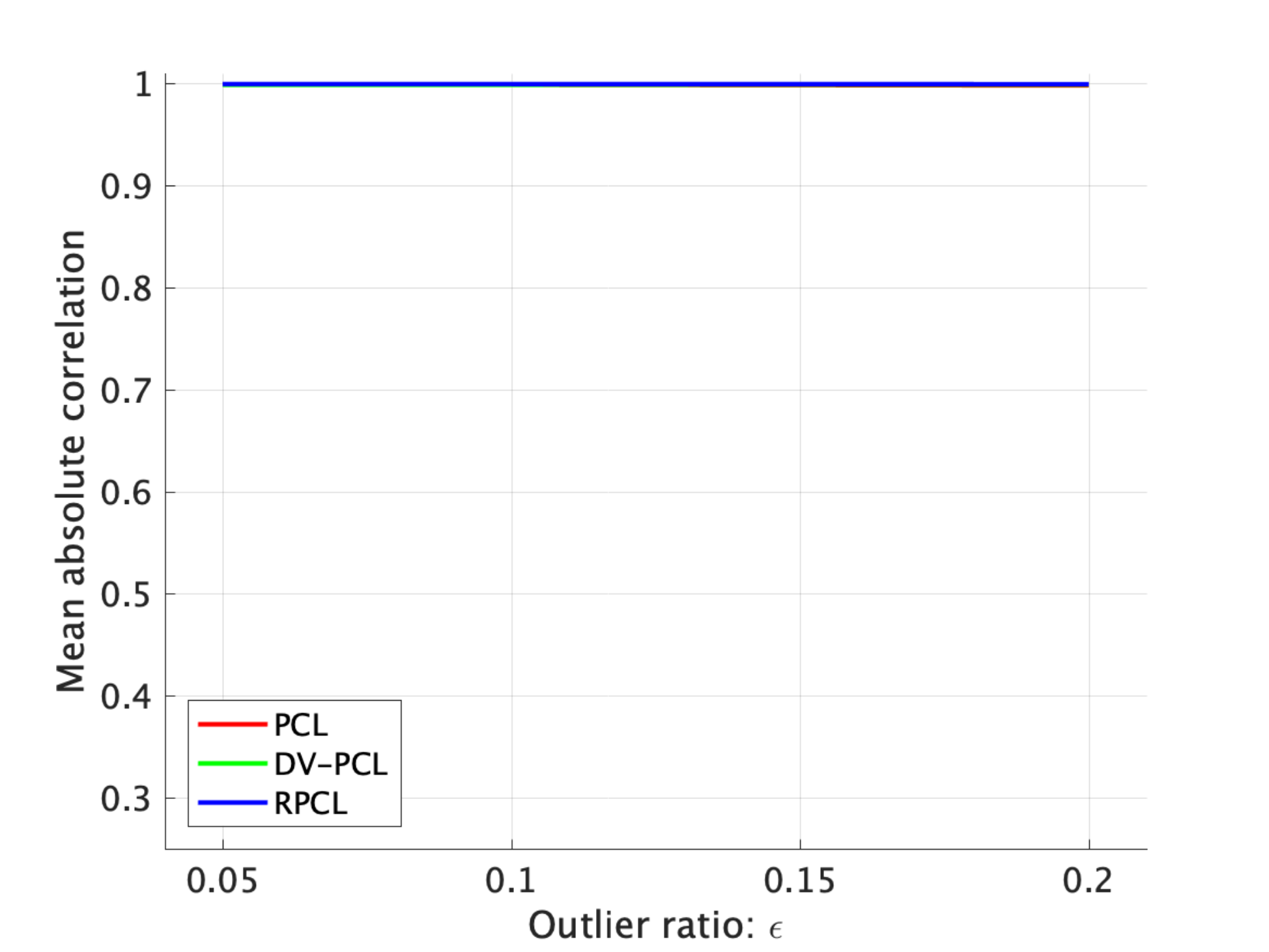}}
    \subfigure[$L=2$]{\includegraphics[width=.32\textwidth]{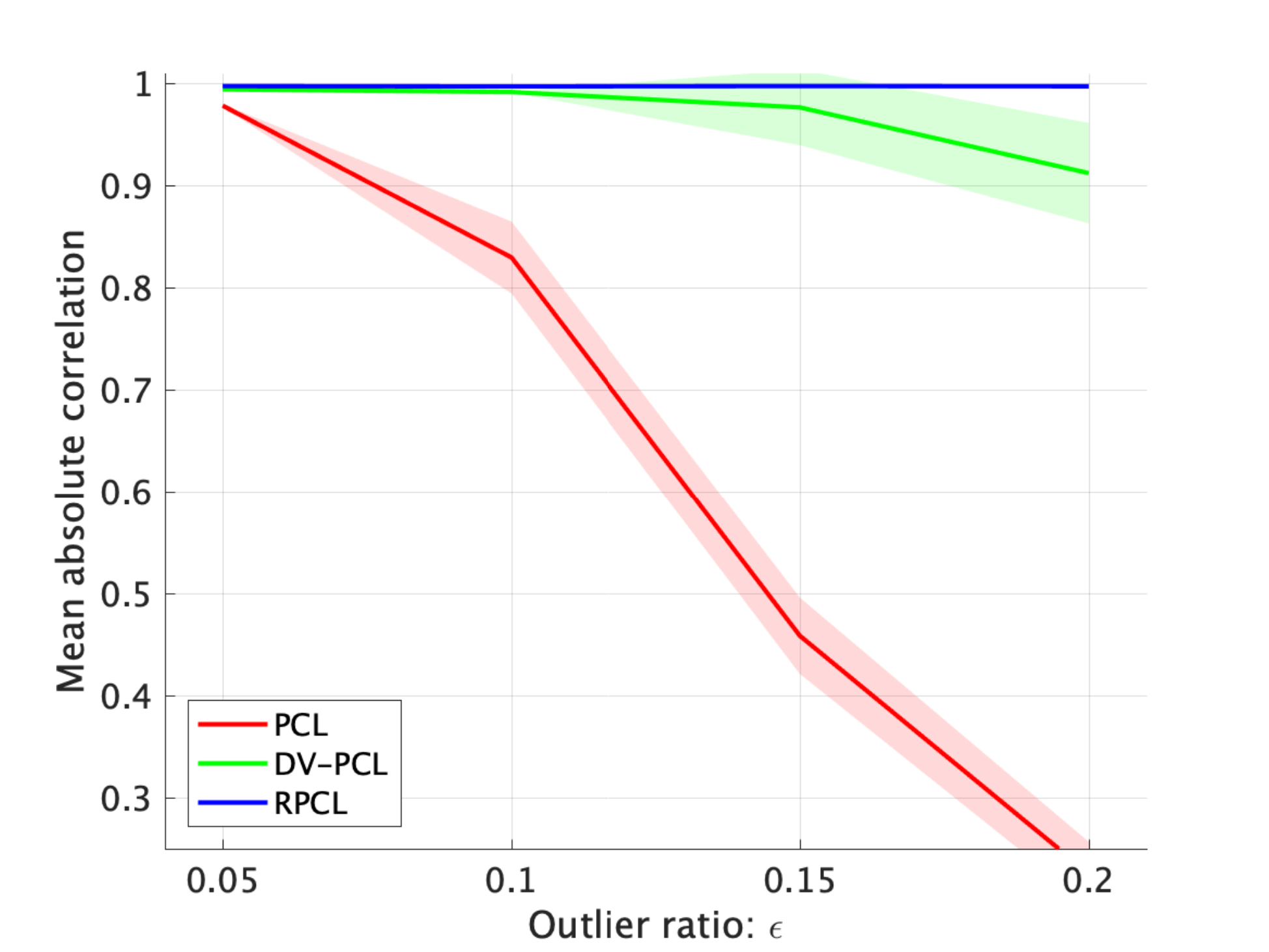}}
    \subfigure[$L=3$]{\includegraphics[width=.32\textwidth]{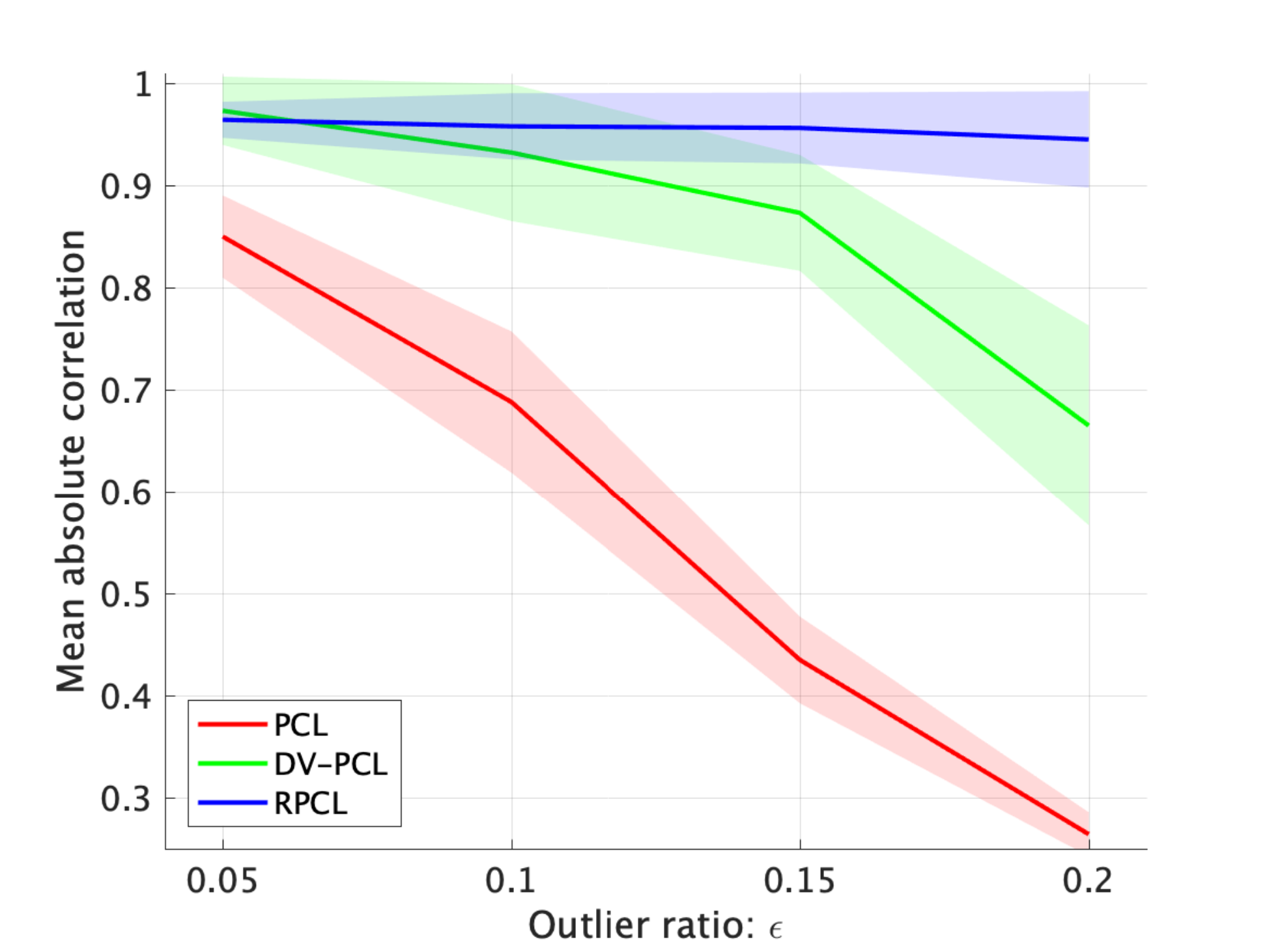}}
    \subfigure[$L=4$]{\includegraphics[width=.32\textwidth]{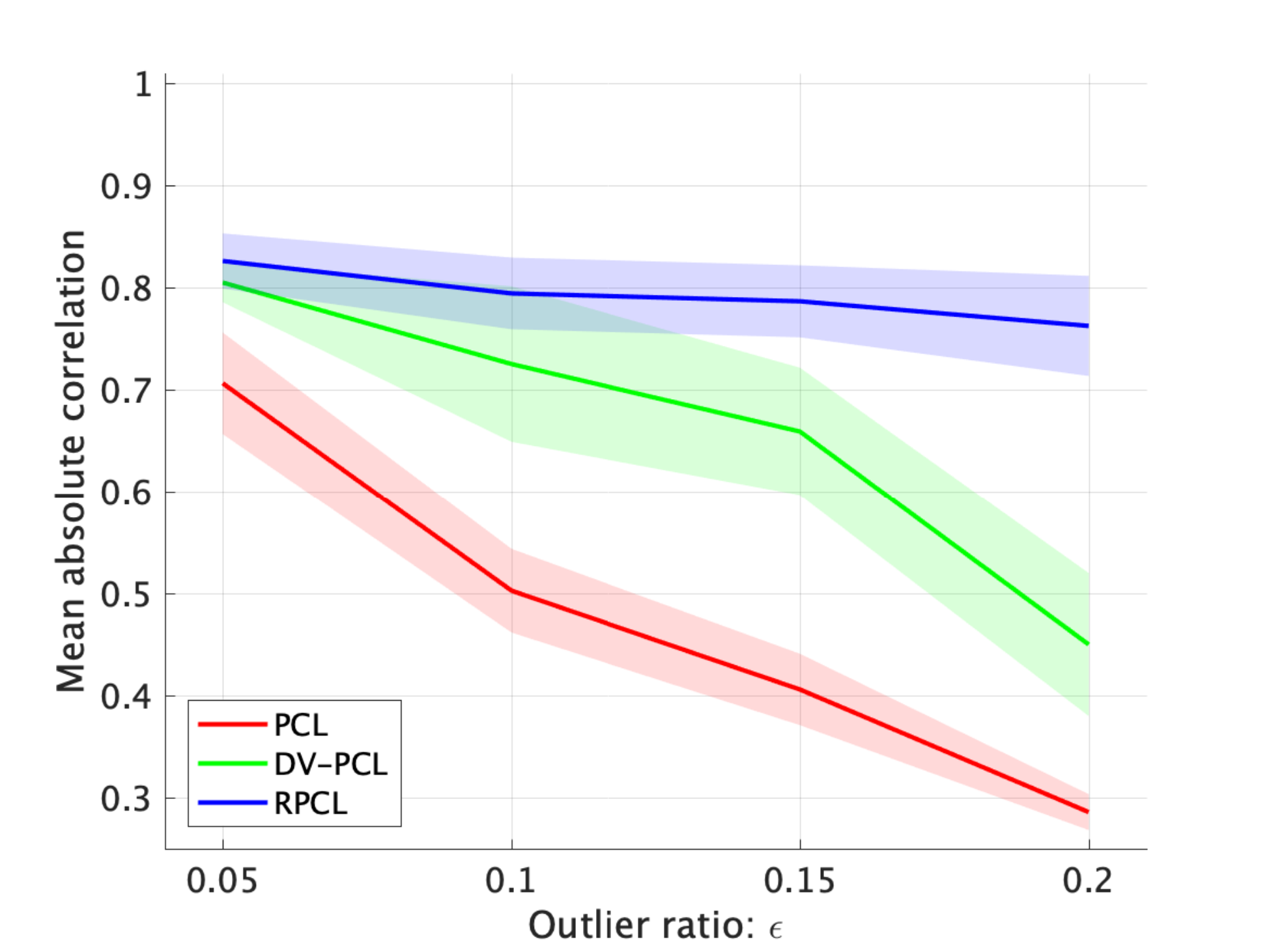}}
    \subfigure[$L=5$]{\includegraphics[width=.32\textwidth]{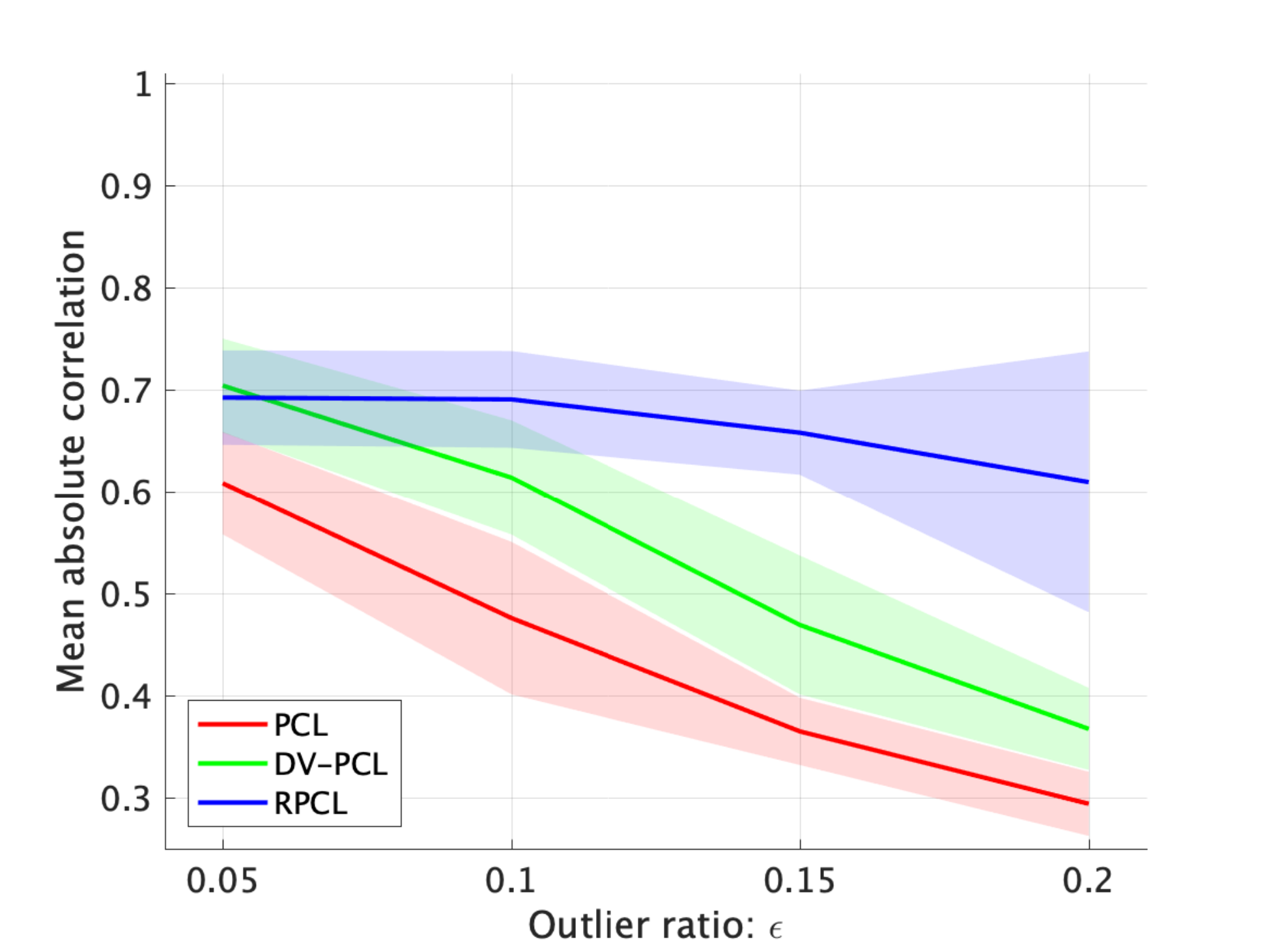}}
    \caption{\label{fig:art-out} Outlier robustness of nonlinear ICA
    methods. The averages of the mean absolute correlation were computed
    over $10$ runs. $L$ denotes the number of layers both in the
    nonlinear mixing function $\bm{f}$ in~\eqref{ICA-model} and
    representation function $\bhx$.}
   \end{figure}

   Fig.\ref{fig:art-out} shows outlier-robustness of each nonlinear ICA
   method in $T=100,000$. When the number of layers is one, all methods
   perform well (Fig.\ref{fig:art-out}(a)). This would be because we
   perform an outlier-robust whitening based on the $\gamma$-cross
   entropy~\citep{chen2013robust} in preprocessing. However, when the
   number of layer is larger than one, the performance of DV-PCL and PCL
   quickly decreases as the contamination ratio $\epsilon$ is increased
   (Fig.\ref{fig:art-out}(b-e)). On the other hand, RPCL keeps high
   correlation values on a wide range of $\epsilon$, and thus is very
   robust against outliers.  These results are consistent with the
   theoretical results in Section~\ref{ssec:analysis-outlier} that
   estimators based on $\Jdv$ (DV-PCL) and $\Jlr$ (PCL) can be seriously
   hampered by outliers, while the $\gamma$-cross entropy (RPCL) is
   promising in the presence of outliers.
   
   Overall, the proposed methods, DV-PCL and RPCL, have respective
   advantages over PCL. DV-PCL seems to be more useful in the limited
   number of samples, while RPCL is very robust against outliers
   particularly when the contamination ratio $\epsilon$ is large.
   \subsubsection{Importance of the dimensionality of complementary data}
   \label{sssec:high-aug-var}
    \begin{figure}[t]
     \centering
     \subfigure{\includegraphics[width=.4\textwidth]{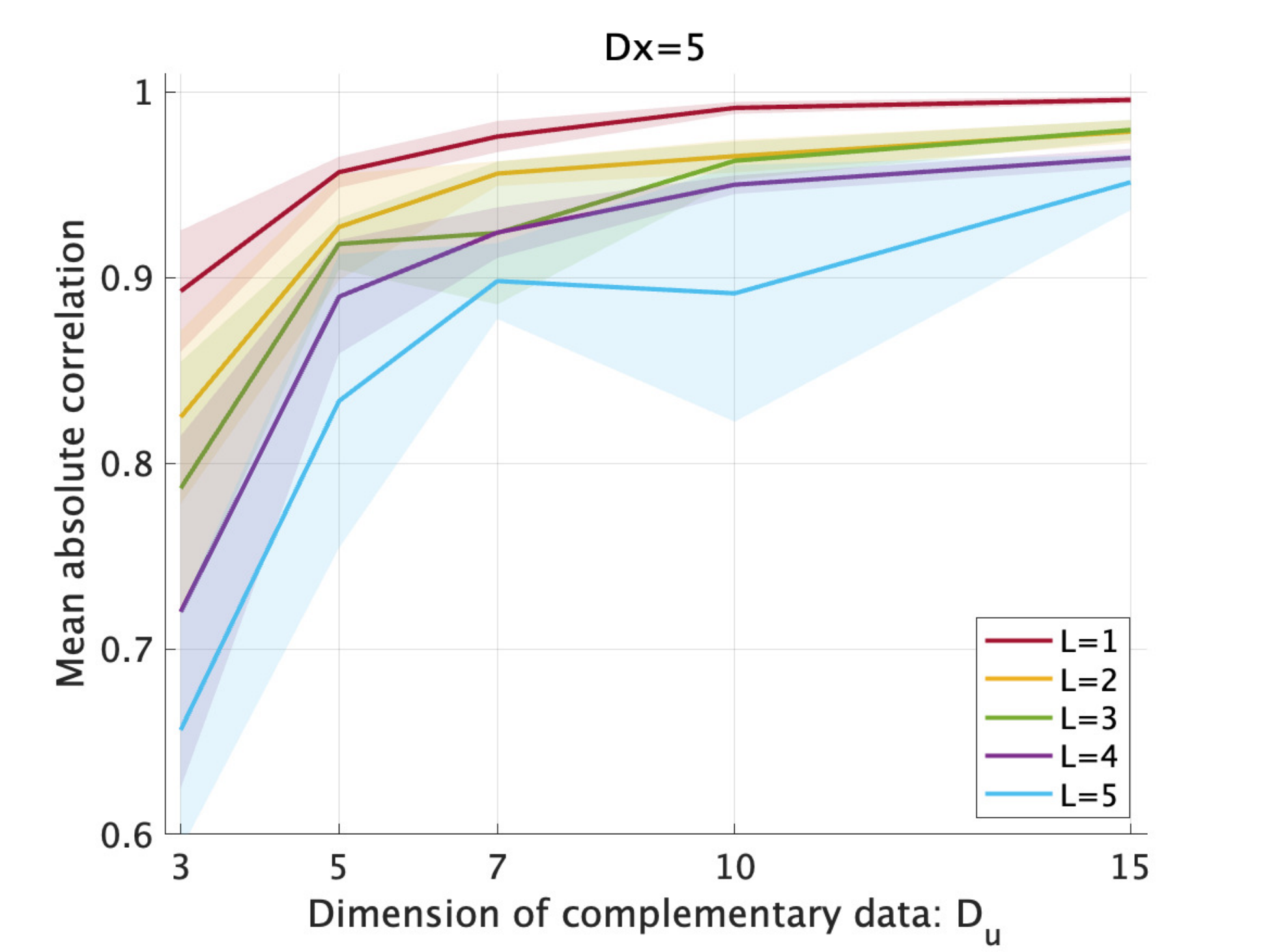}}
     \subfigure{\includegraphics[width=.4\textwidth]{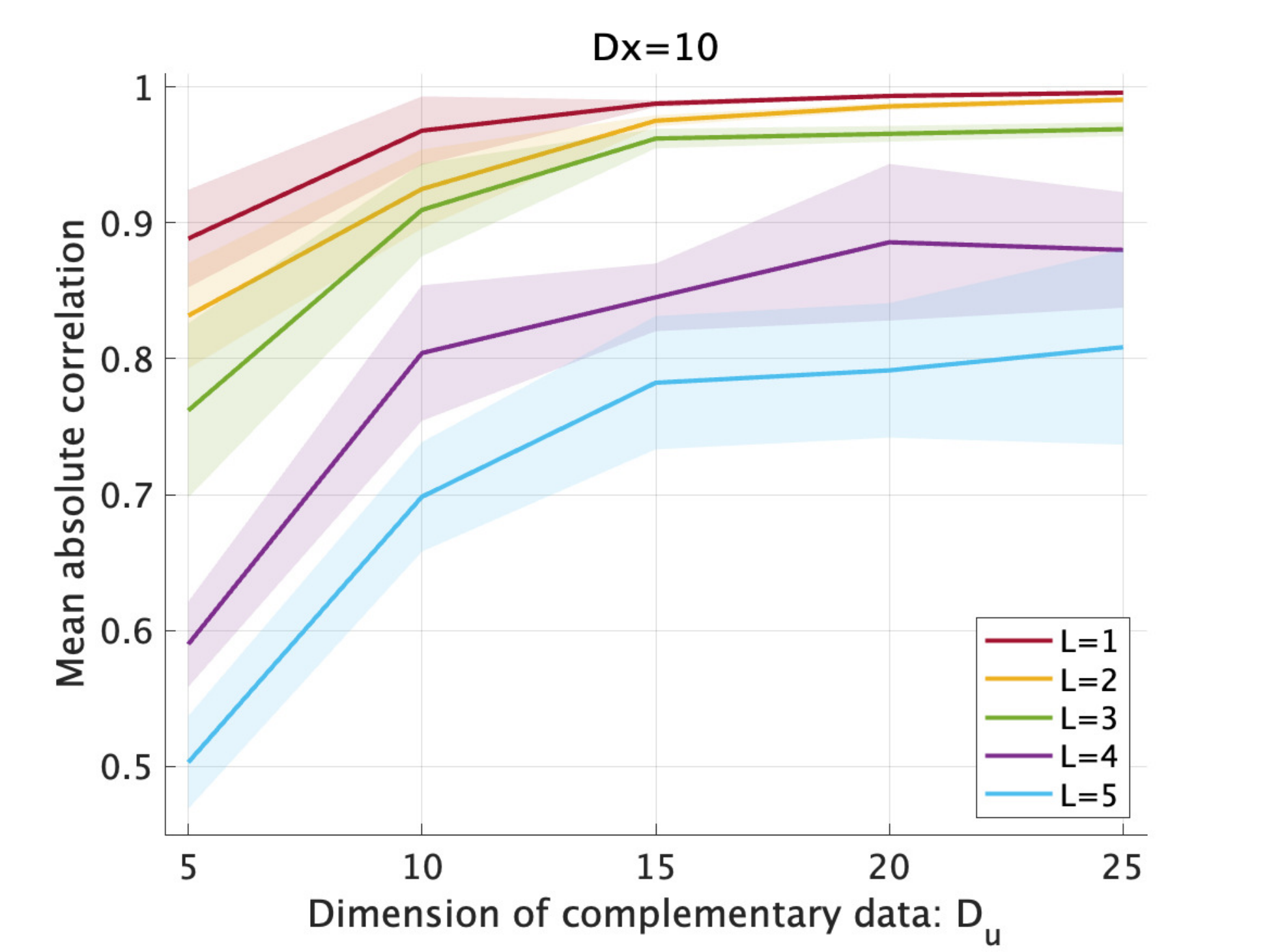}}
     \subfigure{\includegraphics[width=.4\textwidth]{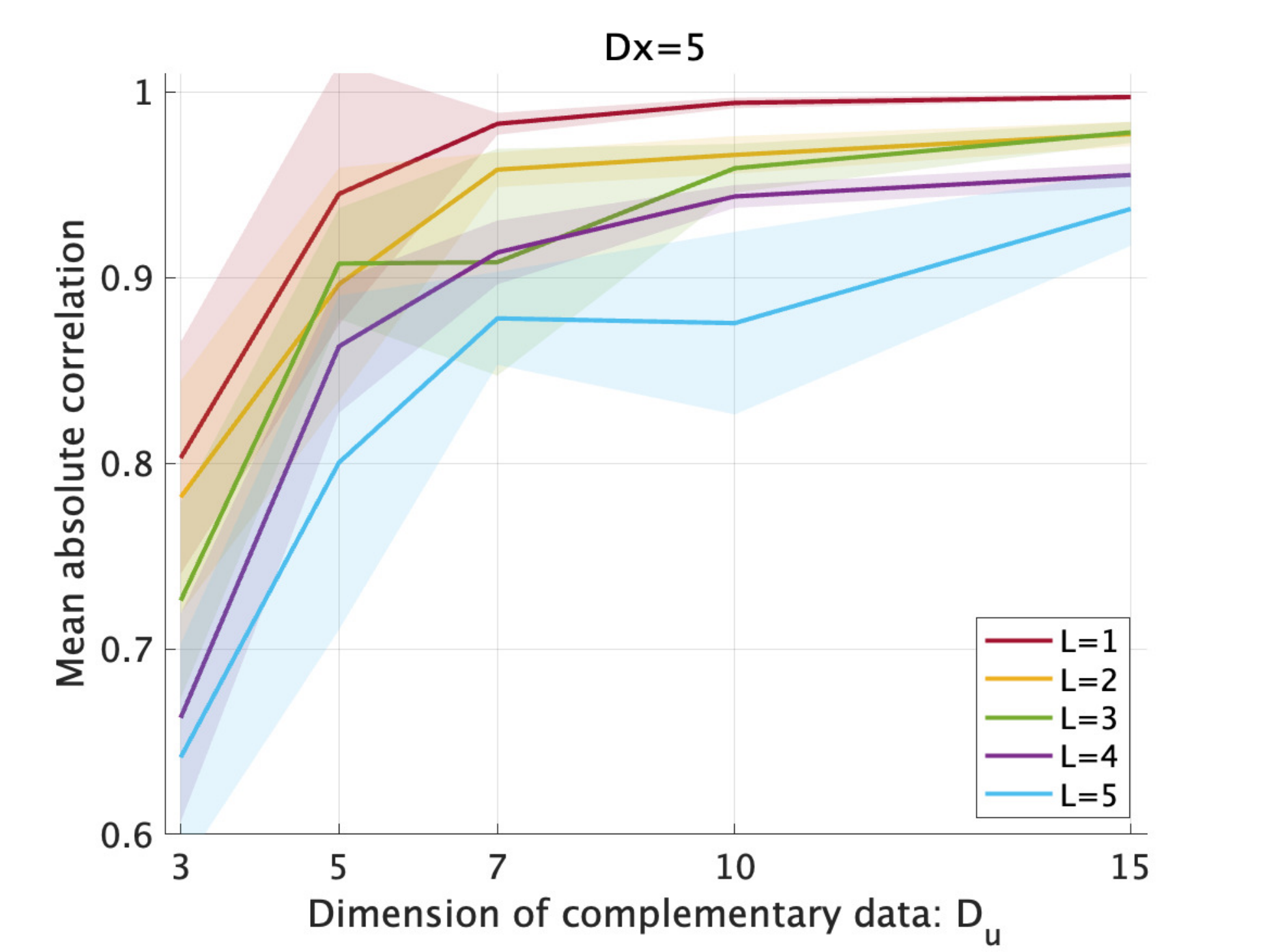}}
     \subfigure{\includegraphics[width=.4\textwidth]{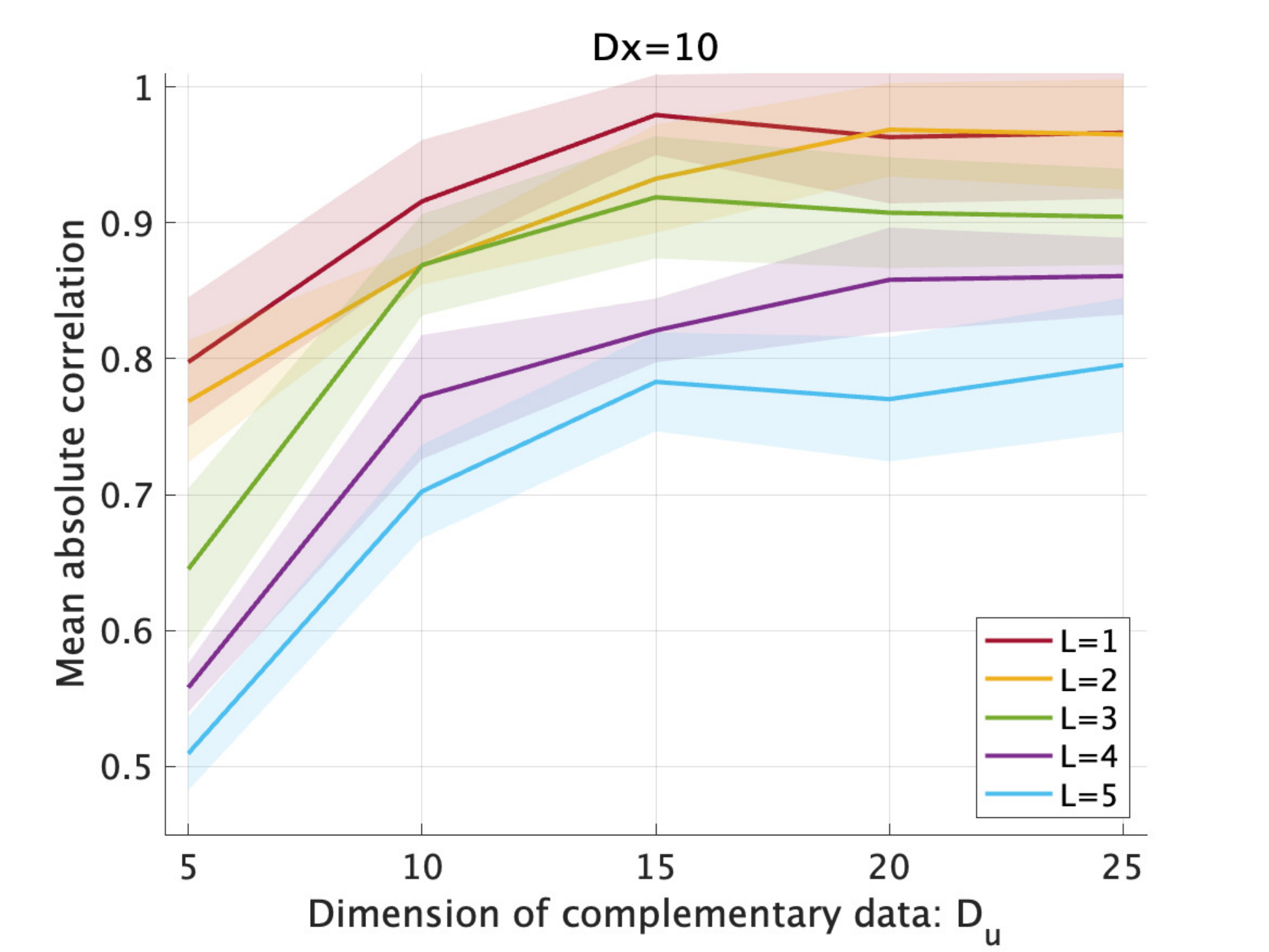}}
     \caption{\label{fig:dimaug} The averages of the mean absolute
     correlation against the dimensionality of complementary data over
     $10$ runs. The left and right plots in the first (second) row are
     $\Dx=5$ and $\Dx=10$ for Donsker-Varadhan variational estimation
     ($\gamma$-cross entropy), respectively.  $L$ denotes the number of
     layers both in the nonlinear mixing function $\bm{f}$
     in~\eqref{ICA-model} and representation function $\bhx$.}
    \end{figure}
   Next, we investigate how the dimensionality of complementary data
   $\bu$ affects the source recovery in nonlinear ICA as implied in
   Proposition~\ref{theo:general-ICA} and
   Theorem~\ref{prop:general-PCL}.
   
   We followed the recovery conditions in
   Theorem~\ref{prop:general-PCL}. In order for the conditional density
   $p(\bm{s}|\bm{u})$ to be differentiable, the conditionally
   independent sources were first generated from
   \begin{align}
    p(\bm{s}|\bm{u})\propto\prod_{i=1}^{\Dx}\exp\left(-
    \log(\cosh(s_i-\bm{w}_i^{\mathrm{s}\top}\bm{u}))\right),
   \label{exp:cond-source2}
   \end{align}
   where $\bm{w}^{\mathrm{s}}_i,~i=1\dots,\Dx$ are $\Du$-dimensional
   vectors randomly determined from the independent uniform density on
   $[-1,1]^{\Du}$. The function $\log(\cosh(\cdot))$ is a smooth
   approximation of the absolute function $|\cdot|$, and thus
   $p(\bm{s}|\bm{u})$ can be regarded as a smoothed version of the
   Laplace density~\eqref{exp:cond-source} in
   Section~\ref{sssec:eff-robust}. This smooth approximation has been
   previously used in linear ICA as well~\citep{hyvarinen1999fast}. In
   this experiment, complementary data samples $\bm{u}(t)$ were simply
   drawn from the independent uniform density on $[0,1]^{\Du}$. The
   total number of samples is $T=100,000$. Input data $\bm{x}$ was
   generated according to~\eqref{ICA-model} where the mixing function
   $\bm{f}$ is modelled by a feedforward neural network with random
   connections.
   
   Here, we used the nonlinear ICA methods based on the $\gamma$-cross
   entropy $\wJga$ (Section~\ref{ssec:RCL-gamma}) and Donsker-Varadhan
   variational estimation $\wJdv$ (Section~\ref{ssec:RCL-DV}). As in
   Section~\ref{sssec:eff-robust}, the representation function $\bhx$
   was modelled by a feedforward neural network where the number of
   hidden units was $4\Dx$, but the final layer was $\Dx$.  The number
   of layers was the same as $\bm{f}$ in the data generative
   model. $\bhu(\bu)$ was modelled by a one-layer neural network without
   the activation function as
   $\bhu(\bu)=\bm{W}_{\mathrm{u}}\bu+\bm{b}_{\mathrm{u}}$ where
   $\bm{W}_{\mathrm{u}}\in\R{\Dx\times\Du}$ and
   $\bm{b}_{\mathrm{u}}\in\R{\Dx}$. Since $p(\bm{s}|\bm{u})$
   in~\eqref{exp:cond-source2} is a smoother density, $r(\bx,\bu)$ was
   also modelled by a smother function than~\eqref{psi-model} in
   Section~\ref{sssec:eff-robust} as follows:
   \begin{align*}
    r(\bm{x},\bm{u})=\sum_{i=1}^{\Dx}
    \log(\cosh(a_{i,1}h_{\mathrm{x},i}(\bm{x})+a_{i,2}h_{\mathrm{u},i}(\bm{u})+b_i))
    -(\bar{a}_{i}\hx{i}(\bm{x})+\bar{b}_i)^2+c.
   \end{align*}
   All parameters were optimized by Adam for $1,500$ epochs with
   mini-batch size $256$ and learning rate $0.001$.  The $\ell_2$
   regularization was also applied as done in
   Section~\ref{sssec:eff-robust}. For the $\gamma$-cross entropy, the
   value of $\gamma$ was increased from $0.1$ to $3.0$ during minibatch
   stochastic gradient at every $100$ epoch.  The performance was
   evaluated by the mean absolute correlation as in
   Section~\ref{sssec:eff-robust}.

   The two plots in the first row of Fig.\ref{fig:dimaug} are results
   for Donsker-Varadhan variational estimation, and clearly show that
   the performance for source recovery depends on the dimensionality of
   complementary data. When $\Du<\Dx$, the mean correlation is small in
   all layers. However, when $\Du$ is larger than or equal to $\Dx$, the
   mean correlation gets significantly larger. This is consistent with
   implication of Theorem~\ref{prop:general-PCL}: In order to recover
   the source components, the dimensionality of complementary data is
   larger than or equal to input data (i.e., $\Dx\leq\Du$ in
   Assumption~(B$'$3)), which has not been revealed in previous work of
   nonlinear ICA. These empirical results clearly support our
   theoretical implications, and suggest to use fairly high-dimensional
   complementary data in practice. Similar results were observed for the
   $\gamma$-cross entropy as well (the second row of
   Fig.\ref{fig:dimaug}).  Furthermore, especially for Donsker-Varadhan
   variational estimation, another interesting point is that
   higher-dimensional complementary data often decreases the variance of
   the mean absolute correlation, and this implies that
   higher-dimensional complementary data takes a role of stabilizing
   estimation as well.
   \subsection{Evaluation on downstream linear classification}
   \label{ssec:linear-classification-exp}
   \begin{figure}[t]
    \centering \includegraphics[width=.4\textwidth]{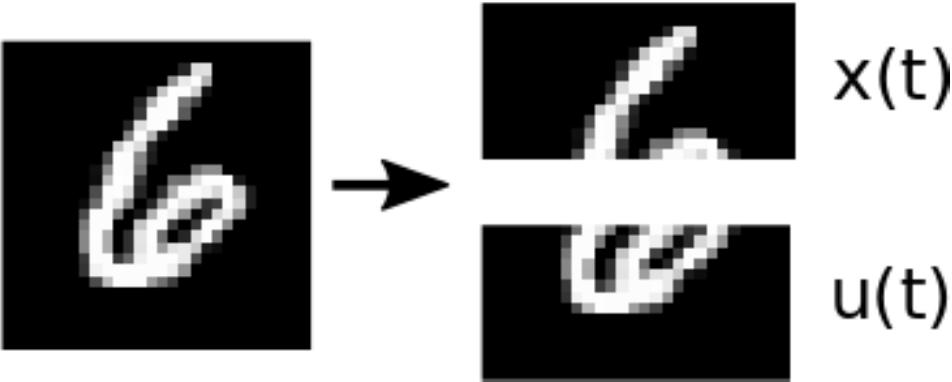}
    \hspace{10mm}
    \includegraphics[width=.4\textwidth]{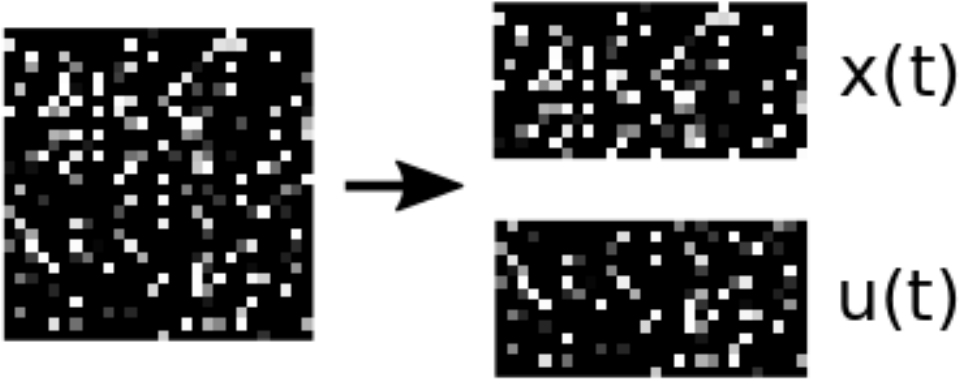}
    \caption{\label{fig:ex-mnist} An example of input and complementary
    data. The half upper images are used for input data samples, while
    the half below images are complementary ones. The right images are
    randomly shuffled versions of left images used as outliers.}
    \centering 
    \includegraphics[width=\textwidth]{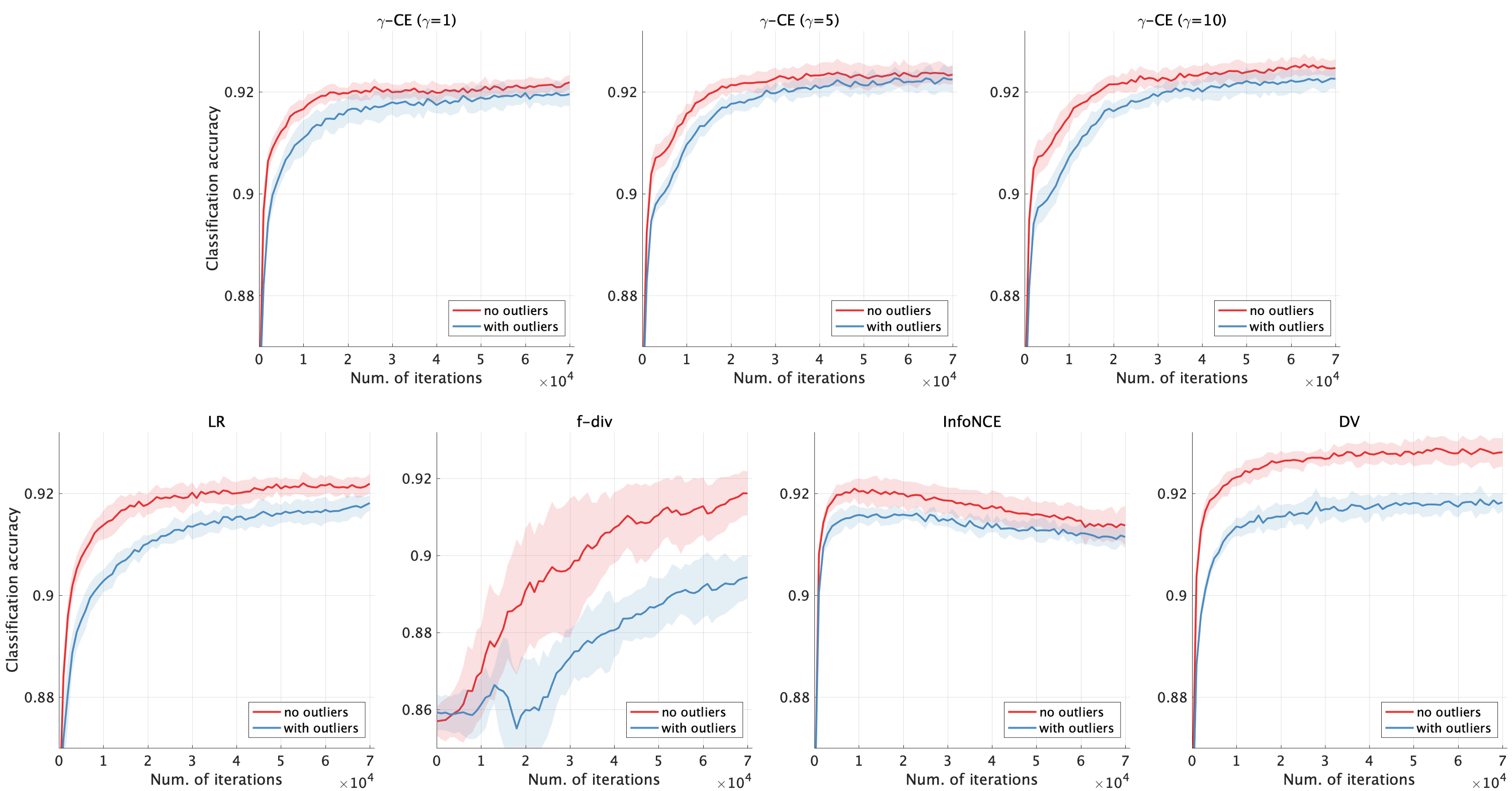}
    \caption{\label{fig:mnist} Averaged classification accuracy for MNIST
    over $10$ runs. For the red lines, training is performed without
    outliers, while the blue lines are training with outliers. Note that
    the scale of the vertical axis only for $f$-div is different.}
  \end{figure}  
  
  \begin{figure}[t]
   \centering
   \includegraphics[width=\textwidth]{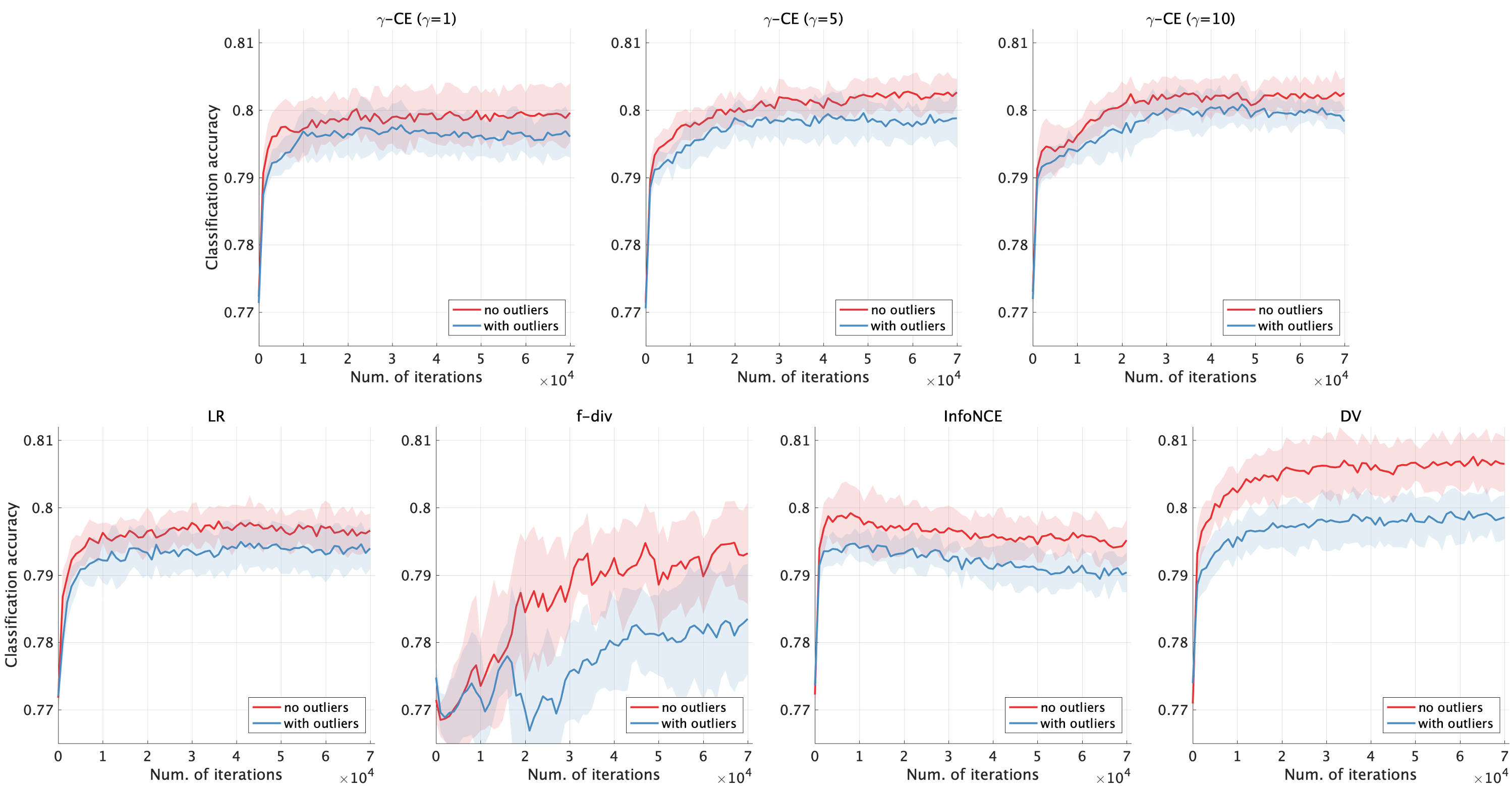}
   \caption{\label{fig:fmnist} Averaged classification accuracy for
   Fashion-MNIST over $10$ runs.}
  \end{figure}  
  \begin{figure}[t]
   \centering 
   \includegraphics[width=\textwidth]{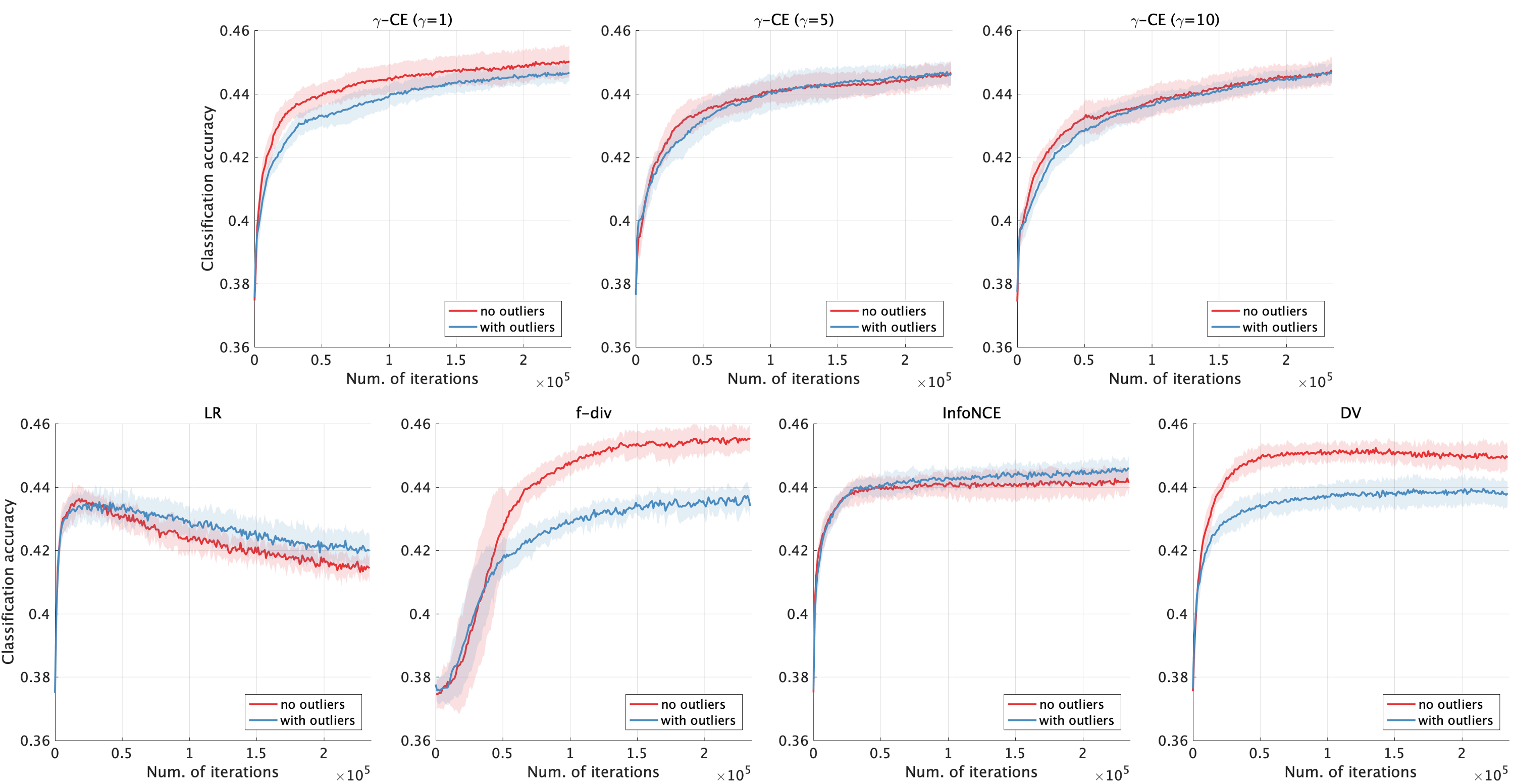}
   \caption{\label{fig:cifar10} Averaged classification accuracy for
   CIFAR-10 over $10$ runs.}
  \end{figure}  

  \renewcommand{\arraystretch}{1}
  \begin{table}[t]
   \begin{center}
    \caption{\label{tab:accuracy} Averages of classification accuracy
    for the last $5,000$ iterations over $10$ runs. Since we recorded
    classification accuracy at every $1,000$ iterations, each method
    includes the $50$ points of the accuracy over $10$ runs. The numbers
    in parentheses indicate standard deviations.  The best and
    comparable methods judged by the t-test at the significance level
    1\% are described in boldface.}
    \vspace{3mm}
    \begin{tabular}{c|c|c|c|c|c}
     \hline $\gamma$-CE ($\gamma=5$) & $\gamma$-CE ($\gamma=10$) & LR & f-div. & InfoNCE & DV  \\ \hline 
     \multicolumn{6}{l}{MNIST}\\
     0.924(0.002) & 0.925(0.002) & 0.921(0.002) & 0.915(0.006) & 0.914(0.003) & {\bf 0.928(0.003)}\\ \hline
     \multicolumn{6}{l}{MNIST \emph{with outliers}}\\
     {\bf 0.922(0.002)} & {\bf 0.922(0.002)} & 0.917(0.002) & 0.893(0.006) & 0.911(0.002) & 0.918(0.002)\\ \hline
     \multicolumn{6}{l}{Fashion-MNIST}\\
     0.802(0.003) & 0.802(0.003) & 0.796(0.003) & 0.794(0.006) & 0.795(0.003) & {\bf 0.807(0.004)}\\ \hline
     \multicolumn{6}{l}{Fashion-MNIST \emph{with outliers}}\\
     {\bf 0.799(0.003)} & {\bf 0.799(0.002)} & 0.794(0.003) & 0.783(0.008) & 0.791(0.003) & {\bf 0.799(0.003)}\\ \hline
     \multicolumn{6}{l}{CIFAR-10}\\
0.446(0.003) & 0.447(0.004) & 0.414(0.004) & {\bf 0.455(0.004)} & 0.442(0.004) & 0.449(0.004)\\ \hline
     \multicolumn{6}{l}{CIFAR-10 \emph{with outliers}}\\
{\bf 0.446(0.003)} & {\bf 0.446(0.003)} & 0.420(0.005) & 0.436(0.004) & {\bf 0.445(0.003)} & 0.438(0.004)\\ \hline
    \end{tabular}
   \end{center}
  \end{table}

  We finally demonstrate how the proposed method based on the
  $\gamma$-cross entropy works on benchmark datasets as done in the
  context of maximization of mutual
  information~\citep{tschannen2019mutual}, and implicitly investigate
  Theorem~\ref{theo:manifold} as well because the representation
  function $\bhx(\bm{x})$ outputs a lower-dimensional feature than input
  data.  In order to evaluate methods for unsupervised representation
  learning, a number of protocols have been previously proposed:
  Multi-scale structural similarity~\citep{wang2003multiscale}, mutual
  information estimation and see more protocols
  in~\citet{hjelm2019learning}. Here, we employ a linear classification
  protocol~\citep{oord2018representation,tian2019contrastive,tschannen2019mutual},
  which consists of two steps: First, a representation function $\bhx$
  is learned with unlabelled data. Second, the learned representation
  function is fixed (i.e., not learned anymore), and the feature
  computed through the representation function is tested on a downstream
  linear classification task using labelled data. The classification
  accuracy is the measure for goodness of the representation function.
  
  More specifically, we followed the experimental protocol
  in~\citet{tschannen2019mutual} \footnote{We slightly modified the
  python codes available
  at~\url{https://github.com/google-research/google-research/tree/master/mutual_information_representation_learning}.},
  which has been used in the context of deep canonical correlation
  analysis as well~\citep{andrew2013deep}. We first divided a single
  image in half, and then used the upper and lower half images as input
  $\bx(t)$ and complementary data samples $\bu(t)$, respectively (Left
  figures in Fig.~\ref{fig:ex-mnist}). Based on these data samples
  $\{\bx(t),\bu(t)\}_{t=1}^T$, we applied five methods based on the
  following objective functions for representation learning:
  \begin{itemize}
   \item \emph{$\gamma$-cross entropy ($\gamma$-CE)} $\tJga$: In order
	 for initialization, we first updated the parameters in
	 $r(\bx,\bu)$ with $\gamma=0.1$ for ten epochs, and then used
	 the updated parameters as the initial parameters for the
	 $\gamma$-cross entropy with a larger $\gamma$ value.

   \item \emph{Logistic regression (LR) in~\eqref{logistic}}: LR can be
	 seen as the limit of $\gamma\to{0}$ in the $\gamma$-cross
	 entropy.

   \item \emph{Variational estimation of f-divergence ($f$-div)
	 in~\eqref{f-div}~\citep{nguyen2008estimating,sugiyama2008direct}}.

   \item \emph{InfoNCE}~\citep{oord2018representation}:
	 \begin{align}
	  \tJnce(r):=E\left[\frac{1}{K}\sum_{i=1}^K\log\frac{e^{r(\bx_i,\bu_i)}}
	  {\frac{1}{K}\sum_{j=1}^Ke^{r(\bx_i,\bu_j)}}\right],
	  \label{infonce}
	 \end{align}    
	 where the expectation is taken over $\prod_{i=1}^{K} p(\bx_i,
	 \bu_i)$. In practice, we estimate~\eqref{infonce} by averaging
	 over multiple (mini-)batches of
	 samples~\citep{tschannen2019mutual}.

   \item \emph{Donsker-Varadhan variational estimation of mutual
	 information (DV)}
	 $\tJdv$~\citep{ruderman2012tighter,belghazi2018mutual}: Unlike
	 nonlinear ICA, this experiment excludes DV from the proposed
	 methods because DV has been already used for maximization of
	 MI~\citep{hjelm2019learning}.
  \end{itemize}  
  
  For all methods, we used the same model $r(\bx,\bu)$ as
  \begin{align*}
   r(\bx,\bu)=\psi(\bhx(\bx),\bhu(\bu))+a(\bhx(\bx))+b(\bhu(\bu)).
  \end{align*}
  The representation functions of $\bhx$ and $\bhu$ were modelled by the
  same neural architecture without parameter sharing: The first two
  hidden layers were convolution layers with the ReLU activation
  function and the third layer, which is the output layer, was a fully
  connected layer without the activation function. Before the output
  layer, we sequentially applied layer normalization~\citep{ba2016layer}
  and average pooling. The output dimensions were fixed at
  $\dx=\du=200$.  $a(\bhx)$ and $b(\bhu)$ were modelled by a one-layer
  feedforward network without activation functions.  By
  following~\citet{oord2018representation}
  and~\citet{tian2019contrastive}, we set
  $\psi(\bhx,\bhu)=\bhx^{\top}\bm{W}\bhu$ where $\bm{W}$ is a $\dx$ by
  $\du$ matrix and learned from data. We optimized all parameters by the
  Adam optimizer~\citep{kingma2015adam}.  After estimating the
  representation function, $\bhx(\bx)$ was fixed and not learned anymore
  on the evaluation phase.  As evaluation, we only trained a linear
  classifier to the learned features $\{\bhx(\bx(t))\}_{t=1}^T$ based on
  multinomial logistic regression. Classification accuracy on test data
  was used as the evaluation metric.

  As datasets, we used the following three classification datasets with
  ten classes\footnote{All datasets were downloaded through the
  tensorflow library.}:
  \begin{itemize}
   \item \emph{MNIST} ($\Dx=392, \Du=392$ and $T=60,000$)
	 
   \item \emph{Fashion-MNIST} ($\Dx=392, \Du=392$ and $T=60,000$)

   \item \emph{CIFAR10} ($\Dx=1,536, \Du=1,536$ and $T=50,000$)
  \end{itemize}
  For MNIST and Fashion-MNIST, we used the learning rate $10^{-4}$ and
  updated the parameters for $300$ epochs, while the learning rate for
  CIFAR10 is $10^{-5}$ and the number of epochs is $1200$. In order to
  demonstrate the robustness to outliers, the training data was
  contaminated by randomly shuffled images with respect to pixels (Right
  figures in Fig.~\ref{fig:ex-mnist}). The contamination ratio of
  outliers was fixed at $0.3$.
  
  Classification accuracy of MNIST on test data samples over iterations
  is plotted in Fig.~\ref{fig:mnist}. In the case of no outliers (i.e.,
  red lines), DV performs the best, while the classification accuracy of
  the $\gamma$-CE and LR is fairly good. For InfoNCE, the classification
  accuracy around the $70,000$-th iteration is less than DV, LR and
  $\gamma$-CE, while $f$-div shows high variance of the classification
  accuracy over iterations. When data is contaminated by outliers (i.e.,
  blue lines), $f$-div, LR and DV show clear performance
  degeneration. On the other hand, the performance of $\gamma$-CE is not
  so influenced by the contamination of outliers. Interestingly,
  estimation based on InfoNCE is also not strongly hampered by outliers,
  but the classification accuracy is worse than $\gamma$-CE around the
  $70000$-th iteration.  The same tendency of the robustness of the
  $\gamma$-cross entropy was observed both for Fashion-MNIST and
  CIFAR-10 (Figs.~\ref{fig:fmnist}
  and~\ref{fig:cifar10}). Table~\ref{tab:accuracy} quantitatively
  indicates that the $\gamma$-cross entropy yields the best
  classification accuracy when data is contaminated by outliers, and
  shows a fairly good permanence even without outliers.  Thus, the
  proposed method based on the $\gamma$-cross entropy can be robust
  against outliers as implied in our theoretical analysis and fairly
  works well even when data is not contaminated by outliers.

  Finally, let us note that these results are not trivial because the
  connection between mutual information and quality of data
  representation has been experimentally demonstrated to be rather
  loose~\citep{tian2019contrastive}. Thus, it was unclear that robust
  estimation always leads to robust representations of
  data. Nonetheless, the proposed method based on the $\gamma$-cross
  entropy performed the best in the presence of outliers. This means
  that the proposed robust method for unsupervised representation
  learning is promising.
 \section{Conclusion}
 \label{sec:conclusion}
 This paper theoretically showed that density ratio estimation plays a
 key role in three frameworks for unsupervised representation learning:
 Maximization of mutual information and nonlinear independent component
 analysis as well as nonlinear subspace estimation, which is a novel
 framework proposed in this paper. Furthermore, we made theoretical
 contributions in each of the three frameworks: We showed that density
 ratio estimation is necessary and sufficient in maximization of mutual
 information, while our analysis revealed a novel insight for source
 recover in nonlinear ICA that the dimensionality of complementary data
 is an important factor for source recovery, which was clearly supported
 by numerical experiments. In addition, the proposed generative model in
 nonlinear subspace estimation is more general than nonlinear ICA in the
 sense that the latent source components are no longer assumed to be
 conditionally independent, and theoretical conditions to estimate a
 nonlinear subspace of the latent source components were
 given. Motivated by the theoretical results, we developed a nonlinear
 ICA method by applying an variation lower-bound of mutual information,
 and proposed an outlier-robust method for unsupervised representation
 learning through density ratio estimation. We also theoretically
 investigated their outlier-robustness. The usefulness of the proposed
 methods were demonstrated through numerical experiments of nonlinear
 ICA and linear classification in a downstream task.
 \acks{The authors would like to thank Dr. Hiroshi Morioka for sharing
 his python codes of permutation contrastive learning with us. Takashi
 Takenouchi was partially supported by JSPS KAKENHI Grant Number
 20K03753 and 19H04071.}
 \appendix
 \section{Proof of Theorem~\ref{prop:maxMI}}
 \label{app:maxMIproof}
 \begin{proof}
  We first recall that $\byx:=\bhx(\bx)$ and $\byu:=\bhu(\bu)$, and
  denote $\bhx^{\perp}(\bx)$ and $\bhu^{\perp}(\bu)$ by $\byx^{\perp}$
  and $\byu^{\perp}$, respectively. Based on Assumption~(A2),
  $(\byx,\byx^{\perp}):=(\bhx(\bx),\bhx^{\perp}(\bx))$ and
  $(\byu,\byu^{\perp}):=(\bhu(\bu),\bhu^{\perp}(\bu))$ are both
  invertible in the sense that there exist functions
  $\bm{g}_{\mathrm{x}}$ and $\bm{g}_{\mathrm{u}}$ such that
  $\bx=\bm{g}_{\mathrm{x}}(\byx,\byx^{\perp})$ and
  $\bu=\bm{g}_{\mathrm{u}}(\byu,\byu^{\perp})$. Based on the invertible
  functions, we first establish the following lemma, which decomposes
  $I(\calX,\calU)$ into three terms:
  \begin{lemma}
   \label{lem:decomposition-MI} Assumptions (A1-2) hold. Then,
   $I(\calX,\calU)$ can be decomposed as follows:
   \begin{align} 
    I(\calX,\calU)&=I(\calYx,\calYu)
    +E_{Y_{\mathrm{u}}}[I(\calYx|\calYu,\calYu^{\perp}|\calYu)]
    +E_{\mathrm{Y_{\mathrm{x}}}}[I(\calYx^{\perp}|\calYx,\calU|\calYx)],
    \label{MI-hy-hx}
   \end{align}
   where $E_{Y_{\mathrm{u}}}$ and $E_{\mathrm{Y_{\mathrm{x}}}}$ denote
   the expectations over $p(\byu)$ and $p(\byx)$ respectively,
   \begin{align*}        
    I(\calYx|\calYu,\calYu^{\perp}|\calYu)&\maketitle=\iint
    p(\byx,\byu^{\perp}|\byu)\log\frac{p(\byx,\byu^{\perp}|\byu)}
    {p(\byx|\byu)p(\byu^{\perp}|\byu)}\intd\byx\intd\byu^{\perp}\\
    I(\calYx^{\perp}|\calYx,\calU|\calYx)&:=\iint
    p(\byx^{\perp},\bm{u}|\byx)\log\frac{p(\byx^{\perp},\bm{u}|\byx)}
    {p(\byx^{\perp}|\byx)p(\bm{u}|\byx)}\intd\byx^{\perp}\intd\bm{u}.
   \end{align*}
  \end{lemma}
  The proof is given in Appendix~\ref{app:proof-decomposition-MI}.  The
  first term on the right-hand side of~\eqref{MI-hy-hx} is mutual
  information between the representation functions of $\bx$ and $\bu$,
  while $I(\calYx^{\perp}|\calYx, \calU|\calYx)$ and
  $I(\calYx|\calYu,\calYu^{\perp}|\calYu)$ measure the conditional
  independence and are equal to $0$ when $\bm{u}\perp\byx^{\perp}|\byx$
  and $\byx\perp\byu^{\perp}|\byu$. It follows from~\eqref{MI-hy-hx}
  that $I(\calX,\calU)=I(\calYx,\calYu)$ if and only if
  $I(\calYx|\calYu,\calYu^{\perp}|\calYu)=0$ and
  $I(\calYx^{\perp}|\calYx,\calU|\calYx)=0$.

  Next, we fix $\bhx$ and $\bhu$ at $\bhx^{\star}$ and $\bhu^{\star}$
  respectively, and substitute them as $\byx=\bhx^{\star}(\bx)$ and
  $\byu=\bhu^{\star}(\bu)$. Then, it is shown
  that~\eqref{univ-approx-MI} implies both
  $I(\calYx|\calYu,\calYu^{\perp}|\calYu)=0$ and
  $I(\calYx^{\perp}|\calYx,\calU|\calYx)=0$.  We first multiply
  $p(\bu)$ to both sides and rewrite~\eqref{univ-approx-MI} as
  \begin{align}
   p(\bm{u}|\bm{x})=p(\bm{u}|\byx), \label{SDRcond}
  \end{align}
  where with the partition function $Z(\byx):=\int
  \exp\left(\psi^{\star}(\byx,\bhu^{\star}(\bu))+a^{\star}(\byx)+b^{\star}(\bhu^{\star}(\bm{u}))\right)
  p(\bu)\intd\bm{u}$,
  \begin{align*}
   p(\bm{u}|\byx):=
  \frac{\exp\left(\psi^{\star}(\byx,\bhu^{\star}(\bm{u})))+a^{\star}(\byx)+b^{\star}(\bhu^{\star}(\bm{u}))\right)p(\bu)}
  {Z(\byx)}.
  \end{align*}
  By the invertible assumption for $(\byx,
  \byx^{\perp})$,~\eqref{SDRcond} can be equivalently expressed as the
  following conditional independence:
  \begin{align}
   \bm{u}\perp\byx^{\perp}|\byx,\label{cond-ind-hx}
  \end{align}   
  implying that
  \begin{align*}
   I(\calYx^{\perp}|\calYx,\calU|\calYx)=0.
  \end{align*}
  Similarly, we can derive the following equation
  from~\eqref{univ-approx-MI}:
  \begin{align}
   p(\bm{x}|\bm{u})=p(\bx|\byu), \label{SDRcond2}
  \end{align}  
  where with the partition function $Z(\byu):=\int
  \exp\left(\psi^{\star}(\bhx^{\star}(\bx),\byu)+a^{\star}(\bhx^{\star}(\bx))+b^{\star}(\byu)\right)
  p(\bx)\intd\bm{x}$,
  \begin{align*}
   p(\bx|\byu):=
   \frac{\exp\left(\psi^{\star}(\bhx^{\star}(\bx),\byu)+a^{\star}(\bhx^{\star}(\bx))+b^{\star}(\byu)\right)p(\bx)}
   {Z(\byu)}.
  \end{align*}  
  Again, \eqref{SDRcond2} implies the conditional independence as
  \begin{align}
   \bm{x}\perp\byu^{\perp}|\byu.  \label{cond-ind-hy}
  \end{align}
  Data processing inequality and conditional
  independence~\eqref{cond-ind-hy} lead to
  \begin{align*}
   I(\calYx|\calYu,\calYu^{\perp}|\calYu)\leq{I}(\calX|\calYu,\calYu^{\perp}|\calYu)
   \underset{\eqref{cond-ind-hy}}{=}0.
  \end{align*}
  Both $I(\calYx|\calYu,\calYu^{\perp}|\calYu)=0$ and
  $I(\calYx^{\perp}|\calYx,\calU|\calYx)=0$ enable us to ensure
  $I(\calX,\calU)=I(\calYx,\calYu)$.

  Conversely, we suppose that $I(\calX,\calU)=I(\calYx,\calYu)$ at
  $\bhx=\bhx^{\star}$ and $\bhu=\bhu^{\star}$. Then, it follows
  from~\eqref{MI-hy-hx} that $I(\calYx|\calYu,\calYu^{\perp}|\calYu)=0$
  and $I(\calYx^{\perp}|\calYx, \calU|\calYx)=0$.
  $I(\calYx^{\perp}|\calYx,\calU|\calYx)=0$ means the conditional
  independence~\eqref{cond-ind-hx}, which can be equivalently expressed
  as~\eqref{SDRcond}. On the other hand, under the invertibility of
  $(\byu,\byu^{\perp})$, $I(\calYx|\calYu,\calYu^{\perp}|\calYu)=0$
  implies
  \begin{align}
   \byx\perp\byu^{\perp}|\byu\quad
   \text{or equivalently}\quad p(\byx|\bu)=p(\byx|\byu). \label{SDRcond3}
  \end{align}
  A simple calculation yields
  \begin{align}
   p(\bx,\bu)&=p(\bu|\bx)p(\bx)
   \underset{\eqref{SDRcond}}{=}p(\bu|\byx)p(\bx)
   \underset{(\star)}{=}\frac{p(\byx|\bu)p(\bu)}{p(\byx)}p(\bx)
   \underset{\eqref{SDRcond3}}{=}\frac{p(\byx|\byu)}{p(\byx)}p(\bx)p(\bu),
   \label{conversion}
  \end{align}
  where we applied Bayes' theorem on $(\star)$. By dividing both sides
  of~\eqref{conversion} by $p(\bx)p(\bu)$, we have
  \begin{align*}
   \log\frac{p(\bx,\bu)}{p(\bx)p(\bu)} = 
   \log{p}(\byx, \byu)-\log{p}(\byx)-\log{p}(\byu).
  \end{align*}
  The existence of the functions is obvious by denoting $\log p(\byx,
  \byu)$, $-\log{p}(\byx)$ and $-\log{p}(\byu)$ by
  $\psi^{\star}(\byx,\byu)$, $a^{\star}(\byx)$ and $b^{\star}(\byu)$
  respectively. The proof is completed.
 \end{proof}

  \subsection{Proof of Lemma~\ref{lem:decomposition-MI}}
  \label{app:proof-decomposition-MI}
  \begin{proof}
   We first decompose $I(\calX,\calU)$ based on the invertibility
   assumption of $(\byx,\byx^{\perp})$ as follows:
   \begin{align}
    I(\calX,\calU)&=\iiint p(\byx,\byx^{\perp},\bm{u})
    \log\frac{p(\byx,\byx^{\perp},\bm{u})}{p(\byx,\byx^{\perp})p(\bm{u})}
    \intd\byx\intd\byx^{\perp}\intd\bm{u}\nonumber\\ &=\iint
    p(\byx,\bm{u})
    \log\frac{p(\byx,\bm{u})}
    {p(\byx)p(\bm{u})}\intd\byx\intd\bm{u}\nonumber\\
    &\qquad+\int{p}(\byx)\left[\iint p(\byx^{\perp},\bm{u}|\byx)
    \log\frac{p(\byx^{\perp},\bm{u}|\byx)}
    {p(\byx^{\perp}|\byx)p(\bm{u}|\byx)}
    \intd\bm{u}\intd\byx^{\perp}\right]\intd\byx\nonumber\\ &=
    I(\calYx,\calU)+E_{\mathrm{Y}_{\mathrm{x}}}[
    I(\calYx^{\perp}|\calYx,\calU|\calYx)],
    \label{MI1}
   \end{align}
   where we used the following relation:
   \begin{align*}
    \frac{p(\byx,\byx^{\perp},\bm{u})}
    {p(\byx,\byx^{\perp})p(\bm{u})}
    &=\frac{p(\byx^{\perp},\bm{u}|\byx)
    p(\byx)}{p(\byx^{\perp}|\byx)p(\byx)p(\bm{u})}
    \cdot\frac{p(\byx,\bm{u})}{p(\byx,\bm{u})}
    =\frac{p(\byx,\bm{u})}{p(\byx)p(\bm{u})}\cdot
    \frac{p(\byx^{\perp},\bm{u}|\byx)}
    {p(\byx^{\perp}|\byx)p(\bm{u}|\byx)}.
   \end{align*}
   Applying the same decomposition as~\eqref{MI1} to
   $I(\calYx,\mathcal{U})$ under the change of variables by the
   invertible $(\byu,\byu^{\perp})$ yields
   \begin{align}
    I(\calYx,\calU)&=\iiint{p}(\byx,\byu,\byu^{\perp})
    \log\frac{p(\byx,\byu,\byu^{\perp})}{p(\byx)p(\byu,\byu^{\perp})}
    \intd\byx\intd\byu\intd\byu^{\perp}\nonumber\\
    &=\iint{p}(\byx,\byu)
    \log\frac{p(\byx,\byu)}{p(\byx)p(\byu)}
    \intd\byx\intd\byu\nonumber\\
    &\qquad
    +\int{p}(\byu)\left[\iint p(\byx,\byu^{\perp}|\byu)
    \log\frac{p(\byx,\byu^{\perp}|\byu)}
    {p(\byu^{\perp}|\byu)p(\byx|\byu)}\intd\byx\intd\byu^{\perp}
    \right]\intd\byu\nonumber\\
    &=I(\calYx,\calYu)+E_{\mathrm{Y}_{\mathrm{u}}}
    [I(\calYx|\calYu,\calYu^{\perp}|\calYu)], \label{MI2}
   \end{align}
   where we used
   \begin{align*}
    \frac{p(\byx,\byu,\byu^{\perp})}{p(\byx)p(\byu,\byu^{\perp})}
    =\frac{p(\byx,\byu^{\perp}|\byu)p(\byu)}
    {p(\byx)p(\byu^{\perp}|\byu)p(\byu)}
    \cdot\frac{p(\byx,\byu)}{p(\byx,\byu)}
    =\frac{p(\byx,\byu)}{p(\byx)p(\byu)}\cdot
    \frac{p(\byx,\byu^{\perp}|\byu)}
    {p(\byu^{\perp}|\byu)p(\byx|\byu)}
   \end{align*}   
   Substituting~\eqref{MI2} into~\eqref{MI1} completes the proof.
  \end{proof}
 \section{Proof of Proposition~\ref{theo:general-ICA}}
 \label{app:general-ICA}
 \begin{proof}
  Let us denote the inverse of the mixing function $\bm{f}$ in the
  generative model~\eqref{ICA-model} as $\bm{g}:=\bm{f}^{-1}$.  From the
  conditional independence of $\bm{s}$ given $\bu$ in Assumption~(B1),
  we obtain the conditional density of $\bx$ given $\bu$ under the
  change of variables by $\bm{s}=\bm{g}(\bm{x})$ as follows:
  \begin{align}
   \log p(\bm{x}|\bu)&=\sum_{i=1}^{\Dx}
   q_i(g_i(\bm{x}),\bu)+\log|\det\bm{J}_{\bm{g}}(\bm{x})|-\log{Z}(\bu),
   \label{eqn:invert-f}
  \end{align}
  where $\bm{g}(\bx)=(g_1(\bx),g_2(\bx),\dots,g_{\Dx}(\bx))^{\top}$ and
  $\bm{J}_{\bm{g}}(\bm{x})$ is the Jacobian of $\bm{g}$ at $\bm{x}$.
  Similarly, by applying the same change of variables for the marginal
  density of $\bm{s}$, the marginal density of $\bm{x}$ is given by
  \begin{align}
   \log{p}(\bm{x})&=\log{p}(\bm{g}(\bm{x}))+\log|\det\bm{J}_{\bm{g}}(\bm{x})|.
  \label{eqn:marginal-x}
  \end{align}
  On the other hand, the log-density ratio
  equation~\eqref{ICA-universal} yields
  \begin{align}
   \log p(\bm{x}|\bm{u})-\log p(\bm{x})=
   \sum_{i=1}^{\Dx}\psi^{\star}_i(\hx{i}^{\star}(\bm{x}),\bhu^{\star}(\bu))
   +a^{\star}(\bhx^{\star}(\bx))+b^{\star}(\bhu^{\star}(\bu)). 
   \label{nICA-univ-approx}
  \end{align}
  Substitution of~\eqref{eqn:invert-f} and~\eqref{eqn:marginal-x} into
  the left-hand side of~\eqref{nICA-univ-approx} cancels out the
  Jacobian term and gives the following equation:
  \begin{align}
   &\sum_{i=1}^{\Dx}q_i(g_i(\bm{x}),\bu)-\log{Z}(\bu)-\log p(\bm{g}(\bx))=   
   \sum_{i=1}^{\Dx}\psi_i^{\star}(\hx{i}^{\star}(\bm{x}),\bhu^{\star}(\bu))
   +a^{\star}(\bhx^{\star}(\bx))+b^{\star}(\bhu^{\star}(\bu)). 
   \label{combination}
  \end{align}
  Then, we compute the gradient of the both sides on~\eqref{combination}
  with respect to $\bu$ as
  \begin{align*}
   \sum_{i=1}^{\Dx} \nabla_{\bu}q_i(g_i(\bm{x}),\bm{u})
   -\nabla_{\bu}\log{Z}(\bu)=\sum_{i=1}^{\Dx}
   \nabla_{\bu}\psi_i^{\star}(\hx{i}^{\star}(\bm{x}),\bhu^{\star}(\bu))
   +\nabla_{\bu}b^{\star}(\bhu^{\star}(\bu)). 
  \end{align*}
  By using the notations of $\bm{z}:=\bhx^{\star}(\bm{x})$ and
  $\bm{v}(\bm{z}):=\bm{g}((\bhx^{\star})^{-1}(\bm{z}))$ where 
  $(\bhx^{\star})^{-1}$ denotes the inverse of $\bhx^{\star}$,
  \begin{align}
   \sum_{i=1}^{\Dx}\nabla_{\bu} q_i(v_i(\bm{z}),\bu)
   -\nabla_{\bu}\log{Z}(\bu)=\sum_{i=1}^{\Dx}
   \nabla_{\bm{u}} \psi_i^{\star}(z_i, \bhu^{\star}(\bm{u}))+\nabla_{\bu}
   b^{\star}(\bhu^{\star}(\bu)),
   \label{eqn:invert-h}
  \end{align}
  where $z_i$ and $v_i$ are the $i$-th elements in $\bm{z}$ and
  $\bm{v}$, respectively.

  Next, we show from~\eqref{eqn:invert-h} that each element in
  $\bm{v}(\bm{z})(=\bm{s})$ is a function of a distinct and single
  element in $\bm{z}(=\bhx^{\star}(\bx))$. To this end, we first take
  the partial derivative to both sides on~\eqref{eqn:invert-h} with
  respect to $z_l$ as
  \begin{align}
   \sum_{i=1}^{\Dx}v_i^{(l)}(\bm{z})
   \nabla_{\bu} q_i^{\prime}(v_i(\bm{z}),\bu)   
   =\frac{\partial}{\partial z_l}
   \left[\nabla_{\bm{u}} \psi_l^{\star}(z_l, \bhu^{\star}(\bm{u}))\right],
   \label{one-differential-eq}   
  \end{align}
  where $q_i^{\prime}(t,\bm{u}):=\parder{q_i(t,\bm{u})}{t}$ and
  $v_i^{(l)}(\bm{z}):=\parder{v_i(\bm{z})}{z_l}$. We further take the
  partial derivative of both sides on~\eqref{one-differential-eq} with
  respect to $z_m$ for $m\neq l$ as
  \begin{align}
   \sum_{i=1}^{\Dx}
   \left[v_i^{(l,m)}(\bm{z})
   \nabla_{\bm{u}}q_i^{\prime}(v_i(\bm{z}),\bm{u})
   +v_i^{(l)}(\bm{z})v_i^{(m)}(\bm{z})
   \nabla_{\bm{u}}q_i^{\prime\prime}(v_i(\bm{z}),\bm{u})\right]
   =\bm{0}, \label{two-differential-eq}
  \end{align}
  where $q_i^{\prime\prime}(t,\bm{u}):=\frac{\partial^2}{\partial
  t^2}q_i(t,\bm{u})$ and
  $v_i^{(l,m)}(\bm{z}):=\frac{\partial^2}{\partial z_l\partial
  z_m}v_i(\bm{z})$. In order to express~\eqref{two-differential-eq} as a
  matrix form, we define the $\Dx(\Dx-1)/2$ by $2\Dx$ matrix consisting
  of a collection of $v_i^{(l)}(\bm{z})v_i^{(m)}(\bm{z})$ and
  $v_i^{(l,m)}(\bm{z})$ with respect to $i,l,m$ ($l\neq{m}$) as
  \begin{align*}
   \bm{M}(\bm{v})&:=\left(
   \begin{array}{cccccc}
    v_1^{(1,2)} & \cdots & v_{\Dx}^{(1,2)} & v_{1}^{(1)}v_{1}^{(2)} & \cdots
     & v_{\Dx}^{(1)}v_{\Dx}^{(2)}\\
    v_1^{(1,3)} & \cdots & v_{\Dx}^{(1,3)} & v_{1}^{(1)}v_{1}^{(3)} & \cdots
     & v_{\Dx}^{(1)}v_{\Dx}^{(3)}\\
    \vdots & \vdots& \vdots & \cdots & \vdots &\vdots\\    
    v_1^{(\Dx-1,\Dx)} & \cdots & v_{\Dx}^{(\Dx-1,\Dx)} & v_{1}^{(\Dx-1)}v_{1}^{(\Dx)} 
    & \cdots & v_{\Dx}^{(\Dx-1)}v_{\Dx}^{(\Dx)}\\
   \end{array}
   \right)\in\R{\frac{\Dx(\Dx-1)}{2}\times{2}\Dx}.
   \end{align*}
  We also define a collection of $q^{\prime}_i(v_i(\bm{z}),\bu)$ and
  $q^{\prime\prime}_i(v_i(\bm{z}),\bu)$ by the following
  $2\Dx$-dimensional vector:
  \begin{align*}
   \bm{w}(\bm{v},\bm{u})&:=\left(q^{\prime}_1(v_1,\bm{u}),\dots,
   q^{\prime}_{\Dx}(v_{\Dx},\bm{u}),q^{\prime\prime}_1(v_1,\bu),
   \dots,q^{\prime\prime}_{\Dx}(v_{\Dx},\bu)\right)^{\top}
   \in\R{{2}\Dx}.
  \end{align*}
  By fixing $\bu$ at $\bu=\bu_1$, \eqref{two-differential-eq} can be
  expressed by using $\bm{M}(\bm{v})$ and $\bm{w}(\bm{v},\bm{u})$ as
  \begin{align}
   \bm{M}(\bm{v})\bm{W}(\bm{v})=\bm{O},
   \label{two-differential-eq-mat}
  \end{align}  
  where
  $\bm{W}(\bm{v}):=\nabla_{\bm{u}}\bm{w}(\bm{v},\bm{u})\Bigr|_{\bu=\bu_1}\in\R{2\Dx\times\Du}$
  and $\bm{O}$ denotes the null matrix.  Since Assumptions (B3-4) ensure
  that the rank of $\bm{W}(\bm{v})$ is $2\Dx$,
  $\bm{W}(\bm{v})\bm{W}(\bm{v})^{\top}$ is a $2\Dx$ by $2\Dx$ invertible
  matrix. Thus, multiplying $\bm{W}(\bm{v})^{\top}$
  to~\eqref{two-differential-eq-mat} and taking the inverse of
  $\bm{W}(\bm{v})\bm{W}(\bm{v})^{\top}$ on the right yields
  \begin{align}
   \bm{M}(\bm{v})=\bm{O}.
   \label{eqn:M-null}
  \end{align}
  Eq.\eqref{eqn:M-null} indicates that $v_i^{(l,m)}(\bm{z})=0$ and
  $v_i^{(l)}(\bm{z})v_i^{(m)}(\bm{z})=0$ for all $i,l,m=1,\dots,\Dx$
  with $l\neq{m}$. This means that each element in $\bm{v}(\bm{z})$ is a
  function of a distinct and single element in $\bm{z}$ because
  $\bm{v}(\bm{z})$ is an invertible function.  Thus, the proof is
  completed.
 \end{proof}
 \section{Rank condition in the exponential family}
 \label{app:rank-inequality}
 We first express the exponential family~\eqref{exp-family} as
 \begin{align*}
  \log{p}(\bm{s}|\bu)=\sum_{k=1}^K\bm{\lambda}_k(\bu)^{\top}\bm{q}_k(\bm{s})-\log{Z}(\bu),
 \end{align*}
 where
 $\bm{\lambda}_k(\bu):=(\lambda_{1k}(\bu),\dots,\lambda_{\Dx{k}}(\bu))^{\top}$
 and $\bm{q}_k(\bm{s}):=(q_{1k}(s_1), \dots,q_{\Dx{k}}(s_{\Dx}))$.
 Then, a simple computation yields
 \begin{align}
  \nabla_{\bm{u}}\bm{w}(\bm{v},\bm{u}) =\sum_{k=1}^K\bm{\Lambda}_k\odot
  \bm{Q}_k, \label{Hadamard}
 \end{align}
 where $\odot$ denotes the Hadamard product (i.e., elementwise
 multiplication of two matrices)
 \begin{align*}
  \bm{\Lambda}_k:=\left(
  \begin{array}{c}
   \nabla_{\bu} \bm{\lambda}_k(\bm{u})\\
   \nabla_{\bu} \bm{\lambda}_k(\bm{u})
  \end{array}
  \right),
 \end{align*}
 and with $q^{\prime}_{ik}(t):=\frac{\intd}{\intd{t}}q_{ik}(t)$ and
 $q^{\prime\prime}_{ik}(t):=\frac{\intd^2}{\intd{t}^2}q_{ik}(t)$ for
 $i=1,\dots,\Dx$,
 \begin{align*}
  {\bm{Q}}_k= \left(
  \begin{array}{cccc}
   q^{\prime}_{1k}(v_1)& q^{\prime}_{1k}(v_1) & \cdots &
    q^{\prime}_{1k}(v_1)\\ 
   \vdots & \vdots & \cdots & \vdots \\
   q^{\prime}_{\Dx{k}}(v_{\Dx})& 
   q^{\prime}_{\Dx{k}}(v_{\Dx})& \cdots &
   q^{\prime}_{\Dx{k}}(v_{\Dx}) \\
   q^{\prime\prime}_{1k}(v_1)& q^{\prime\prime}_{1k}(v_1) & \cdots &
    q^{\prime\prime}_{1k}(v_1)\\ 
   \vdots & \vdots & \cdots & \vdots \\
   q^{\prime\prime}_{\Dx{k}}(v_{\Dx})& 
   q^{\prime\prime}_{\Dx{k}}(v_{\Dx})& \cdots &
   q^{\prime\prime}_{\Dx{k}}(v_{\Dx})
  \end{array}   
  \right).
 \end{align*}
 The rank of $\bm{\Lambda}_k$ is at most $\Dx$ under Assumption~(B3)
 because $\bm{\Lambda}_k$ is a vertical concatenation of the two same
 matrices of $\nabla_{\bu}\bm{\lambda}_k(\bm{u})\in\R{\Dx\times\Du}$. On
 the other hand, $\bm{Q}_k$ is a rank-one matrix because the column
 vectors in ${\bm{Q}}_k$ are all same. Thus, the rank factorization
 ensures that the following decompositions of $\bm{\Lambda}_k$ and
 $\bm{Q}_k$ exist:
 \begin{align}
  \bm{\Lambda}_k=\sum_{i=1}^{\Dx} \bm{a}_{ik}\bm{b}_{ik}^{\top}
  \quad~\text{and}\quad~{\bm{Q}}_k=\bm{c}_k\bm{d}_k^{\top},
  \label{rank-one-matrices}
 \end{align}
 where $\bm{a}_{ik}, \bm{c}_k\in\R{2\Dx}$ and
 $\bm{b}_{ik}, \bm{d}_k\in\R{\Du}$. Substituting~\eqref{rank-one-matrices}
 into~\eqref{Hadamard} yields
 \begin{align}
  \nabla_{\bm{u}}\bm{w}(\bm{v},\bm{u})=\sum_{k=1}^K\sum_{i=1}^{\Dx} 
  (\bm{a}_{ik}\odot\bm{c}_k)(\bm{b}_{ik}\odot\bm{d}_k)^{\top}.
  \label{exp-equatuon}
 \end{align}
 indicating that the rank of $\nabla_{\bm{u}}\bm{w}(\bm{v},\bm{u})$ is
 at most $2\Dx$ when $K>1$. On the other hand, when $K=1$,
 \eqref{exp-equatuon} is given by
 \begin{align*}
  \nabla_{\bm{u}}\bm{w}(\bm{v},\bm{u})=\sum_{i=1}^{\Dx} 
  (\bm{a}_{i1}\odot\bm{c}_1)(\bm{b}_{i1}\odot\bm{d}_1)^{\top},
 \end{align*}
 and thus the rank is at most $\Dx$.
 \section{Proof of Theorem~\ref{prop:general-PCL}}
 \label{app:proof-general-PCL}
 Our proof is based on the following lemmas:
 \begin{lemma}
  \label{lem:diagonals} Suppose that $\bm{D}(\bm{v})$ is an $n$ by $n$
  diagonal matrix whose diagonals $d_{i}(\bm{v})$ is a function of
  $\bm{v}\in\R{n}$, and $\bm{A}$ and $\bm{B}$ are $n$ by $n$ constant
  matrices. Furthermore, the following assumptions are made:
  \begin{enumerate}
   \item[(1)] There exist $n$ points $\bm{v}_1,
	      \bm{v}_2,\dots,\bm{v}_n$ such that
	      $\bm{d}(\bm{v}_1),\bm{d}(\bm{v}_2),\dots,\bm{d}(\bm{v}_n)$
	      are linearly independent where
	      $\bm{d}(\bm{v}):=(d_{1}(\bm{v}),d_{2}(\bm{v}),\dots,d_{n}(\bm{v}))^{\top}$
	      is the vector of the diagonal elements in
	      $\bm{D}(\bm{v})$.

   \item[(2)] $\bm{A}$ and $\bm{B}$ are of full-rank.
  \end{enumerate}
  Then, when $\bm{A}\bm{D}(\bm{v})\bm{B}$ is a diagonal matrix at least
  at $n$ points $\bm{v}_1, \bm{v}_2,\dots,\bm{v}_n$, then both $\bm{A}$
  and $\bm{B}$ are diagonal matrices multiplied by a permutation matrix.
 \end{lemma}
 
 \begin{lemma}[Theorem~4.4.8 in~\citet{harville2006matrix}]
  \label{lem:rank} Suppose that $\bm{A}$ is an $m$ by $n$ matrix of rank
  $r$. For any $m$ by $r$ matrix $\bm{B}$ and $r$ by $n$ matrix $\bm{T}$
  such that $\bm{A}=\bm{B}\bm{T}$, both $\bm{B}$ and $\bm{T}$ have rank
  $r$.
 \end{lemma}
 Lemma~\ref{lem:diagonals} is proved in Appendix~\ref{app:diagonals},
 while the reader may refer to the proof of Theorem~4.4.8
 in~\citet{harville2006matrix} for Lemma~\ref{lem:rank}.
  \subsection{Main proof}
  \begin{proof}
   We start by using the same notations and following the same line of
   the proof until~\eqref{combination} in
   Section~\ref{app:general-ICA}. Then, we obtain the following equation
   under Assumptions~(B$^{\prime}$1-2,6):
    \begin{align}
     &\sum_{i=1}^{\Dx} q_i(g_i(\bm{x}),\hq{i}(\bu))
     -\log{Z}(\bu)-\log p(\bm{g}(\bx))\nonumber\\
     &\qquad=\sum_{i=1}^{\Dx}
     \psi_i^{\star}(\hx{i}^{\star}(\bm{x}),\hu{i}^{\star}(\bu))
     +a^{\star}(\bhx^{\star}(\bx))+b^{\star}(\bhu^{\star}(\bu)),
     \label{eqn:general-PCL-comb}
    \end{align}
   where $\bm{g}$ denotes the inverse of the mixing function $\bm{f}$
   and $\bm{g}(\bx)=(g_1(\bx),g_2(\bx),\dots,g_{\Dx}(\bx))^{\top}$.  Let
   us denote $\bm{z}:=\bhx^{\star}(\bm{x})$,
   $\bm{v}(\bm{z}):=\bm{g}((\bhx^{\star})^{-1}(\bm{z}))$,
   $\ru{i}:=\hu{i}^{\star}(\bu)$ and $\rq{i}:=\hq{i}(\bu)$. Then, taking
   a cross-derivative with respect to $v_l$ and $\rq{m}$ to both side
   on~\eqref{eqn:general-PCL-comb} yields
    \begin{align}
     \sum_{i=1}^{\Dx} \frac{\partial^2 q_i(v_i,\rq{i})}{\partial
     v_l\partial \rq{m}}=\sum_{i=1}^{\Dx} \frac{\partial z_i}{\partial v_l}
     \frac{\partial^2 \psi^{\star}_i(z_i,\ru{i})} {\partial z_i\partial \ru{i}}
     \sum_{j=1}^{\Dx} \frac{\partial \ru{i}}{\partial u_j} \frac{\partial
     u_j}{\partial \rq{m}}.  \label{cross-der}
    \end{align} 
    To express~\eqref{cross-der} as a matrix form, we define the following
    $\Dx$ by $\Dx$ diagonal matrices:
    \begin{align*}
     \bm{D}_q(\bm{v},\brq)&:=\left(
     \begin{array}{cccccc}
      \frac{\partial^2 q_1(v_1,\rq{1})}{\partial v_1\partial
       \rq{1}} & 0 & \cdots & 0\\
      0 & \frac{\partial^2 q_2(v_2,\rq{2})}{\partial v_2\partial
       \rq{2}} &  \cdots  & 0\\
      \vdots & \vdots& \vdots &\vdots\\    
      0 & 0 &  \cdots &     
       \frac{\partial^2 q_{\Dx}(v_{\Dx},\rq{\Dx})}{\partial v_{\Dx}\partial \rq{\Dx}}\\
     \end{array}
     \right)\in\R{\Dx\times\Dx}\\
     \bm{D}_{\psi}(\bm{v},\bru)&:=
     \left(
     \begin{array}{cccccc}
      \frac{\partial^2 \psi^{\star}_1(v_1,\ru{1})}{\partial v_1\partial
       \ru{1}} & 0 & \cdots & 0\\
      0 & \frac{\partial^2 \psi^{\star}_2(v_2,\ru{2})}{\partial v_2\partial
       \ru{2}} &  \cdots  & 0\\
      \vdots & \vdots& \vdots &\vdots\\    
      0 & 0 &  \cdots &     
       \frac{\partial^2 \psi^{\star}_{\Dx}(v_{\Dx},\ru{\Dx})}
       {\partial v_{\Dx}\partial \ru{\Dx}}\\
     \end{array}
     \right)\in\R{\Dx\times\Dx},   
    \end{align*}
    where $\brq=(\rq{1},\rq{2},\dots,\rq{\Dx})^{\top}$ and
    $\bru=(\ru{1},\ru{2},\dots,\ru{\Dx})^{\top}$.
    Then,~\eqref{cross-der} can be expressed as
    \begin{align}
     \bm{D}_{\mathrm{q}}(\bm{v},\brq)
     =\bm{J}_{\bm{z}}(\bm{v})\bm{D}_{\psi}(\bm{v},\bru)
     \bm{J}_{\bru}(\bu)\bm{J}_{\bu}(\brq),   
     \label{cross-der-mat-org}
    \end{align}
   where
   $\bm{J}_{\bm{z}}(\bm{v}):=\nabla_{\bm{v}}\bm{z}\in\R{\Dx\times\Dx}$,
   $\bm{J}_{\bru}(\bu):=\nabla_{\bu}\bru\in\R{\Dx\times\Du}$ and
   $\bm{J}_{\bu}(\brq):=\nabla_{\brq}\bu\in\R{\Du\times\Dx}$ are the
   Jacobian matrices of $\bm{z}$, $\bru$ and $\bu$, respectively.  We
   fix $\bu$ at $\bu=\bu_1$ and express~\eqref{cross-der-mat-org} as
   \begin{align}
    \bm{D}_{\mathrm{q}}^{1}(\bm{v})
    =\bm{J}_{\bm{z}}(\bm{v})\bm{D}_{\psi}^{1}(\bm{v}) \bm{J}^{1},
    \label{cross-der-mat}
   \end{align}
    where
    $\bm{D}_{\mathrm{q}}^{1}(\bm{v}):=\bm{D}_{\mathrm{q}}(\bm{v},\brq)$
    at $\brq=\bm{\lambda}(\bu_1)$,
    $\bm{D}_{\psi}^{1}(\bm{v}):=\bm{D}_{\psi}(\bm{v},\bru)$ at
    $\bru=\bhu^{\star}(\bu_1)$, and
    $\bm{J}^{1}:=\bm{J}_{\bru}(\bu_1)\bm{J}_{\bu}(\brq)$ at
    $\brq=\bm{\lambda}(\bu_1)$.  We can have a similar matrix
    expression as~\eqref{cross-der-mat} at $\bu=\bu_2$ as
    \begin{align}
     \bm{D}_{\mathrm{q}}^{2}(\bm{v})
     =\bm{J}_{\bm{z}}(\bm{v})\bm{D}_{\psi}^{2}(\bm{v})\bm{J}^{2}.
     \label{cross-der-mat2}
    \end{align}
    Here, we note that both $\bm{J}^{1}\in\R{\Dx\times\Dx}$ and
   $\bm{J}^{2}\in\R{\Dx\times\Dx}$ have at most rank $\Dx$ by
    Assumption~(B$'$3).
    
    Next, we confirm that $\bJ^{1}$, $\bJ^{2}$,
   $\bm{D}_{\psi}^{1}(\bm{v})$ and $\bm{D}_{\psi}^{2}(\bm{v})$ for
   all $\bm{v}$ have rank $\Dx$, and thus are invertible.  To this end,
   we denote the $i$-th diagonals in $\bm{D}_{\mathrm{q}}^{1}(\bm{v})$
   and $\bm{D}_{\mathrm{q}}^{2}(\bm{v})$ by
    \begin{align*}
     \alpha_i^{1}(v_i):=\frac{\partial^2{q}_i(v_i,r)}{\partial{v_i}\partial{r}}\Bigr|_{r=\hq{i}(\bu_1)}
     \quad\text{and}\quad
     \alpha_i^{2}(v_i):=\frac{\partial^2{q}_i(v_i,r)}{\partial{v_i}\partial{r}}\Bigr|_{r=\hq{i}(\bu_2)},
    \end{align*}
   respectively.  From Assumptions~(B$^{\prime}$4),
   $\bm{D}_{\mathrm{q}}^{1}(\bm{v})$ has nonzero diagonals
   $\alpha_i^{1}(v_i)$ and rank $\Dx$ for all $\bm{v}$.  Then,
   applying Lemma~\ref{lem:rank} to~\eqref{cross-der-mat} ensures that
   $\bm{D}_{\psi}^{1}(\bm{v})\bJ^{1}$ have rank $\Dx$.  By applying
   Lemma~\ref{lem:rank} to~$\bm{D}_{\psi}^{1}(\bm{v})\bJ^{1}$ again,
   it can be shown that both $\bm{D}_{\psi}^{1}(\bm{v})$ and
   $\bJ^{1}$ also have rank $\Dx$. Similarly, we can prove that both
   $\bm{D}_{\psi}^{2}(\bm{v})$ and $\bJ^{2}$ have rank $\Dx$ under
   Assumption~(B$'$4).
   
   Finally, we show that $\bm{J}_{\bm{z}}(\bm{v})$ is the product of a
   permutation and diagonal matrices. Since $\bm{D}_{\psi}^{1}(\bm{v})$
   and $\bJ^{1}$ are invertible, from~\eqref{cross-der-mat}, we have
   \begin{align}
    \bm{J}_{\bm{z}}(\bm{v})=\bm{D}_{\mathrm{q}}^{1}(\bm{v})
    [\bm{J}^{1}]^{-1}[\bm{D}_{\psi}^{1}(\bm{v})]^{-1}.
    \label{Jz-jacobian}
   \end{align}
   Substituting~\eqref{Jz-jacobian} into~\eqref{cross-der-mat2}, and
   multiplying $\bm{J}^{1}$ and the inverses of
   $\bm{D}_{\mathrm{q}}^{1}(\bm{v})$ and $\bm{J}^{2}$
   to both sides yields
   \begin{align}
    \bm{J}^{1}\left[[\bm{D}_{\mathrm{q}}^{1}(\bm{v})]^{-1}
    \bm{D}_{\mathrm{q}}^{2}(\bm{v})\right][\bm{J}^{2}]^{-1}
    =[\bm{D}_{\psi}^{1}(\bm{v})]^{-1} \bm{D}_{\psi}^{2}(\bm{v}).
    \label{diag-eq}
   \end{align}
   Then, we compactly express all of diagonals in
   $[\bm{D}_{\mathrm{q}}^{1}(\bm{v})]^{-1}\bm{D}_{\mathrm{q}}^{2}(\bm{v})$
   as the following vector:
   \begin{align*}
    \bm{\alpha}(\bm{v}):=\left(\frac{\alpha_1^{2}(v_1)}{\alpha_1^{1}(v_1)},
    \frac{\alpha_2^{2}(v_2)}{\alpha_2^{1}(v_2)},\dots,\frac{\alpha_{\Dx}^{2}(v_{\Dx})}{\alpha_{\Dx}^{1}(v_{\Dx})}\right)^{\top}.     
   \end{align*}    
   By Assumption~(B$'$5), $\bm{\alpha}(\bm{v}_1), \bm{\alpha}(\bm{v}_2),
   \dots, \bm{\alpha}(\bm{v}_{\Dx})$ are linearly independent. Since
   $\bm{J}^{1}$ and $\bm{J}^{2}$ are proved to have rank $\Dx$ and thus
   constant full-rank matrices, applying Lemma~\ref{lem:diagonals}
   ensures that $\bm{J}^{1}$ and $\bm{J}^{2}$ are diagonal matrices
   multiplied by a permutation matrix. Thus, it follows
   from~\eqref{Jz-jacobian} that $\bm{J}_{\bm{z}}(\bm{v})$ is also a
   diagonal matrix multiplied by a permutation matrix. Thus, each
   $v_i(\bm{z})(=s_i)$ corresponds to a single and distinct element in
   $\bm{z}(=\bhx^{\star}(\bx))$. The proof is completed.
  \end{proof}
  \subsection{Proof of Lemma~\ref{lem:diagonals}}
  \label{app:diagonals}
  \begin{proof}
   Let us first denote the $(l,m)$-th elements of $\bm{A}$ and $\bm{B}$
   by $a_{lm}$ and $b_{lm}$. The $(l,m)$-th element in
   $\bm{A}\bm{D}(\bm{v})\bm{B}$ is given by
   \begin{align*}
    \sum_{i=1}^{n}d_{i}(\bm{v})a_{li}b_{im}.
   \end{align*}
   Since off-diagonal elements in $\bm{A}\bm{D}(\bm{v})\bm{B}$ are all
   zeros at $\bm{v}=\bm{v}_j,~j=1,\dots,n$,
   \begin{align}
    \sum_{i=1}^{n}d_{i}(\bm{v}_j)a_{li}b_{im}=\bm{d}(\bm{v}_j)^{\top}\bm{c}^{lm}=0
    \quad(l\neq m), \label{elements}
   \end{align}
   where 
   \begin{align*}
    \bm{c}^{lm}&:=(a_{l1}b_{1m}, a_{l2}b_{2m}, \dots, a_{ln}b_{nm})^{\top}.
   \end{align*}
   Collecting $\bm{d}(\bm{v})$ over the $n$ points
   $\bm{v}_1,\bm{v}_2,\dots,\bm{v}_{n}$ based on~\eqref{elements} yields
   \begin{align}
    \widetilde{\bm{D}}^{\top}\bm{c}^{lm}=\bm{0}, \label{eqn:c-zero}
   \end{align}
   where
   \begin{align*}
    \widetilde{\bm{D}}&:=[\bm{d}(\bm{v}_1),\bm{d}(\bm{v}_2),\dots,\bm{d}(\bm{v}_n)]\in\R{n\times{n}}.
   \end{align*}
   By Assumption~(1), $\widetilde{\bm{D}}$ is of full-rank and
   invertible. Thus, from~\eqref{eqn:c-zero}, we obtain
   $\bm{c}^{lm}=\bm{0}$ for $l,m=1,\dots,n$ with $l\neq{m}$, indicating
   that
   \begin{align}
    a_{li}b_{im}=0\quad(i,l,m=1,\dots,n~\text{and}~l\neq{m}).
    \label{nonzero-diagonals}
   \end{align} 
  
   Next, we show that both the $i$-th column vector in $\bm{A}$ and the
   $i$-th row vector in $\bm{B}$ have a single nonzero element, while
   the other elements are zeros. We first suppose that the $i$-th column
   vector in $\bm{A}$ has at least two nonzeros elements such that
   $a_{li}\neq{0}$ and $a_{l'i}\neq{0}$ for $l\neq{l}'$. $a_{li}\neq{0}$
   implies that $b_{im}=0$ except for $m=l$, but $a_{l'i}\neq{0}$
   ensures $b_{im}=0$ at $m=l$. Thus, the $i$-th row vector in $\bm{B}$
   must be the zero vector. However, this contradicts to the assumption
   that $\bm{B}$ is of full-rank. Therefore, the $i$-th column vector of
   $\bm{A}$ must have the single nonzero element. Similarly, we can
   prove that the $i$-th row vector in $\bm{B}$ also has a single
   nonzero element. Thus, by the full-rank assumption of $\bm{A}$ and
   $\bm{B}$, all column and row vectors in $\bm{A}$ and $\bm{B}$ must
   have the single nonzero element at distinct positions. Thus, $\bm{A}$
   and $\bm{B}$ are equal to diagonal matrices multiplied by a
   permutation matrix.
  \end{proof}
 \section{Proof of Theorem~\ref{theo:manifold}}
 \label{app:manifold}
 \begin{proof}
  Taking the gradient of both sides on~\eqref{manifold-universal} with
  respect to $\bu$ yields
  \begin{align}
   \nabla_{\bm{u}}\log p(\bm{x}|\bm{u})=\nabla_{\bu}
   \psi^{\star}(\bhx^{\star}(\bx),\bhu^{\star}(\bu))
   +\nabla_{\bu}b^{\star}(\bhu^{\star}(\bu)).
   \label{conditional-pdf-sn2}
  \end{align}
  By the change of variables by $\bx=\bm{f}(\bm{s},\bm{n})$ in the
  generative model~\eqref{generative-model2}, we
  re-express~\eqref{conditional-pdf-sn2} as
  \begin{align}
   \nabla_{\bu}\log p(\bm{s},\bm{n}|\bm{u})      
   =\nabla_{\bu}\psi^{\star}(\bhx^{\star}(\bm{f}(\bm{s},\bm{n})),\bhu^{\star}(\bu))
   +\nabla_{\bu}b^{\star}(\bhu^{\star}(\bu)),
   \label{mani-comb}
  \end{align}
  where we note that the Jacobian term due to the change of variables is
  deleted by the differential operator $\nabla_{\bu}$.
  
  Next, with $\bm{v}:=\bhx^{\star}(\bm{f}(\bm{s},\bm{n}))$, we compute
  the gradient of the left-hand side on~\eqref{mani-comb} with respect
  to $\bm{n}$ as
  \begin{align}
   \nabla_{\bm{n}}\nabla_{\bu}\log p(\bm{s},\bm{n}|\bm{u})
   &=\nabla_{\bm{n}}[\nabla_{\bu} \{\log p(\bm{s}|\bm{u})+\log
   p(\bm{n})\}]\nonumber\\ &=\nabla_{\bm{n}}[\nabla_{\bu}\log
   p(\bm{s}|\bm{u})]\nonumber\\ &=\bm{O},
   \label{null-mat}
  \end{align}  
  where we used Assumption~(C1) that $\bm{s}\perp\bm{n}$,
  $\bm{u}\perp\bm{n}$ and $\bm{s}\not\perp\bm{u}$. On the other hand,
  the gradient of the right-hand side on~\eqref{mani-comb} is given by
  \begin{align}
   \nabla_{\bm{n}}[\nabla_{\bu}\psi^{\star}(\bm{v},\bhu^{\star}(\bu))]
   =[\nabla_{\bm{v}}\nabla_{\bu}\psi^{\star}(\bm{v},\bhu^{\star}(\bu))]
   \bJ_{\bm{v}}(\bm{n}), \label{right-grad}
  \end{align}
  where
  $\nabla_{\bm{v}}\nabla_{\bu}\psi^{\star}(\bm{v},\bhu^{\star}(\bu))\in\R{\Du\times\dx}$
  and
  $\bJ_{\bm{v}}(\bm{n}):=\nabla_{\bm{n}}\bm{v}\in\R{\dx\times(\Dx-\dx)}$
  is the Jacobian matrix of $\bm{v}$ at $\bm{n}$.  We fix $\bu$ at
  $\bu=\bu_1$ and equate~\eqref{right-grad} with~\eqref{null-mat} based
  on~\eqref{mani-comb} as
  \begin{align}
   \bm{M}(\bm{v})\bJ_{\bm{v}}(\bm{n})=\bm{O},
   \label{zero-equation}
  \end{align}
  where
  $\bm{M}(\bm{v}):=\nabla_{\bm{v}}\nabla_{\bu}\psi^{\star}(\bm{v},\bhu^{\star}(\bu))\bigr|_{\bu=\bu_1}\in\R{\Du\times\dx}$.
  By Assumption~(C4), the rank of $\bm{M}(\bm{v})$ is $\dx$, and thus
  the inverse of $\bm{M}(\bm{v})^{\top}\bm{M}(\bm{v})$ exists.
  Multiplying $\bm{M}(\bm{v})^{\top}$ on the left and taking the inverse
  of $\bm{M}(\bm{v})^{\top}\bm{M}(\bm{v})$ to~\eqref{zero-equation}
  indicates $\bJ_{\bm{v}}(\bm{n})=\bm{O}$.  Recall that
  $\bm{v}=\bhx^{\star}\circ\bm{f}(\bm{s},\bm{n})=\bhx^{\star}(\bx)$. Since
  $\bm{f}(\bm{s},\bm{n})$ and $\bhx^{\star}(\bm{x})$ are invertible and
  surjective respectively, $\bJ_{\bm{v}}(\bm{n})=\bm{O}$ ensures that
  $\bm{v}$ is a (vector-valued) function of only $\bm{s}$. Thus, the
  proof is completed.
 \end{proof}  
 \section{Derivation of~\eqref{two-grads}}
 \label{app:two-grads} 
 By Assumption~(C1) that $\bm{s}\perp\bm{n}$, $\bm{u}\perp\bm{n}$ and
 $\bm{s}\not\perp\bm{u}$, \eqref{mani-comb} can be written as
 \begin{align*}
  \nabla_{\bu}\log p(\bm{s}|\bm{u})      
  =\nabla_{\bu}\psi^{\star}(\bm{v},\bhu^{\star}(\bu))
  +\nabla_{\bu}b^{\star}(\bhu^{\star}(\bu)),
 \end{align*}
 where $\bm{v}:=\bhx^{\star}(\bm{f}(\bm{s},\bm{n}))$. Taking the
 gradient with respect to $\bm{v}$ yields
 \begin{align*}
  \bm{J}_{\bm{v}}(\bm{s})[\nabla_{\bm{s}}\nabla_{\bu}\log p(\bm{s}|\bm{u})]
  =\nabla_{\bm{v}}\nabla_{\bu}\psi^{\star}(\bm{v},\bhu^{\star}(\bu)),
 \end{align*}
 where $\bm{J}_{\bm{v}}(\bm{s}):=\nabla_{\bm{s}}\bm{v}$ denotes the
 Jacobian of $\bm{v}$ at $\bm{s}$. Thus, we obtain~\eqref{two-grads} by
 fixing $\bm{u}$ at $\bm{u}=\bm{u}_1$.
 
 \section{Proof of Proposition~\ref{prop:IF}}
 \label{app:IF}
 \begin{proof}
  Firstly, we consider the contamination model~$2$.  By definition, the
  estimator $\vtheta^{\star}$ satisfies the following equation:
  \begin{align}
   \left.\nabla_{\vtheta}\left\{ \Exu[r_{\vtheta}(\bX,\bU)]-\log
   \Extu[e^{r_{\vtheta}(\bX,\bU)}]\right\} \right|_{\vtheta=\vtheta^{\star}}=\bm{0},
   \label{equiv-dv}
  \end{align}
  while $\vtheta_{\epsilon}$ associated with the contaminated densities
  fulfills
  \begin{align}
   \left.\nabla_{\vtheta} \left\{
   \bExu[r_{\vtheta}(\bX,\bU)]-\log \bExtu[e^{r_{\vtheta}(\bX,\bU)}]\right\}
   \right|_{\vtheta=\vtheta_{\epsilon}}=\bm{0}, \label{equiv.vd}
  \end{align}
  where $\bExu$ and $\bExtu$ denote the expectations over
  $\bar{p}(\bx,\bu)$ and $\bar{p}(\bx)\bar{p}(\bu)$, respectively.
  Applying the Taylor series expansion of~\eqref{equiv.vd} around
  $\vtheta^{\star}$ yields
  \begin{align}
   \bm{0}&=\left. \nabla_{\vtheta} \left\{ \bExu[r_{\vtheta}(\bX,\bU)]-\log
   \bExtu[e^{r_{\vtheta}(\bX,\bU)}]\right\} \right|_{\vtheta=\vtheta^{\star}}+
   \bVdv(\vtheta_{\epsilon}-\vtheta^{\star})+O(\|\vtheta_{\epsilon}-\vtheta^{\star}\|^2).
  \label{taylor-dv}
  \end{align}
  where $\bVdv$ is the Hessian matrix at $\vtheta=\vtheta^{\star}$ and
  defined by
  \begin{align*}
   \bVdv:=\nabla_{\vtheta}\nabla_{\vtheta}\left.\left\{
   \bExu[r_{\vtheta}(\bX,\bU)]-\log \bExtu[e^{r_{\vtheta}(\bX,\bU)}]\right\}
   \right|_{\vtheta=\vtheta^{\star}}.
  \end{align*}

  Let us denote $\nabla_{\vtheta}r_{\vtheta}(\bx,\bu)$ by
  $\bm{g}_{\vtheta}(\bx,\bu)$. We recall that the contaminated densities
  in the contamination model~2 are defined as
  \begin{align*}
   \bar{p}(\bx,\bu)&=(1-\epsilon)p(\bx,\bu)
   +\epsilon\delta_{\bar{\bx}}(\bx)\delta_{\bar{\bu}}(\bu)\\
   \bar{p}(\bx)&=(1-\epsilon)p(\bx)+\epsilon\delta_{\bar{\bx}}(\bx)\\
   \bar{p}(\bu)&=(1-\epsilon)p(\bu)+\epsilon\delta_{\bar{\bu}}(\bu).
  \end{align*}
  Then, with sufficiently small $\epsilon$, the first term on the
  right-hand side of~\eqref{taylor-dv} is given by
  \begin{align}
   \left.\nabla_{\vtheta} \left\{ \bExu[r_{\vtheta}(\bX,\bU)]-\log
   \bExtu[e^{r_{\vtheta}(\bX,\bU)}]\right\}
   \right|_{\vtheta=\vtheta^{\star}}
   &=\bExu[\bm{g}_{\vtheta^{\star}}(\bX,\bU)]
   -\frac{\bExtu[e^{r_{\vtheta^{\star}}(\bX,\bU)}\bm{g}_{\vtheta^{\star}}(\bX,\bU)]}{\bExtu[e^{r_{\vtheta^{\star}}(\bX,\bU)}]}
   \nonumber\\
   &\simeq\bExu[\bm{g}_{\vtheta^{\star}}(\bX,\bU)]
   -\bExtu[e^{r_{\vtheta^{\star}}(\bX,\bU)}\bm{g}_{\vtheta^{\star}}(\bX,\bU)],
   \label{grad-part-dv1}
  \end{align}
  where $\simeq$ denotes the equality up to terms of $\epsilon^2$, and
  we applied the following relation:
  \begin{align*}
   \bExtu[e^{r_{\vtheta^{\star}}(\bX,\bU)}]
   &\simeq\Extu[e^{r_{\vtheta^{\star}}(\bX,\bU)}]+\epsilon
   \left(\Ex[e^{r_{\vtheta^{\star}}(\bX,\bar{\bu})}]+\Eu[e^{r_{\vtheta^{\star}}(\bar{\bx},\bU)}]
   -2\Extu[e^{r_{\vtheta^{\star}}(\bX,\bU)}]\right)
   =1,
  \end{align*}
  where we used 
  \begin{align}
   \Extu[e^{r_{\vtheta^{\star}}(\bX,\bU)}]
   &=\Extu\left[\frac{p(\bX,\bU)}{p(\bX)p(\bU)}\right]=1,
   \label{IF-tmp1}\\
   \Ex[e^{r_{\vtheta^{\star}} (\bX,\bar{\bu})}]
   =\int p(\bx|\bar{\bu})d\bx&=1\quad\text{and}\quad
   \Eu[e^{r_{\vtheta^{\star}}(\bar{\bx},\bU)}]
   =\int{p}(\bu|\bar{\bx})d\bu=1,\label{IF-tmp2}
  \end{align}
  based on the
  assumption that
  $r_{\vtheta^{\star}}(\bx,\bu)=\log\frac{p(\bx,\bu)}{p(\bx)p(\bu)}$.
  Then, we substitute
  \begin{align*}
   \bExu[\bm{g}_{\vtheta^{\star}}(\bX,\bU)]=(1-\epsilon)\Exu[\bm{g}_{\vtheta^{\star}}(\bX,\bU)]+\epsilon\bm{g}_{\vtheta^{\star}}(\bar{\bx},\bar{\bu}),
  \end{align*}
  and
  \begin{align*}
   \bExtu[e^{r_{\vtheta^{\star}}(\bX,\bU)}\bm{g}_{\vtheta}(\bX,\bU)]
   &\simeq(1-\epsilon)\Extu[e^{r_{\vtheta^{\star}}(\bX,\bU)}\bm{g}_{\vtheta^{\star}}(\bX,\bU)]
   +\epsilon
   \left(\Ex[e^{r_{\vtheta^{\star}}(\bX,\bar{\bu})}\bm{g}_{\vtheta^{\star}}(\bX,\bar{\bu})]
   \right.\\
   &\qquad\qquad\left.
   +\Eu[e^{r_{\vtheta^{\star}}(\bar{\bx},\bU)}\bm{g}_{\vtheta^{\star}}(\bar{\bx},\bU)]
   -\Extu[e^{r_{\vtheta^{\star}}(\bX,\bU)}\bm{g}_{\vtheta^{\star}}(\bX,\bU)]\right),
  \end{align*}
  into~\eqref{grad-part-dv1}, and have 
  \begin{align}
   &\left.\nabla_{\vtheta} \left\{ \bExu[r_{\vtheta}(\bX,\bU)]-\log
   \bExtu[e^{r_{\vtheta}(\bX,\bU)}]\right\}
   \right|_{\vtheta=\vtheta^{\star}}\nonumber\\
   &\simeq(1-\epsilon)\left(\Exu[\bm{g}_{\vtheta^{\star}}(\bX,\bU)]
   -\Extu[e^{r_{\vtheta^{\star}}(\bX,\bU)}\bm{g}_{\vtheta^{\star}}(\bX,\bU)]\right)
   \nonumber \\
   &~+\epsilon\left(\bm{g}_{\vtheta^{\star}}(\bar{\bx},\bar{\bu})
   +\Extu[e^{r_{\vtheta^{\star}}(\bX,\bU)}\bm{g}_{\vtheta^{\star}}(\bX,\bU)]   
   -\Ex[e^{r_{\vtheta^{\star}}(\bX,\bar{\bu})}\bm{g}_{\vtheta^{\star}}(\bX,\bar{\bu})]
   -\Eu[e^{r_{\vtheta^{\star}}(\bar{\bx},\bU)}\bm{g}_{\vtheta^{\star}}(\bar{\bx},\bU)] \right)\nonumber\\
   &=\epsilon\left(
   \bm{g}_{\vtheta^{\star}}(\bar{\bx},\bar{\bu})
   +\Exu[\bm{g}_{\vtheta^{\star}}(\bX,\bU)]
   -\Ex[e^{r_{\vtheta^{\star}}(\bX,\bar{\bu})}\bm{g}_{\vtheta^{\star}}(\bX,\bar{\bu})
   -\Eu[e^{r_{\vtheta^{\star}}(\bar{\bx},\bU)}\bm{g}_{\vtheta^{\star}}(\bar{\bx},\bU)]\right),
   \label{grad-part-dv}
  \end{align}
  where we applied
  $\Exu[\bm{g}_{\vtheta^{\star}}(\bX,\bU)]=\Extu[e^{r_{\vtheta^{\star}}(\bX,\bU)}\bm{g}_{\vtheta^{\star}}(\bX,\bU)]$,
  which is proved from~\eqref{equiv-dv} and~\eqref{IF-tmp1} as follows:
  \begin{align*}
   \Exu[\bm{g}_{\vtheta^{\star}}(\bX,\bU)]
   -\Extu[e^{r_{\vtheta^{\star}}(\bX,\bU)}\bm{g}_{\vtheta^{\star}}(\bX,\bU)]
   &=\Exu[\bm{g}_{\vtheta^{\star}}(\bX,\bU)]
   -\frac{\Extu[e^{r_{\vtheta^{\star}}(\bX,\bU)}\bm{g}_{\vtheta^{\star}}(\bX,\bU)]}{\Extu[e^{r_{\vtheta^{\star}}(\bX,\bU)}]}\nonumber\\
   &=\nabla_{\vtheta}\left\{ \Exu[r_{\vtheta}(\bX,\bU)] -\left.\log
   \Extu[e^{r_{\vtheta}(\bX,\bU)}]\right\}\right|_{\vtheta=\vtheta^{\star}}\nonumber\\
   &=\bm{0}.
  \end{align*}
  
  Substituting~\eqref{grad-part-dv} into~\eqref{taylor-dv} yields
  \begin{align}
   \bm{0}&
   \simeq\epsilon\left\{   
   \bm{g}_{\vtheta^{\star}}(\bar{\bx},\bar{\bu})
   +\Exu[\bm{g}_{\vtheta^{\star}}(\bX,\bU)]
   -\Ex[e^{r_{\vtheta^{\star}}(\bX,\bar{\bu})}\bm{g}_{\vtheta^{\star}}(\bX,\bar{\bu})]
   -\Eu[e^{r_{\vtheta^{\star}}(\bar{\bx},\bU)}\bm{g}_{\vtheta^{\star}}(\bar{\bx},\bU)]
   \right\}\nonumber \\
   &\qquad+
   \bVdv(\vtheta_{\epsilon}-\vtheta^{\star})+O(\|\vtheta_{\epsilon}-\vtheta^{\star}\|^2)
   \label{taylor-dv2}
  \end{align}
  In the limit of $\epsilon\to{0}$,
  $O(\|\vtheta_{\epsilon}-\vtheta^{\star}\|^2)$ in~\eqref{taylor-dv2} quickly
  converges to zero and can be negligible, and $\bVdv$ approaches
  \begin{align*}
   -\Exu[\bm{g}_{\vtheta^{\star}}(\bX,\bU)\bm{g}_{\vtheta^{\star}}(\bX,\bU)^{\top}]+
   \Exu[\bm{g}_{\vtheta^{\star}}(\bX,\bU)]\Exu[\bm{g}_{\vtheta^{\star}}(\bX,\bU)]^{\top}=:\Vdv,
  \end{align*}
  where we applied~\eqref{IF-tmp1}.  By dividing both sides
  on~\eqref{taylor-dv2} by $\epsilon$ and taking the limit of
  $\epsilon\to{0}$, we obtain
  \begin{align*}
   &\Vdv\IFdv(\bar{\bx},\bar{\bu})\\
   &\qquad=
   \bm{g}_{\vtheta^{\star}}(\bar{\bx},\bar{\bu})
   +\Exu[\bm{g}_{\vtheta^{\star}}(\bX,\bU)]
   -\Ex[e^{r_{\vtheta^{\star}}(\bX,\bar{\bu})}\bm{g}_{\vtheta^{\star}}(\bX,\bar{\bu})]
   -\Eu[e^{r_{\vtheta^{\star}}(\bar{\bx},\bU)}\bm{g}_{\vtheta^{\star}}(\bar{\bx},\bU)].
  \end{align*}
  Thus, applying the inverse of $\Vdv$ yields~\eqref{IFdv2}.
  
  For the contamination model~1, we recall that the contaminated
  densities are given as follows:
  \begin{align*}
   \bar{p}(\bx,\bu)&=(1-\epsilon)p(\bx,\bu)+\epsilon\delta_{{\bar{\bu}}}(\bu)p(\bx)\\
   \bar{p}(\bx)&=p(\bx)\\
   \bar{p}(\bu)&=(1-\epsilon)p(\bu)+\epsilon\delta_{{\bar{\bu}}}(\bu).
  \end{align*}
  By following the same line of the proof, we reach the influence
  function~\eqref{IFdv1} for the contamination model~1.
 \end{proof}
 \section{Proof of Proposition~\ref{prop:strong-robust}}
 \label{app:strong-robust}
 \subsection{Main proof}
 We employ the following lemma, which is proved in
 Appendix~\ref{app:n-conditions}:
 \begin{lemma}   
  \label{prop:nu-conditions} Constant $\nu_1$ is sufficiently small
  under Assumptions~(D1-2).
 \end{lemma}
 Thus, we next focus on proving that~\eqref{strong-robust} holds when
 $\nu_1$ is sufficiently small.  Denoting $e^{r(\bx,\bu)}$ by
 $\varphi(\bx,\bu)$, for the contamination model~1, the expectations over
 $\bar{p}(\bx,\bu)$ and $\bar{p}(\bx)\bar{p}(\bu)$ can be expressed as
 \begin{align*}
  \bExu\left[\left(\frac{\varphi(\bX,\bU)^{\gamma+1}}
  {1+\varphi(\bX,\bU)^{\gamma+1}}\right)^{\frac{\gamma}{\gamma+1}}\right]
  &=(1-\epsilon)\Exu\left[\left(\frac{\varphi(\bX,\bU)^{\gamma+1}}
  {1+\varphi(\bX,\bU)^{\gamma+1}}\right)^{\frac{\gamma}{\gamma+1}}\right]\\   
  &\qquad+\epsilon\iint
   p(\bx)\delta(\bu|\bx)\left(\frac{\varphi(\bx,\bu)^{\gamma+1}}
  {1+\varphi(\bx,\bu)^{\gamma+1}} \right)^{\frac{\gamma}{\gamma+1}}
   \hspace{-3mm}\intd\bx\intd\bu\\
  \bExtu\left[\left(\frac{1} {1+\varphi(\bX,\bU)^{\gamma+1}}
  \right)^{\frac{\gamma}{\gamma+1}}\right]
  &=(1-\epsilon)\Extu\left[\left(\frac{1}
  {1+\varphi(\bX,\bU)^{\gamma+1}}\right)^{\frac{\gamma}{\gamma+1}}\right]\\
  &\qquad+\epsilon\iint{p}(\bx)\delta(\bu)\left(\frac{1}{1+\varphi(\bx,\bu)^{\gamma+1}}
 \right)^{\frac{\gamma}{\gamma+1}} \hspace{-3mm}\intd\bx\intd\bu.
 \end{align*}
 Substituting these expectations into $\bJga(\varphi)$ yields
 \begin{align*}
  \bJga(\varphi)&=-\frac{1}{\gamma}\log
  \left[\Exu\left[\left(\frac{\varphi(\bX,\bU)^{\gamma+1}}
  {1+\varphi(\bX,\bU)^{\gamma+1}}\right)^{\frac{\gamma}{\gamma+1}}\right]
  +\Extu\left[\left(\frac{1}
  {1+\varphi(\bX,\bU)^{\gamma+1}}\right)^{\frac{\gamma}{\gamma+1}}\right]
  +\frac{\epsilon}{1-\epsilon}\nu_1\right]\\
  &\qquad-\frac{1}{\gamma}\log(1-\epsilon).
 \end{align*}
 By applying $\log(z+\nu_1)=\log(z) +O(\nu_1)$ with a sufficiently small
  $\nu_1$,
 \begin{align*}
  \bJga(\varphi)&=-\frac{1}{\gamma}\log
  \left[\Exu\left[\left(\frac{\varphi(\bX,\bU)^{\gamma+1}}
  {1+\varphi(\bX,\bU)^{\gamma+1}}\right)^{\frac{\gamma}{\gamma+1}}\right]
  +\Extu\left[\left(\frac{1}
  {1+\varphi(\bX,\bU)^{\gamma+1}}\right)^{\frac{\gamma}{\gamma+1}}\right]\right]\\
  &\qquad-\frac{1}{\gamma}\log(1-\epsilon)
  +O\left(\frac{\epsilon}{1-\epsilon}\nu_1\right)\\
  &=\Jga(\varphi)-\frac{1}{\gamma}\log(1-\epsilon)
  +O\left(\frac{\epsilon}{1-\epsilon}\nu_1\right).
 \end{align*}
 This completes the proof.
 \subsection{Proof of Lemma~\ref{prop:nu-conditions}}
 \label{app:n-conditions}
 We recall that $\nu_1$ is defined by
 \begin{align*}
  \nu_1&=\underbrace{
  \iint p(\bx)\delta(\bu|\bx)\left(\frac{e^{(\gamma+1)r(\bx,\bu)}}
  {1+e^{(\gamma+1)r(\bx,\bu)}}
  \right)^{\frac{\gamma}{\gamma+1}} \intd\bx\intd\bu}_{\text{(A)}}
  +\underbrace{\iint
  p(\bx)\delta(\bu)\left(\frac{1}{1+e^{(\gamma+1)r(\bx,\bu)}}
  \right)^{\frac{\gamma}{\gamma+1}} \intd\bx\intd\bu}_{\text{(B)}}.
 \end{align*}
 Here, we complete the proof by showing below that two terms (A) and (B)
 on the right-hand side are sufficiently small under Assumptions~(D1-2).
    \paragraph{Term~(A):}
    We apply the Cauchy–Schwartz inequality to Term~(A) as
    \begin{align}
     &\iint \delta(\bu|\bx)p(\bx)\left(\frac{e^{(\gamma+1)r(\bx,\bu)}}
     {1+e^{(\gamma+1)r(\bx,\bu)}}
     \right)^{\frac{\gamma}{\gamma+1}}\hspace{-3mm}\intd\bx\intd\bu
     \nonumber\\ &\qquad\leq C_A\left(\iint\delta(\bu|\bx)p(\bx)
     e^{\gamma\{\psi(\bhx(\bx),\bhu(\bu))+{a}(\bhx(\bx))\}}
     \intd\bx\intd\bu\right)^{\frac{1}{2}},
    \end{align}
    where
    \begin{align*}
     C_A:=\left(\iint\delta(\bu|\bx)p(\bx)
     \left(\frac{e^{(\gamma+1)\{\psi(\bhx(\bx),\bhu(\bu))+{a}(\bhx(\bx))\}/2}}
     {e^{-(\gamma+1)b(\bhu(\bu))}
     +e^{(\gamma+1)\{\psi(\bhx(\bx),\bhu(\bu))+a(\bhx(\bx))\}}}
     \right)^{\frac{2\gamma}{\gamma+1}}
     \intd\bx\intd\bu\right)^{\frac{1}{2}}.
    \end{align*}
    By Assumption~(D1), the constant $C_A$ is finite. Thus,
    Assumption~(D2) ensures that Term~(A) is sufficiently small.
    \paragraph{Term~(B):}
    Applying the Cauchy–Schwartz inequality yields the upper bound of
    Term~(B) as
    \begin{align}
     &\iint \delta(\bu)p(\bx)\left(\frac{1}{1+e^{(\gamma+1)r(\bx,\bu)}}
     \right)^{\frac{\gamma}{\gamma+1}}\hspace{-3mm}\intd\bx\intd\bu
     \leq C_{\rm B}\left( \int
     \delta(\bu)e^{-\gamma{b}(\bhu(\bu))}\intd\bu \right)^{\frac{1}{2}},
     \label{ineq:termB}
    \end{align}
    where
    \begin{align*}
     C_{\rm B}:=\left(\iint\delta(\bu)p(\bx)
     \left(\frac{e^{-(\gamma+1){b}(\bhu(\bu))/2}}
     {e^{-(\gamma+1)b(\bhu(\bu))}
     +e^{(\gamma+1)\{\psi(\bhx(\bx),\bhu(\bu))+a(\bhx(\bx))\}}}
     \right)^{\frac{2\gamma}{\gamma+1}}
     \intd\bx\intd\bu\right)^{\frac{1}{2}}.
    \end{align*}
    By~\eqref{ineq:termB} and Assumption~(D2), Term~(B) is also
    sufficiently small.
    
 \section{Semistrong robustness for the contamination model~2}
 \label{app:semistrong-robust}
 For the contamination model~2, the following proposition is
 established:
   \begin{proposition}
    \label{prop:strong-robust2}~Let us define a constant $\nu_2$ as
    \begin{align*}
     &\nu_2:=\iint \delta(\bx,\bu)
     \left(\frac{e^{(\gamma+1)r(\bx,\bu)}}{1+e^{(\gamma+1)r(\bx,\bu)}}
     \right)^{\frac{\gamma}{\gamma+1}}\hspace{-3mm} \intd\bx\intd\bu
     +(1-\epsilon)\iint
     p(\bx)\delta(\bu)
     \left(\frac{1}{1+e^{(\gamma+1)r(\bx,\bu)}}
     \right)^{\frac{\gamma}{\gamma+1}}\hspace{-3mm} \intd\bx\intd\bu\\
     &+(1-\epsilon)\iint\delta(\bx)p(\bu)
     \left(\frac{1}{1+e^{(\gamma+1)r(\bx,\bu)}}
     \right)^{\frac{\gamma}{\gamma+1}}\hspace{-3mm} \intd\bx\intd\bu
     +\epsilon\iint\delta(\bx)\delta(\bu)
     \left(\frac{1}{1+e^{(\gamma+1)r(\bx,\bu)}}
     \right)^{\frac{\gamma}{\gamma+1}}\hspace{-3mm}\intd\bx\intd\bu.
    \end{align*}
    We make the following assumptions:
    \begin{enumerate}[(D$'$1)]
     \item Assume that
	   \begin{align*}
	    &\iint\delta(\bx,\bu)
	    \left(\frac{e^{(\gamma+1)\psi(\bhx(\bx),\bhu(\bu))/2}}
	    {e^{-(\gamma+1)\{a(\bhx(\bx))+b(\bhu(\bu))\}}
	    +e^{(\gamma+1)\psi(\bhx(\bx),\bhu(\bu))}}
	    \right)^{\frac{2\gamma}{\gamma+1}}
	    \intd\bx\intd\bu<\infty,\\
	    &\iint \{p(\bx)\delta(\bu)+\delta(\bx)p(\bu)+\delta(\bx)\delta(\bu)\}\\
	    &\qquad\times\left(\frac{e^{-(\gamma+1)\{a(\bhx(\bx))+b(\bhu(\bu))\}/2}}
	    {e^{-(\gamma+1)\{a(\bhx(\bx))+b(\bhu(\bu))\}}
	   +e^{(\gamma+1)\psi(\bhx(\bx),\bhu(\bu))}}
	    \right)^{\frac{2\gamma}{\gamma+1}}
	    \intd\bx\intd\bu<\infty
	   \end{align*}
	   
     \item The following integrals are sufficiently small.
	   \begin{align*}
	    \iint\delta(\bx,\bu)
	    e^{\gamma\psi(\bhx(\bx),\bhu(\bu))}\intd\bu\intd\bx,
	    ~\int\delta(\bx)
	    e^{-\gamma{a}(\bhx(\bx))}\intd\bu\intd\bx
	    ~~\text{\and}~~
	    \int\delta(\bu)e^{-\gamma{b}(\bhu(\bu))}\intd\bu.
	   \end{align*}	  
    \end{enumerate}
    Then, it holds that $\nu_2$ is sufficiently small and
    \begin{align}
     \bJga(r)&=-\frac{1}{\gamma}\log\left[
     \exp\left(-\gamma\Jga(r)\right)-\epsilon{I}(r)\right]
     -\frac{1}{\gamma}\log(1-\epsilon)
     +O\left(\frac{\epsilon}{1-\epsilon}\nu_2\right),
     \label{semistrong-robust}
    \end{align}
    where 
    \begin{align*}
     I(r):=\iint p(\bx)p(\bu)
     \left(\frac{1}{1+e^{(\gamma+1)r(\bx,\bu)}}
     \right)^{\frac{\gamma}{\gamma+1}} \intd\bx\intd\bu.
    \end{align*}
   \end{proposition}
   The proof is essentially the same as
   Proposition~\ref{prop:strong-robust}, and thus is omitted.
   Proposition~\ref{prop:strong-robust2} indicates when $\nu_2$ is
   sufficiently small, minimization of $\bJga(r)$ is almost equal to
   minimization of
   \begin{align}
    -\frac{1}{\gamma}\log\left[
    e^{-\gamma\Jga(r)}-\epsilon{I}(r)\right].
    \label{semistrong-robust}
   \end{align}
   Unlike the contamination model~1
   (Proposition~\ref{prop:strong-robust}), there exists an extra term
   $\epsilon{I}(r)$ inside the logarithm
   in~\eqref{semistrong-robust}. However, this extra term may not cause
   any bias in terms of representation learning because the minimizer
   of~\eqref{semistrong-robust} is given by
   \begin{align}
    \log \frac{p(\bx,\bu)}{p(\bx)p(\bu)}-\log(1-\epsilon).
    \label{semistrong-solution}
   \end{align}
   Since our goal of representation learning is to estimate the
   log-density ratio up to a constant, the
   minimizer~\eqref{semistrong-solution} means that robust
   representation learning would be possible even for the contamination
   model~2 under the condition that $\nu_2$ is sufficiently small. In
   addition, again, the contamination ratio $\epsilon$ is not
   necessarily assumed to be small.
   
   Assumptions~(D$'$1-2) have the same implications as
   Assumptions~(D1-2) in Proposition~\ref{prop:strong-robust} (See
   Section~\ref{ssec:strong-robust}), and thus would reflect the typical
   contamination of outliers.  Thus, the condition that $\nu_2$ is
   sufficiently small would be fairly reasonable.  Indeed,
   Section~\ref{sec:exp} experimentally demonstrates that our method for
   representation learning is very robust against outliers.
   
 \bibliography{../research/papers}

\end{document}